\documentclass[a4paper,12pt,titlepage,oneside]{kumsthesis}
\pdfoutput=1

\usepackage[utf8]{inputenc}
\UseRawInputEncoding
\usepackage{url}
\usepackage[bottom]{footmisc}
\usepackage[caption=false,font=footnotesize]{subfig}
\usepackage{natbib}
\usepackage{amsmath}
\usepackage{amssymb}
\usepackage{amsfonts}
\usepackage{breakcites}
\usepackage{color}
\usepackage[dvipsnames]{xcolor}
\usepackage{etoolbox}
\usepackage{dblfloatfix}
\usepackage{bbm}
\usepackage{xspace}
\usepackage{bbm}
\usepackage[separate-uncertainty=true]{siunitx}
\usepackage{booktabs}
\usepackage{rotating}
\usepackage[accsupp]{axessibility}
\usepackage{multirow}
\usepackage{multicol}
\usepackage{caption}
\usepackage{float}
\usepackage{nccmath}
\usepackage[hidelinks]{hyperref}

\graphicspath{{./Figures/}}

\usepackage{xspace}
\usepackage{epstopdf}

\newcommand \COMMENT  [1] {}       %

\begin{document}

\newcommand{\Perp}{\perp\!\!\! \perp}
\newcommand{\bK}{\mathbf{K}}
\newcommand{\bX}{\mathbf{X}}
\newcommand{\bY}{\mathbf{Y}}
\newcommand{\bk}{\mathbf{k}}
\newcommand{\bx}{\mathbf{x}}
\newcommand{\by}{\mathbf{y}}
\newcommand{\bhy}{\hat{\mathbf{y}}}
\newcommand{\bty}{\tilde{\mathbf{y}}}
\newcommand{\bG}{\mathbf{G}}
\newcommand{\bI}{\mathbf{I}}
\newcommand{\bg}{\mathbf{g}}
\newcommand{\bS}{\mathbf{S}}
\newcommand{\bs}{\mathbf{s}}
\newcommand{\bM}{\mathbf{M}}
\newcommand{\bw}{\mathbf{w}}
\newcommand{\eye}{\mathbf{I}}
\newcommand{\bU}{\mathbf{U}}
\newcommand{\bV}{\mathbf{V}}
\newcommand{\bW}{\mathbf{W}}
\newcommand{\bn}{\mathbf{n}}
\newcommand{\bv}{\mathbf{v}}
\newcommand{\bwv}{\mathbf{wv}}
\newcommand{\bq}{\mathbf{q}}
\newcommand{\bR}{\mathbf{R}}
\newcommand{\bi}{\mathbf{i}}
\newcommand{\bj}{\mathbf{j}}
\newcommand{\bp}{\mathbf{p}}
\newcommand{\bt}{\mathbf{t}}
\newcommand{\bJ}{\mathbf{J}}
\newcommand{\bu}{\mathbf{u}}
\newcommand{\bB}{\mathbf{B}}
\newcommand{\bD}{\mathbf{D}}
\newcommand{\bz}{\mathbf{z}}
\newcommand{\bP}{\mathbf{P}}
\newcommand{\bC}{\mathbf{C}}
\newcommand{\bA}{\mathbf{A}}
\newcommand{\bZ}{\mathbf{Z}}
\newcommand{\bff}{\mathbf{f}}
\newcommand{\bF}{\mathbf{F}}
\newcommand{\bo}{\mathbf{o}}
\newcommand{\bO}{\mathbf{O}}
\newcommand{\bc}{\mathbf{c}}
\newcommand{\bm}{\mathbf{m}}
\newcommand{\bT}{\mathbf{T}}
\newcommand{\bQ}{\mathbf{Q}}
\newcommand{\bL}{\mathbf{L}}
\newcommand{\bl}{\mathbf{l}}
\newcommand{\ba}{\mathbf{a}}
\newcommand{\bE}{\mathbf{E}}
\newcommand{\bH}{\mathbf{H}}
\newcommand{\bd}{\mathbf{d}}
\newcommand{\br}{\mathbf{r}}
\newcommand{\be}{\mathbf{e}}
\newcommand{\bb}{\mathbf{b}}
\newcommand{\bh}{\mathbf{h}}
\newcommand{\bhh}{\hat{\mathbf{h}}}
\newcommand{\btheta}{\boldsymbol{\theta}}
\newcommand{\bTheta}{\boldsymbol{\Theta}}
\newcommand{\bpi}{\boldsymbol{\pi}}
\newcommand{\bphi}{\boldsymbol{\phi}}
\newcommand{\bpsi}{\boldsymbol{\psi}}
\newcommand{\bPhi}{\boldsymbol{\Phi}}
\newcommand{\bmu}{\boldsymbol{\mu}}
\newcommand{\bsigma}{\boldsymbol{\sigma}}
\newcommand{\bSigma}{\boldsymbol{\Sigma}}
\newcommand{\bGamma}{\boldsymbol{\Gamma}}
\newcommand{\bbeta}{\boldsymbol{\beta}}
\newcommand{\bomega}{\boldsymbol{\omega}}
\newcommand{\blambda}{\boldsymbol{\lambda}}
\newcommand{\bLambda}{\boldsymbol{\Lambda}}
\newcommand{\bkappa}{\boldsymbol{\kappa}}
\newcommand{\btau}{\boldsymbol{\tau}}
\newcommand{\balpha}{\boldsymbol{\alpha}}
\newcommand{\nR}{\mathbb{R}}
\newcommand{\nN}{\mathbb{N}}
\newcommand{\nL}{\mathbb{L}}
\newcommand{\nE}{\mathbb{E}}
\newcommand{\cN}{\mathcal{N}}
\newcommand{\cM}{\mathcal{M}}
\newcommand{\cR}{\mathcal{R}}
\newcommand{\cB}{\mathcal{B}}
\newcommand{\cL}{\mathcal{L}}
\newcommand{\cH}{\mathcal{H}}
\newcommand{\cS}{\mathcal{S}}
\newcommand{\cT}{\mathcal{T}}
\newcommand{\cO}{\mathcal{O}}
\newcommand{\cC}{\mathcal{C}}
\newcommand{\cP}{\mathcal{P}}
\newcommand{\cE}{\mathcal{E}}
\newcommand{\cI}{\mathcal{I}}
\newcommand{\cF}{\mathcal{F}}
\newcommand{\cK}{\mathcal{K}}
\newcommand{\cY}{\mathcal{Y}}
\newcommand{\cX}{\mathcal{X}}
\def\bgamma{\boldsymbol\gamma}

\newcommand{\specialcell}[2][c]{%
  \begin{tabular}[#1]{@{}c@{}}#2\end{tabular}}

\newcommand{\figref}[1]{\Fig~\ref{#1}}
\newcommand{\secref}[1]{Section~\ref{#1}}
\newcommand{\algref}[1]{Algorithm~\ref{#1}}
\newcommand{\eqnref}[1]{Eq.~\eqref{#1}}
\newcommand{\tabref}[1]{Table~\ref{#1}}

\newcommand{\rulesep}{\unskip\ \vrule\ }

\newcommand{\KLD}[2]{D_{\mathrm{KL}} \Big(#1 \mid\mid #2 \Big)}

\renewcommand{\b}{\ensuremath{\mathbf}}

\def\mc{\mathcal}
\def\mb{\mathbf}

\newcommand{\T}{^{\raisemath{-1pt}{\mathsf{T}}}}

\makeatletter
\DeclareRobustCommand\onedot{\futurelet\@let@token\@onedot}
\def\@onedot{\ifx\@let@token.\else.\null\fi\xspace}
\def\eg{e.g\onedot} \def\Eg{E.g\onedot}
\def\ie{i.e\onedot} \def\Ie{I.e\onedot}
\def\cf{cf\onedot} \def\Cf{Cf\onedot}
\def\etc{etc\onedot} \def\vs{vs\onedot}
\def\wrt{wrt\onedot}
\def\dof{d.o.f\onedot}
\def\etal{et~al\onedot} \def\iid{i.i.d\onedot}
\def\Fig{Fig\onedot} \def\Eqn{Eqn\onedot} \def\Sec{Sec\onedot} \def\Alg{Alg\onedot}
\makeatother

\newcommand{\xdownarrow}[1]{%
  {\left\downarrow\vbox to #1{}\right.\kern-\nulldelimiterspace}
}

\newcommand{\xuparrow}[1]{%
  {\left\uparrow\vbox to #1{}\right.\kern-\nulldelimiterspace}
}

\newcommand*\rot{\rotatebox{90}}
\newcommand{\boldparagraph}[1]{\vspace{0.15cm}\noindent{\bf #1:} }
\newcommand{\boldquestion}[1]{\vspace{0.2cm}\noindent{\bf #1} }

\newcommand{\ka}[1]{ \noindent {\color{blue} {\bf Kaan:} {#1}} } 
\newcommand{\ftm}[1]{ \noindent {\color{magenta} {\bf Fatma:} {#1}} }

\Title{Stochastic Future Prediction in Real World Driving Scenarios} \Author{Adil Kaan Akan} \Year{June 3, 2022}
\TTitle{Y\"{u}ksek Lisans Tez Ba\c{s}l{\i}\u{g}{\i}} \TYear{3 Haziran 2022}
\Program{Computer Engineering}
\TProgram{Bilgisayar M\"{u}hendisli\u{g}i}
\Signature{Assist. Prof. Fatma G\"{u}ney (Advisor)}
\Signature{Assoc. Prof. Aykut Erdem}
\Signature{Prof. J\"{u}rgen Gall}

\prelimpages
\titlepage
\thesissignaturepage
\dedication{To all my beloved ones}
\abstract{
Uncertainty plays a key role in future prediction. The future is uncertain. That means there might be many possible futures. A future prediction method should cover the whole possibilities to be robust. In autonomous driving, covering multiple modes in the prediction part is crucially important to make safety-critical decisions. Although computer vision systems have advanced tremendously in recent years, future prediction remains difficult today. Several examples are uncertainty of the future, the requirement of full scene understanding, and the noisy outputs space. In this thesis, we propose solutions to these challenges by modeling the motion explicitly in a stochastic way and learning the temporal dynamics in a latent space.

Firstly, we propose to use the motion information in the scene in stochastic video prediction. By separately modeling the appearance of the scene and the motion in the scene, the static and the dynamic parts are partitioned. The dynamic part is predicted by the explicit motion. Since the motion is predicted, the noisy pixel space is not used. We demonstrate the benefits of using the motion information scene explicitly.

Secondly, we propose to separate the scene into the 3D structure and the motion, which correspond to static and dynamic parts, in video prediction in driving scenarios. The static part is handled by the 3D structure and the motion of the ego-vehicle, whereas the dynamic part is handled by the remaining motion in the scene. The remaining motion is captured by explicitly conditioning the dynamic part on top of the static one. We demonstrate the improvements of the conditioning and structure-aware separation.

Finally, motivated by the compact representation, we propose a method for stochastic future prediction in Bird's-Eye View representation. Using a new formulation, we approach the problem as a state prediction rather than a trajectory prediction. For that purpose, we choose to use more powerful label-aware latent variables to generate more diverse and admissible future trajectories. Extensive evaluations show that both the diversity and the accuracy of the future trajectories significantly improved, especially in challenging cases of spatially far regions and temporally long spans.
}
\oz{Belirsizlik gelecek tahmininde \c{c}ok kritik bir rol oynamaktad{\i}r. Gelecek belirsizdir ve bu bir\c{c}ok gelecek ihtimali oldu\u{g}unu g\"{o}sterir. Bir gelecek tahmini y\"{o}ntemi g\"{u}venilir olabilmek i\c{c}in b\"{u}t\"{u}n ihtimalleri kapsamal{\i}d{\i}r. Otonom s\"{u}r\"{u}\c{s}te emniyet a\c{c}{\i}s{\i}ndan kritik kararlar verebilmek i\c{c}in t\"{u}m modlar{\i} kapsamak a\c{s}{\i}r{\i} derecede \"{o}nemlidir. Bilgisayarl{\i} g\"{o}r\"{u} y\"{o}ntemleri son y{\i}llarda \c{c}ok h{\i}zl{\i} \c{s}ekilde ilerse de gelecek tahmini hala zor olmaya devam etmektedir. Bu zorluklardan birka\c{c} tanesi gelece\u{g}in belirsizli\u{g}i, t\"{u}m sahnenin anla\c{s}{\i}lmas{\i}n{\i}n gereklili\u{g}i ve piksellerin g\"{u}r\"{u}lt\"{u}l\"{u} olmas{\i}d{\i}r. Bu tezde, bu problemlere hareket bilgisini modelleyerek ve zamansal dinamikleri \"{o}\u{g}renerek \c{c}\"{o}z\"{u}mler sunuyoruz.

\.{I}lk olarak, sahnedeki gelecek bilgisinin a\c{c}{\i}k bir \c{s}ekilde i\c{s}lenmesini sunarak video alan{\i}nda gelecek tahmini y\"{o}ntemi yapan bir y\"{o}ntem sunuyoruz. Bu y\"{o}ntem g\"{o}r\"{u}n\"{u}\c{s}\"{u} ve hareketi ayr{\i} ayr{\i} modelleyerek sahnenin dura\u{g}an ve hareketli k{\i}s{\i}mlar{\i}n{\i} ay{\i}rmaktad{\i}r. Bu sayede piksellerin g\"{u}r\"{u}lt\"{u}l\"{u} uzay{\i}ndan ka\c{c}arak hareket bilgisiyle tahmin yapmak m\"{u}mk\"{u}n olmu\c{s}tur. Bu y\"{o}ntemin kazan\c{c}lar{\i}n{\i} detayl{\i} bir \c{s}ekilde g\"{o}stermekteyiz.

\.{I}kinci olarak, sahneyi 3 boyutlu yap{\i} ve artakalan hareket olacak \c{s}ekilde ikiye ay{\i}rmay{\i} sunuyoruz. Bu ayr{\i}\c{s}{\i}m ile sahnedeki dura\u{g}an ve hareketli k{\i}s{\i}mlar{\i} ayr{\i} ayr{\i} modelleyebiliyoruz. Dura\u{g}an k{\i}s{\i}mlar 3 boyutlu yap{\i}yla ve hareketli k{\i}s{\i}mlar ise artakalan hareket bilgisiyle tahmin edilmektedir. Artakalan hareket i\c{c}in \"{o}nce 3 boyutlu yap{\i}daki hareketi tahmin edip bu hareket \"{u}st\"{u}ne ko\c{s}ullanarak tahmin yap{\i}lmaktad{\i}r. Bu ayr{\i}\c{s}{\i}m{\i}n ve ko\c{s}ullanman{\i}n etkilerini detayl{\i} bir \c{s}ekilde g\"{o}stermekteyiz.

Son olarak, kompakt g\"{o}sterimden etkilenerek, gelecek tahmini problemini ku\c{s}bak{\i}\c{s}{\i} uzaya ta\c{s}{\i}may{\i} sunuyoruz. Yeni bir formulasyon kullanarak gelecek tahmininden ziyade gelecek durum tahmini yapman{\i}n problemi daha da kolayla\c{s}t{\i}raca\u{g}{\i}n{\i} sunmaktay{\i}z. Bu ama\c{c}la, gelecek bilgisinin kullan{\i}ld{\i}\u{g}{\i} bir da\u{g}{\i}l{\i}m kullanarak daha da\u{g}{\i}n{\i}k ve kurallara uygun y\"{o}r\"{u}nge tahminleri yap{\i}lmaktad{\i}r. Yapt{\i}\u{g}{\i}m{\i}z detayl{\i} deneyler ile bu y\"{o}ntemin daha ger\c{c}ek\c{c}i tahminler yapt{\i}\u{g}{\i} g\"{o}sterilmi\c{s}tir.}

\acknowledgments{I would like to express my deepest appreciation and gratitude to Asst. Prof. Fatma G\"{u}ney. This work would not have been possible without her continuous support and motivation. I also want to express my deepest gratitude for being not only my advisor but also my mentor. It has been a privilege to work with you.

I would like to thank the rest of my thesis committee: Assoc. Prof. Aykut Erdem, Prof. J\"{u}rgen Gall for agreeing to be in my committee and for evaluating my work.

Also, I would like to thank Assoc. Prof. Aykut Erdem and Assoc. Prof. Erkut Erdem for their contributions to this work.

I would like to thank all the members of AVG. With a special mention to Sadra Safadoust, for all their help and contribution to this work, \c{c}a\u{g}an Selim \c{C}oban, Ege Onat \"{o}zs\"{u}er, for guiding me throughout my studies.

Also, I would like to thank all of my friends. Especially, Berkay U\u{g}ur \c{S}enocak for helpful discussions and code reviews, Batuhan Ulu\c{c}e\c{s}me for his guidance in Ko\c{c} University, G\"{u}ne\c{s} \c{C}epi\c{c} and Melih Kaan Kayap{\i}nar for their support, Metehan G\"{u}lp{\i}nar for his entertaining conversations in hard times, Salih Karag\"{o}z for his constructive feedback.

I would like to thank my mother, Dilek, my father, Mehmet, and my little sister, Ay\c{c}a, for their unconditional support and the strength you have given to me. I always felt your support and motivation with me. It would have been very hard to complete this work with your belief in me.

Last but not least, I would like to thank Aleyna G\"{u}rkan, who became everything to me, supported me for everything I wanted to pursue, motivated me to stay at night for work, and tolerated me around deadlines. I honestly could not bear the burden without her endless encouragement, unwavering motivation, joyful and helpful accompany and warm love.
}

\tableofcontents
\listoftables
\listoffigures
\abbreviations{
\begin{tabular}{lp{1cm}l}
    PSNR & & Peak Signal to Noise Ratio \\
    SSIM & & Structural Similarity \\
    LPIPS & & Learned Perceptual Image Patch Similarity \\ 
	FVD & & Frechet Video Distance \\
	LSTM & & Long Short Term Memory Networks \\
	MLP & &  Multi Layer Perceptron \\
	2D & &  2 Dimensional \\
    3D & &   3 Dimensional \\
    CNN & &   Convolutional Neural Networks \\
\end{tabular}
}
\textpages

\chapter{Introduction}
\label{chapter:introduction}

Videos can be defined as a set of image frames, and naturally, they contain much more information than images. The amount of information is not limited to pixels, and it also includes the temporal information in the videos, \emph{namely} motion. The temporal information both eases and complicates the understanding of the videos. The information in the image level scales up when the number of image frames in the video increases. However, reasoning about the video gets more manageable since the amount of information increase with the number of image frames. For example, objects which are not fully visible in a few frames can be seen in the other frames. Videos include not only foreground objects that are moving but also background, which has complex information in most of the scenarios, \eg driving scenarios. To understand the videos fully, one needs to figure out the environment, present agents or foreground objects that have an independent motion, relations between the objects, and motion of the background, if there is any. Moreover, the amount of information is not limited to these. The information at the pixel level is enormous because of the large resolution of the videos. 

Since birth, we have a constant visual stream of data in our brains, and we naturally learn to process visual information like videos. As soon as we open our eyes to the world, we monitor the world around us and try to reason about it. We, humans, with this much experience, can reason about the videos without much effort. We can understand the objects moving around us and their relation to other objects. Our brain is capable of understanding the scene around us at the video level.

In autonomous driving, the ego-vehicle perceives the world as images at each time step. At each time step, the ego-vehicle takes pictures of the environment around it and makes several decisions according to the surrounding environment such as acceleration, stopping or turning. Before making these decisions, the ego-vehicle needs to understand the world around it, detect and track the moving objects, predict their future trajectories and plan the optimal route. For these tasks, the ego-vehicle needs to consider not only the information at the current time but also the past information. Since the ego-vehicle uses the image at each time steps, it has a stack of images captured at separate time steps, \emph{a video}. 

The amount of information in the autonomous driving is immense and to fully solve the problem, the ego-agent needs to be proficient at several sub-tasks such as detection, prediction and planning. Image level reasoning may not suitable to this end and video level understanding might be necessary to conquer the problem because the temporal information in the video can introduce a lot meaningful cues.

To capture all the information in the videos, prediction might be appropriate task. Humans can easily predict the future of the video because they can grasp all the information in the current and the past frames of the video. They can reason about the background and the objects, their motion, and their relation to other objects. Therefore, the prediction task might be the suitable task to fully understand because one needs to consume all the information until the current time before the prediction.

In this thesis, we focus on video future prediction in different representations. To predict the future, an agent first needs to understand its environment and its surrounding objects. Moreover, it must know the past states of both itself and surrounding objects. In order to predict the future, the agent should understand its current and past states, relate the environment and the surrounding objects to each other, then, conditioned on this information, and infer the future. Future prediction has been studied in different representations such as video frames \cite{Akan2021ICCV, Denton2018ICML, Franceschi2020ICML, Finn2016NeurIPS, Babaeizadeh2018ICLR, Lee2018ARXIV, Akan2022SLAMP3D, Akan2022SLAMP3D_ARXIV, Castrejon2019ICCV, Villegas2019NeurIPS}, key points \cite{villegas2017ICML, Minderer2019NeurIPS} , coordinate space \cite{liu2021multimodal, gu2021densetnt, zhao2020tnt, gao2020vectornet, tang2019multiple, Aydemir2022Arxiv} and Bird’s-eye view \cite{Hu2021ICCV, Akan2022Arxiv, Hendy2020CVPRW}.  In the first part of the thesis, we propose a novel method for stochastic video prediction that exploits the motion history in the scene. In the second part, we propose another video prediction method that decomposes the scene into static and dynamic parts and models each part using domain knowledge of real-world driving datasets. In the third part, we propose a future instance segmentation method in Bird's-eye view representation from multiple camera images. In this part, we briefly analyze the problems. 

\section{Motivation}

\subsection{Video Prediction}

Videos contain visual information enriched by motion.
Motion is a useful cue for reasoning about human activities
or interactions between objects in a video. Given a few
initial frames of a video, our goal is to predict several frames into the future, as realistically as possible. By looking at a few frames, humans can predict what will happen next. Surprisingly, they can even attribute semantic meanings to random dots and recognize motion patterns~\cite{Johansson1973}. This shows
the importance of motion to infer the dynamics of the video
and to predict the future frames.

\begin{figure}
    \small
    \centering
    \begin{tabular}{c|cccc}
            \multicolumn{1}{c|}{\textbf{Context}}  &
            \multicolumn{4}{|c}{\textbf{Predicted Frames}} 
            \\
            $t = 2$ &
            $t = 3$ & 
            $t = 5$ & 
            $t = 10$ &
            $t = 20$ \\
            \hline
            \raisebox{-.4\height}{\includegraphics[scale=0.56]{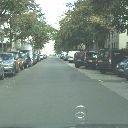}} &
            \raisebox{-.4\height}{\includegraphics[scale=0.56]{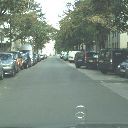}} & 
            \raisebox{-.4\height}{\includegraphics[scale=0.56]{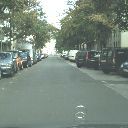}} &
            \raisebox{-.4\height}{\includegraphics[scale=0.56]{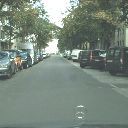}} &
            \raisebox{-.4\height}{\includegraphics[scale=0.56]{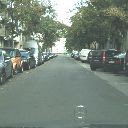}}
            \\[15.5pt]
    \end{tabular}
    \caption[Video Prediction Problem Definition]{
    The goal in video prediction is to predict the next several frames into the future given a few initial frames. In this example, the first two frames are given and the model tries to predict the next 18 of them. The images are from Cityscapes~\cite{Cordts2016CVPR} dataset.
    }
    \label{fig:video-pred-prob-def}
\end{figure}

Video prediction can be defined as predicting future video frames, given a few initial ones. An example definition of the problem can be seen in \figref{fig:video-pred-prob-def}.

Motion cues have been heavily utilized for future frame
prediction in computer vision. A common approach is to factorize the video into static and dynamic components~\cite{Walker2015ICCV, Liu2017ICCV, Lu2017CVPR, Fan2019AAAI, Gao2019ICCV, Lotter2017ICLR, Brabandere2016NeurIPS, Vondrick2017CVPR}. First, most of the previous methods are deterministic and fail to model the uncertainty of the future. Second, motion is typically interpreted as local changes from one frame to the next. However, changes in
motion follow certain patterns when observed over some
time interval. Consider scenarios where objects move with
near-constant velocity, or humans repeating atomic actions
in videos. Regularities in motion can be very informative
for future frame prediction.

Video prediction methods predicts the target future frames directly in pixel space. That is, they encode both the motion and content of the video into latent variables or specific variables and decode those variable into images using neural networks. However, directly modeling latent variables to the image space is difficult because of the large resolution of the images and the number of possible values that a pixel can take. Several methods~\cite{villegas2017ICML, Minderer2019NeurIPS} first try to transfer the prediction task into a more meaningful space, \eg key point space, and then from that space, future image frames are decoded. As we cover before, motion is a crucial information in a video. The motion in the video would ease the process of prediction. The future of the video can be predicted first transferring prediction task into the motion space and then, the target frames can be decoded, similar to key point approaches. Most of the motion in a video is global across the video rather than a local changes throughout the video. Therefore, by propagating the past motion information into the future steps can lead better predictions. 

Stochastic methods have been proposed to model the inherent uncertainty of the future in videos. Earlier methods encode the dynamics of the video in stochastic latent variables which are decoded to future frames in a deterministic way~\cite{Denton2018ICML}. Instead, we first assume that both appearance and motion are encoded in the stochastic latent variables and decode them separately into appearance and motion predictions in a deterministic way. Inspired by the previous deterministic methods~\cite{Finn2016NeurIPS, Liu2017ICCV, Gao2019ICCV}, we also estimate a mask relating the two. Both appearance and motion decoders are expected to predict the full frame but they might fail due to occlusions around motion boundaries. Intuitively, we predict a probabilistic mask from the results of the appearance and motion decoders to combine them into a more accurate final prediction. Our model learns to use motion cues in the dynamic parts and relies on appearance in the occluded regions.

In the first two parts of this thesis, we focus on the problem of stochastic video prediction. Different from classical video prediction, in stochastic video prediction, the aim is to generate different futures given the same initial frames. That is, the models learn the multimodality of the data and generate a diverse set of future frames. There are different challenges in this problem. For example, the most important challenge is the uncertainty of the future. Given a few initial frames of video, the future of the video is unknown to models, and there are multiple possible futures. Therefore, the models need to learn the underlying multimodal distribution to generate a set of diverse futures. In the first two chapters, we propose two different stochastic video prediction methods.

\subsection{Bird's-Eye View Representation}

Future prediction with sequential visual data has been studied from different perspectives. Stochastic video prediction methods, as we cover in  operate in the pixel space and learn to predict future frames conditioned on the previous frames. These methods achieve impressive results by modeling the uncertainty of the future with stochasticity, however, long-term predictions in real-world sequences tend to be quite blurry due to the complexity of directly predicting pixels. A more practical approach is to consider a compact representation that is tightly connected to the modalities required for the downstream application. In self-driving, understanding the 3D properties of the scene and the motion of other agents plays a key role in future predictions. The bird's-eye view~(BEV) representation meets these requirements by first fusing information from multiple cameras into a 3D point cloud and then projecting the points to the ground plane~\cite{Philion2020ECCV}. This leads to a compact representation where the future location and motion of multiple agents in the scene can be reliably predicted. In this paper, we explore the potential of stochastic future prediction for self-driving to produce admissible and diverse results in long sequences with an efficient and compact BEV representation. 

Future prediction from the BEV representation has been recently proposed in FIERY~\cite{Hu2021ICCV}. The BEV representations of past frames are first related in a temporal model and then used to learn two distributions representing the present and the future. Based on sampling from one of these distributions depending on train or test time, various future modalities are predicted with a recurrent model. For planning, long term multiple future predictions are crucial, however, the predictions of FIERY degrade over longer time spans due to the limited representation capability of a single distribution for increasing diversity in longer predictions. Following the official implementation, the predictions do not differ significantly based on random samples but converge to the mean, therefore lack diversity. We start from the same BEV representation and predict the same output modalities to be comparable. Differently, instead of two distributions for the present and the future, we propose to learn time-dependent distributions by predicting a residual change at each time step to better capture long-term dynamics. Furthermore, we show that by sampling a random variable at each time step, we can increase the diversity of future predictions while still being accurate and efficient. For efficiency, we use a state-space model \cite{Murphy2023Prob} instead of costly auto-regressive models.

The main idea behind the efficiency of the state-space model is to decouple the learning of dynamics and the generation of predictions~\cite{Franceschi2020ICML}. We first learn a low dimensional latent space from the BEV representation to capture the dynamics and then learn to decode some output modalities from the predictions in that latent space. These output modalities show the location and the motion of the agents in the scene.
We can train our dynamics model independently by learning to match the BEV representations of future frames that are predicted by our model to the ones that are extracted from a pre-trained BEV segmentation model~\cite{Philion2020ECCV}. Through residual updates to the latent space, our model can capture the changes to the BEV representation over time. This way, the only information we use from the future is the BEV representation of future frames.
Another option is to encode the future modalities to predict and provide the model with this encoded representation to learn a future distribution \cite{Hu2021ICCV}. We experiment with both options in this paper. While providing labels in the future distribution leads to better performance, learning the dynamics becomes dependent on the labels. From the BEV predictions, we train a decoder to predict the location and the motion of future instances in a supervised manner. These output modalities increase the interpretability of the predicted BEV representations that can be used for learning a driving policy in future work.

\begin{figure}[ht!]
    \centering
    \includegraphics[width=\linewidth]{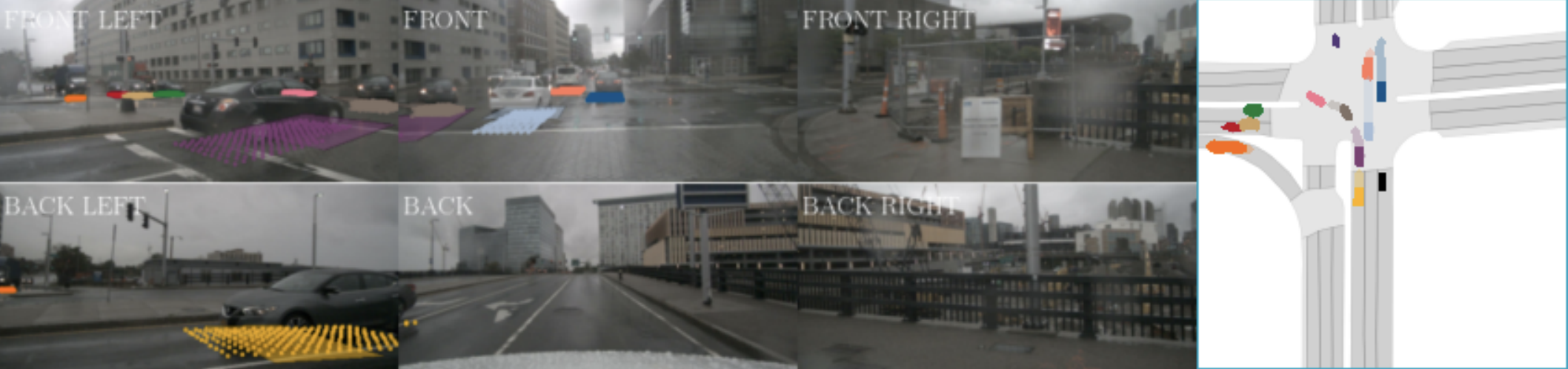}
    \caption[Bird's-Eye View (BEV) Representation]{\textbf{Bird's-Eye View (BEV) Representation} BEV representation can be constructed from multiple cameras of the ego-vehicle. The images are from nuScenes dataset~\cite{Caesar2020CVPR} and for the BEV predictions and 2D projected points FIERY~\cite{Hu2021ICCV} is used.
    }
    \label{fig:bev-pred-prob-def}
\end{figure}

In the last part of this thesis, we focus on future prediction in Bird's-Eye view~(BEV) space. Instead of directly predicting the future in pixel space, we choose to use a more compact representation, namely BEV. In self-driving, understanding the 3D properties of the scene and the motion of other agents plays a key role in future predictions. The bird's-eye view~(BEV) representation meets these requirements by first fusing information from multiple cameras into a 3D point cloud and then projecting the points to the ground plane~\cite{Philion2020ECCV}. An example BEV representation can be seen in \figref{fig:bev-pred-prob-def}. This leads to a compact representation where the future location and motion of multiple agents in the scene can be reliably predicted. In the last chapter, we propose a novel stochastic future prediction method for self-driving utilizing BEV representation.

\section{Proposed Methods and Models}

First, we present a new method for stochastic video future frame prediction. We propose to include the motion history in the scene in order to generate more realistic future frames. The motion in the scene was continuous, and by learning to model the history of the motion, we can propagate it to future frames. We named our model SLAMP, which stands for ``Stochastic Latent Appearance and Motion Prediction" because our model predicts future frames both in pixel space (\textbf{appearance}) and motion space (\textbf{motion}). 

Our major contribution is to exploit the existing motion history in the scene. With the explicit modeling of the motion history, our model predicts the motion between consecutive frames and predicts the future frames in the motion space. Besides in motion space, SLAMP predicts the future frames in pixel space. With these two predictions, SLAMP achieves a decomposition of the scene by attending pixel predictions for the static or occluded parts of the scene and motion predictions for the dynamic parts of the scene. SLAMP achieves state-of-the-art results on challenging real-world datasets with a dynamic background while performing on par with the previous methods on generic video prediction datasets such as KTH~\cite{Schuldt2004CVPR} and BAIR~\cite{EbertFLL17}. 

Second, we present another novel method for stochastic video future frame prediction, specifically for real-world autonomous driving scenarios. We propose to model static parts of the scene with ``Structure and Motion", \eg Ego-motion, instead of pixel space appearance prediction. We name our model as SLAMP-3D because of its 3D-aware static extension to SLAMP.

The first contribution of SLAMP-3D is that we decompose the scene into static and dynamic components similar to SLAMP. Instead of using pixel space for the static part, we propose to model the structure and the ego-motion together by exploiting the domain knowledge~\cite{Zhou2017CVPR, Godard2019ICCV, Safadoust2021THREEDV}. With this change in the static part, our work is much more suitable for the real-world driving datasets, KITTI~\cite{Geiger2012CVPR, Geiger2013IJRR} and Cityscapes~\cite{Cordts2016CVPR}. 

The camera motion in the scene can be explained by the structure and the ego-motion of the scene. If the scene is static, with no motion except the motion of the ego-vehicle, then camera motion can be explained perfectly with structure and ego-motion. However, in real-world scenarios, this assumption fails because there will be many independently moving objects whose motion cannot be explained by ego-motion. Therefore, as the second contribution in SLAMP-3D, we propose to condition the dynamic component on top of the static component to predict the residual flow, the remaining motion after the modeled ego-motion. As in the case of SLAMP, SLAMP-3D exploits the motion history and predicts the future frame in the motion space as an optical flow from the source frame to the target frame. Differently from SLAMP, the predicted optical flow is a \emph{residual flow} which remains in the scene due to independently moving objects. With this conditioning, SLAMP-3D achieves improved results, especially for the foreground objects in the dynamic scenes.

Third, we present a novel method for future instance segmentation method in Bird's-eye view from multiple cameras. We propose to formulate the prediction task as learning temporal dynamics through stochastic residual updates to a latent representation of the state at each time step. We name our model ``StretchBEV" because our method is much more capable than the existing ones (\emph{streched}) both temporally and spatially.

The main contribution is formulating the problem as learning temporal dynamics. We propose to use a state-space model with stochastic residual updates to a latent representation at each time step. With this formulation, the prediction problem suffices to a state prediction problem, which we handle with stochastic residual updates. With its strong latent variable, \emph{separated for each time step}, our methods can generate highly diverse future predictions while preserving the accuracy of the predictions. Our proposed approach clearly outperforms the state-of-the-art in various metrics evaluating the decoded predictions, especially by large margins in challenging cases of spatially far regions and temporally long spans. We also evaluate the diversity of the predictions both quantitatively and qualitatively and clearly show that our method outperforms the existing methods.

\section{Contributions and Novelties}

Our contributions are as follows:

\begin{itemize}
    \item We propose ``SLAMP", a novel method for stochastic video future frame prediction.
    \item We achieve comparable performance for generic video prediction datasets and outperform the existing ones in real-world scenarios.
    \item Our method not only predicts the future frame but also predicts the future optical flow without seeing the target frame.
    \item We propose ``SLAMP-3D", a novel 3D aware stochastic video prediction method that can predict future frames of the video using decomposition of 3D-aware scene structure and residual motion.
    \item Our method achieves comparable performance on real-world driving datasets while having a much more inference speed
    \item Our method requires ~1 second for generating 10 samples of a scene which is ~40 $\times$ faster than the state-of-the-art method without a significant loss of performance.
    \item We propose StretchBEV, a novel method for predicting future instances and their segmentation masks.
    \item We clearly outperform state-of-the-art methods, especially by large margins in challenging cases of spatially far regions and temporally long spans.
    \item Our method has the same inference speed as the existing ones while having much more performance and diversity.
\end{itemize}

\subsection{Publications}

\begin{itemize}
    \item Adil Kaan Akan, Fatma G\"{u}ney, ``StretchBEV: Stretching Future Instance Prediction Spatially and Temporally", Under Review
    \item Adil Kaan Akan, Sadra Safadoust, Erkut Erdem, Aykut Erdem, Fatma G\"{u}ney, ``Stochastic Video Prediction with Structure and Motion", Under Review
    \item Adil Kaan Akan, Sadra Safadoust, Erkut Erdem, Aykut Erdem, Fatma G\"{u}ney, ``SLAMP: Stochastic Latent Appearance and Motion Prediction", Proceedings of the IEEE/CVF International Conference on Computer Vision (ICCV), 2021.
\end{itemize}

\section{Outline of the Thesis}

Chapter \ref{chap:background} gives a brief background on autoregressive models and state-space models which we use in our proposed models.

Chapter \ref{chap:rw} summarizes the previous work done on the future prediction, specifically on video and Bird's-Eye view representation.

Chapter \ref{chapter:slamp} introduces the SLAMP model and its contribution, \emph{motion history}. We first discuss the existing work done in the area of stochastic video prediction. Then, we present our method and its details. In the end, we demonstrate detailed experiments and comparison to existing state-of-the-art methods in this chapter.

Chapter \ref{chapter:chap3} describes the SLAMP-3D and its decomposition strategy for 3D structure and residual motion. The details about its decomposition and conditioning are provided in this chapter. Moreover, we present detailed experiments and comparisons to existing work.

Chapter \ref{chap:chapter4} explains the StretchBEV and its novel formulation of temporal dynamics updates. We first discuss the previous work done in Bird's-eye view representation space, then we explain our formulation. In the end, we show experimental results that achieve new state-of-the-art in various metrics. 

Chapter \ref{chap:conc} concludes this thesis with a brief summary and presents possible directions for future work.

{\clearpage
\thispagestyle{plain}
~
\newpage}
\chapter{Background}
\label{chap:background}

In this chapter, we cover methods of modelling of future prediction, specifically autoregressive models and state-space models. We introduce the basic concepts following the previous work~\cite{Denton2018ICML, Franceschi2020ICML} since our proposed methods are building on top of them.

\section{Autoregressive Models}
\label{sec:bg-autoregressive}

For Autoregressive Models, we follow SVG~\cite{Denton2018ICML}, where the future information is encoded into latent variable using a learned prior and posterior distribution pair. 

\boldparagraph{SVG Architecture}

Denton and Fergus~\cite{Denton2018ICML} introduced SVG, an autoregressive model for stochastic video prediction. SVG encodes the future information into latent variables. They propose two different versions for their models, fixed-prior and learned-prior. In both versions, the posterior distribution is learned from the future time steps. However, the prior distribution changes depending on the version. In the fixed prior, the prior distribution is assumed to be a fixed standard Gaussian distribution. On the other hand, in the learned prior, the prior distribution is learnt from the previous and current time steps. They showed that learned prior outperforms the fixed one because the prior distribution has a flexibility to adapt according to the current time step. Below, we cover the steps of the algorithm.

SVG encodes each image into a feature space using CNNs. 

\begin{ceqn}
\begin{align}
\bh_{t-1} &= Enc(\bx_{t-1}) \\
\bh_t &= Enc(\bx_{t}) \nonumber
\end{align}
\end{ceqn}

where $\bx_{1:T}$ are the frames of the video, $Enc$ is CNN-based encoder and, $h_{1:T}$ are the encoded features of the corresponding video frame. Then, from the encoded features of current and future time steps, two separate distribution pairs, prior and posterior, are learnt. 

\begin{ceqn}
\begin{eqnarray}
\mu_{\psi}(t), \sigma_{\psi}(t) &= \mathrm{LSTM}_{\psi}(h_{t-1}) \\
\mu_{\phi}(t), \sigma_{\phi}(t) &= \mathrm{LSTM}_{\phi}(h_{t}) \nonumber
\end{eqnarray}
\end{ceqn}

where $\mu_{\psi}, \sigma_{\psi}$ are neural networks learning the prior distribution and $\mu_{\phi}, \sigma_{\phi}$ are neural networks learning the posterior distribution. After sampling the latent variable from the corresponding distribution, posterior in training and prior in the evaluation, the next frame's features are predicted by a separate model from combination of previous frame's features, $\bh_{t-1}$ and the latent variable, $\bz_t$. Note that in the fixed prior version, the prior distribution is not learnt. Therefore, the networks, $\mu_{\psi}, \sigma_{\psi}$, do not exist and the latent variable is sampled from $\mathcal{N}(0, 1)$.

\begin{ceqn}
\begin{equation}
    \bh_t = \mathrm{LSTM}_{\theta}(\bh_{t-1}, \bz_t)
\end{equation}
\end{ceqn}

After predicting the next frame's features, it is straightforward to decode it using CNNs into next frame. 

\begin{ceqn}
\begin{equation}
    \bx_t = Dec(\bh_t)
\end{equation}
\end{ceqn}

where Dec is a CNN-based decoder and decodes the features into video frames. 

The whole process is repeated for each time step, where the model  uses its predictions for the unseen steps. This means that the error accumulates throughout the generation process and towards to end, the generated video frames get blurry.

\section{State-Space Models}
\label{sec:bg-ssm}

As opposed to interleaving process of auto-regressive models, state-space models separate the frame generation from the modelling of dynamics \cite{Gregor2019ICLR}. State-of-the-art method SRVP \cite{Franceschi2020ICML} proposes a state-space model for video generation with deterministic state transitions representing residual change between the frames. In this chapter, we follow SRVP~\cite{Franceschi2020ICML}, where they consider the problem of stochastic video prediction as a distribution of possible future frames given the conditioning frames of a video, and briefly summarize the approach.

\boldparagraph{SRVP Architecture}

Franceschi~\etal~\cite{Franceschi2020ICML} introduced SRVP, a state-space model for stochastic video prediction model. SRVP encodes the scene into a stochastic latent variable and uses time-independent residual updates to create a different state representation for each time step. The SRVP separates the content and the dynamic by using a separate deterministic content network to encode the scene details and using it for the decoding of future frames. In this way, the latent state will learn the dynamics, \eg motion in the scene, and the content part will learn the details of the scene. 

In each time step, SRVP~\cite{Franceschi2020ICML} creates a state variable, $\by_t$, by using a stochastic residual network to update previous states into future ones. 

\begin{ceqn}
\begin{equation}
    \by_{t + \Delta t} = \by_{t} + \Delta t \cdot f_{\theta} \big (\by_{t}, \bz_{\lfloor t \rfloor + 1} \big ).
\end{equation}
\end{ceqn}

where $\by_{t}$ is the state representation at time $t$, $\Delta t$ is a hyperparameter that defines the size of the updates, $\bz$ is the latent variable sampled from prior or posterior distribution according to train or inference time. The prior distribution is learnt from current state, $\by_t$ whereas the posterior distribution is learnt from the future video frames, $\bx_{t+1:T}$.

\begin{ceqn}
\begin{equation}
    \bz_{t} \sim \mathcal{N}(\mu_{\theta}(\by_t), \sigma_{\theta}(\by_t)\mathrm{I})
\end{equation}
\end{ceqn}

where $\mu_{\theta}$ and $\sigma_{\theta}$ are the neural networks learning the prior distribution from the current state. 

As stated before, SRVP decompose the content and dynamics into separate spaces. SRVP learns a specific content variable, which is purely deterministic and stays the same throughout the video. That is, they assume the background is static and does not change. Since the content variable is available during the decoding of each frame, the network forces the latent variables to focus on the dynamics rather than the content of the video.

\begin{ceqn}
\begin{equation}
\label{eq:bg-content}
    \bw = \bc_{\psi}(\bx_1, \bx_2,\dots,\bx_K)
\end{equation}
\end{ceqn}

where $\bw$ is the content variable which stays the same throughout the video, $\bc_{\psi}$ is a neural network calculating the content variable, $\bx_1, \bx_2,\dots,\bx_K$ are the corresponding image frames.

After creating state representation for each time step and calculating the content variable, SRVP decodes all of them into video frames, \eg images, which can be seen in \eqref{eq:srvp_decode}.

\begin{ceqn}
\begin{equation}
    \label{eq:srvp_decode}
    \bx_{t} =  g_{\theta} \big (\by_{t}, \bw \big ).
\end{equation}
\end{ceqn}

where $\bx_t$ is the image frame at time $t$, $g_{\theta}$ is the decoder parameterized by $\theta$, $\by_t$ is the state representation at time $t$, $\bw$ is the content feature extracted by deterministic content network.

\section{Comparison between Autoregressive and State-Space Models}
\label{sec:bg-comp}

As we cover both models in this chapter, we briefly compare them in this section.

Autoregressive models use their predictions to propagate into the future. To predict further into the future, models need to use their predictions. This brings two drawbacks. First, the predictions are not 100\% correct. Each prediction has an error, even if the error is too small. Since the predictions are fed back into the models, the errors accumulate as the model rolls out into the future. Second, their run-time performance is slow compared to state-space models. Since the model needs to use its predictions further into the future, it first needs to generate the previous time steps. For example, to generate two steps into the future, the model needs to generate one step into the future. This sequential mechanism costs their run-time performance.

State-space models move the problem of prediction into a different state representation than the input representation, which is much smaller than the input most of the time. In this way, the prediction is made on the state-space, which is much faster due to the small dimensionality of the state-space. When all the states are predicted, the states are decoded into the input or output representation, which simply separates the prediction and generation. However, the problem with state-space models is that they assume the background static. For example, SRVP generates a deterministic content variable that stays the same throughout the video, see \eqref{eq:bg-content}. The assumption of static background holds for several datasets such as KTH~\cite{Schueldt2004CVPR} and BAIR~\cite{Ebert2017CORL}. However, in real-world scenarios, this assumption breaks. For example, in the driving scenarios, the background is not still, moving according to the ego-motion. As we show in Chapter~\ref{chapter:slamp}, SRVP performs poorly on two autonomous driving datasets with moving background.
{\clearpage
\thispagestyle{plain}
~
\newpage}
\chapter{Literature Review}
\label{chap:rw}

\section{Future Prediction in Video}
\label{sec:rw-video}

In this section, we cover the previous work focusing on video prediction. We first focus on the previous work specifically in \secref{sec:rw-svp}, then we cover the work done on video decomposition, and related works on 3D structure and motion in Sections~\ref{sec:rw-decomp} and \ref{sec:rw-3d}, respectively.

\subsection{Stochastic Video Prediction}
\label{sec:rw-svp}

SV2P~\cite{Babaeizadeh2018ICLR} and SVG~\cite{Denton2018ICML} are the first to model the stochasticity in video sequences using latent variables. The input from past frames are encoded in a posterior distribution to generate the future frames. In a stochastic framework, learning is performed by maximizing the likelihood of the observed data and minimizing the distance of the posterior distribution to a prior distribution, either fixed~\cite{Babaeizadeh2018ICLR} or learned from previous frames~\cite{Denton2018ICML}. Since time-variance is proven crucial by these previous works, we sample a latent variable at every time step. Sampled random variables are fed to a frame predictor, modelled recurrently using an LSTM. 
Typically, each distribution, including the prior and the posterior, is modeled with a recurrent model such as an LSTM.

There are various extensions to the basic formulation of stochastic video generation.
Villegas~\etal~\cite{Villegas2019NeurIPS} replace the linear LSTMs with convolutional ones at the cost of increasing the number of parameters. Castrejon~\etal~\cite{Castrejon2019ICCV} introduce a hierarchical representation to model latent variables at different scales which increases the complexity. Karapetyan~\etal~\cite{Karapetyan2022Arxiv} proposes usage of hierarchical recurrent networks to generate high quality video frames using multi-scale architectures.
Assouel~\etal~\cite{assouel2022CLEAR} introduce Variational Independent Modules to learn entity-centric representations and their transitions throughout the video. 
Babaeizadeh~\etal~\cite{Babaeizadeh2021Arxiv} over-parameterize an existing architecture to first overfit to the training set, and then use data augmentation to generalize to the validation or test sets.
Lee~\etal~\cite{Lee2018ARXIV} incorporate an adversarial loss into the stochastic framework to generate sharper images, at the cost of less diverse results.
We also use convolutional LSTMs to generate diverse and sharp-looking results without any adversarial losses by first reducing the spatial resolution to reduce the cost. Future prediction is typically performed in the pixel space but there are other representations such as keypoints \cite{villegas2017ICML, Minderer2019NeurIPS} , coordinates \cite{gu2021densetnt, gao2020vectornet} and bird’s-eye view \cite{Hu2021ICCV, Akan2022Arxiv}.

Optical flow has been used before in future prediction~\cite{Li2018ECCV, Lu2017CVPR}. Li \etal~\cite{Li2018ECCV} generate future frames from a still image by using optical flow generated by an off-the-shelf model, whereas we compute flow as part of prediction. Lu \etal~\cite{Lu2017CVPR} use optical flow for video extrapolation and interpolation without modeling stochasticity. Long-term video extrapolation results show the limitation of this work in terms of predicting future due to relatively small motion magnitudes considered in extrapolation. Differently from flow, Xue \etal~\cite{Xue2016NeurIPS} model the motion as image differences using cross convolutions. 

\subsection{Video Decomposition}
\label{sec:rw-decomp}

The previous work explored motion cues for video generation either explicitly with optical flow \cite{Walker2015ICCV, Walker2016ECCV, Liang2017ICCV, Liu2017ICCV, Lu2017CVPR, Fan2019AAAI, Gao2019ICCV} or implicitly with temporal differences \cite{Lotter2017ICLR} or pixel-level transformations \cite{Brabandere2016NeurIPS, Vondrick2017CVPR}. There are some common factors among these methods such as using recurrent models \cite{Shi2015NeurIPS, Lotter2017ICLR, Fan2019AAAI}, specific processing of dynamic parts \cite{Brabandere2016NeurIPS, Liang2017ICCV, Fan2019AAAI, Gao2019ICCV}, utilizing a mask \cite{Finn2016NeurIPS, Liu2017ICCV, Gao2019ICCV}, and adversarial training \cite{Vondrick2017CVPR, Lu2017CVPR}. In our models, we also use recurrent models, predict a mask, and separately process motion, but in a stochastic way.

Specifically, state-space model SRVP~\cite{Franceschi2020ICML} learns a content variable from the first few frames which remains unchanged while predicting the future frames.
As we will show in the next chapter, Chapter~\ref{chapter:slamp}, the content variable in SRVP cannot handle changes in the background. In addition to pixel space, SLAMP separately models changes in motion and keeps track of a motion history. This reduces the role of the pixel decoder to recover occlusions, \eg around motion boundaries. 
We propose to decompose the scene as static and dynamic where the static part not only considers the ego-motion but also the structure. This allows us to differentiate the motion in the background from the motion in the foreground which leads to a more meaningful scene decomposition for driving.
Disentanglement of explicit groups of factors has been explored before for generating 3D body models depending on the body pose and the shape~\cite{Detlefsen2019NeurIPS}. Inspired by the conditioning of the body pose on the shape, in SLAMP-3D, we condition the motion of the dynamic part on the static part to achieve a disentanglement between the two types of motion in driving.

\subsection{Structure and Motion}
\label{sec:rw-3d}

Our approach, SLAMP-3D is related to view synthesis~\cite{Garg2016ECCV} where the target view can be synthesized by warping a source view based on the depth estimation from the target view~\cite{Jaderberg2015NeurIPS}. The difference between the synthesized target view and the original one can be used for self-supervised training.
Monocular depth estimation approaches~\cite{Zhou2017CVPR, Zhan2018CVPR, Wang2018CVPR, Bian2019NeurIPS, Mahjourian2018CVPR, Godard2019ICCV, Guizilini2020CVPR} generalize view synthesis to adjacent frames by also estimating the relative pose from one frame to the next. These methods can successfully model the structure and motion in the static part of the scene. However, the motion of independently moving objects remains as a source of error.
More recent works~\cite{Yin2018CVPR, Zou2018ECCV, Chen2019ICCV, Ranjan2019CVPR, Luo2019PAMI} estimate the residual optical flow to model the motion of independently moving objects. In addition, some of these methods model the consistency between depth and optical flow~\cite{Zou2018ECCV}, and also motion segmentation~\cite{Ranjan2019CVPR, Luo2019PAMI, Safadoust2021THREEDV}. In SLAMP-3D, we similarly learn the decomposition of the world into static and dynamic parts but in longer sequences by exploiting the history from previous frames to predict future frames with uncertainty.

\section{Future Prediction in Bird's-Eye View}

In this section, we first compare the autoregressive models and state-space models in \secref{sec:rw-comp}. Then, we focus on the future prediction in driving scenarios in \secref{sec:rw-driving}.

Our work, StretchBEV introduced in Chapter~\ref{chap:chapter4}, fits into the stochastic future prediction framework, producing long term, diverse predictions, however, we predict future in the Bird's-Eye view (BEV) space instead of the noisy pixel space.

\subsection{Autoregressive Models vs. State-Space Models}
\label{sec:rw-comp}

As introduced in \secref{sec:bg-comp}, autoregressive method have several problems such as error accumulation and low runtime speed. The performance of auto-regressive methods can be improved by increasing the network capacity~\cite{Villegas2019NeurIPS} or introducing a hierarchy into the latent variables~\cite{Castrejon2019ICCV}, which also increase the complexity of these methods. Due to complexity of predicting pixels, another group of work moves away from the pixel space to the keypoints~\cite{Minderer2019NeurIPS} or to the motion space by incorporating motion history as we propose in both SLAMP and SLAMP-3D. Our proposed approach follows a similar strategy by performing future prediction in the BEV representation, but more efficiently by avoiding auto-regressive predictions.

Auto-regressive strategy requires encoding the predictions, leading to high computational cost and creates a tight coupling between the temporal dynamics and the generation process~\cite{Gregor2019ICLR,Rubanova2019NeurIPS}. 

On the other hand, the state-space models~(SSM) break this coupling by separating the learning of dynamics from the generation process, resulting in a computationally more efficient approach. Furthermore, learned states can be directly used in reinforcement learning~\cite{Gregor2019ICLR} and can be interpreted~\cite{Rubanova2019NeurIPS}, making SSMs particularly appealing for self-driving. Earlier SSMs with variational deep inference suffer from complicated inference schemes and typically target low-dimensional data~\cite{Krishnan2017AAAI,Karl2017ICLR}. An efficient training strategy with a temporal model based on residual updates is proposed for high-dimensional video prediction in the state of the art SSM~\cite{Franceschi2020ICML}. We adapt a similar residual update strategy for predicting future BEV representations. We also experimentally show that the content variable for the static part of the scene is not as effective in the BEV space as it is in pixel space~\cite{Franceschi2020ICML}. 

\subsection{Future Prediction in Driving Scenarios}
\label{sec:rw-driving}

The typical approach to the prediction problem in self-driving is to first perform detection and tracking, and then do the trajectory prediction~\cite{Chai2020CORL,Hong2019CVPR}. In these methods, errors are propagated at each step. There are some recent methods~\cite{Luo2018CVPR,Casas2018CORL,Casas2020ICRA,Djuric2021IV} which directly address the prediction problem from the sensory input including LiDAR, radar, and other sensors. These methods also typically rely on an HD map of the environment. Due to high performance and efficiency of end-to-end approach, we follow a similar approach for future prediction but using cameras only and without relying on HD maps.

Despite their efficiency, most of the previous work focus on the most likely prediction~\cite{Casas2018CORL} or only models the uncertainty regarding the ego-vehicle's trajectory~\cite{Casas2020ICRA,Djuric2021IV}. The motion prediction methods which consider the behavior of all the agents in the scene typically assume a top-down rasterised representation as input, \eg Argoverse setting~\cite{Chang2019CVPR}. Even then, multiple future prediction problem is typically addressed by generating a fixed number of predictions~\cite{Gao2020CVPR,Liang2020ECCV}, for example by estimating the likelihood of multiple target locations~\cite{Zhao2020CORL,Gu2021ICCV}. There are some exceptions~\cite{Tang2019NeurIPS,Sriram2020ECCV,Huang2020RAL} which directly address the diversity aspect with a probabilistic framework. These works, especially the ones using a deep variational framework~\cite{Tang2019NeurIPS,Sriram2020ECCV} are similar to our approach in spirit, however, they operate in the coordinate space by assuming the availability of a top-down map where locations of agents are marked. In this thesis, we aim to learn this top-down BEV representation from multiple cameras by also segmenting the agents in the scene.

FIERY~\cite{Hu2021ICCV} is the first to address probabilistic future prediction from multiple cameras. However, future predictions are limited both in terms of diversity and length considering the typical video prediction setting. In Chapter~\ref{chap:chapter4}, we propose a probabilistic future prediction method that can generate diverse predictions extending to different temporal horizons with a stochastic temporal dynamics model.

{\clearpage
\thispagestyle{plain}
~
\newpage}
\chapter{SLAMP: Stochastic Latent Appearance and Motion Prediction}
\label{chapter:slamp}

\section{Introduction}
\label{sec:chap2-intro}

Motion is an important clue in the videos. The temporal connection between the frames is crucial for improving the performance of any task in video representation. Most of the time, the motion in the video stays the same or does not change much throughout the video duration. For example, if a person walks at the beginning of a video, then most probably, the person will continue walking. With this insight, we introduce the concept of motion history. The motion history corresponds to explicitly modeling the past motion in the video and propagating it to the future of the video. In this way, the future frames are generated using the motion information in the video.

In this chapter, we introduce SLAMP, which is a stochastic future prediction method for videos. In SLAMP, we exploit the \emph{motion history} concept and generate the future frames using the motion information. 

Our method is summarized as follows: We first sample latent variables from two separate distributions, one for appearance and one for motion, which is learnt from the images. The latent variables contain the information from the future time steps, similar to SVG~\cite{Denton2018ICML}. After, we simultaneously generate the future target frame in both the pixel space and the motion space using the sampled latent variables. The pixel space corresponds to that generating the target frame as an image. On the other hand, the motion space corresponds to generating the motion from the current frame to the target frame, then synthesizing the target frame. So, SLAMP predicts the future frame in two different modalities. In the end, a mask is predicted to combine both predictions in the best way. 

\begin{figure}[h!]
\centering
\includegraphics[width=1\linewidth]{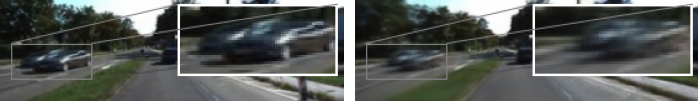}
   \includegraphics[width=1\linewidth]{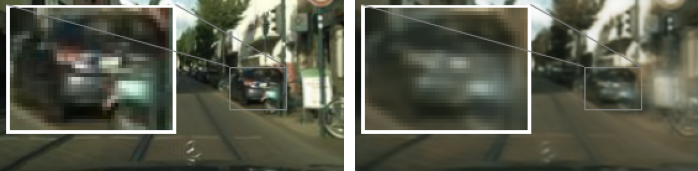}
\caption{Comparison of the first prediction frames (11th) SLAMP~(\textbf{left}) \vs state-of-the-art method, SRVP~\cite{Franceschi2020ICML}~(\textbf{right}) on KITTI~\cite{Geiger2013IJRR}~(\textbf{top}) and Cityscapes~\cite{Cordts2016CVPR}~(\textbf{bottom}) datasets. Our method can predict both foreground and background objects better than SRVP. Full sequence predictions can be seen in Appendix.}
\label{fig:teaser-slamp}
\end{figure}

Using the motion history in a stochastic way boosts performance in challenging real-world driving scenarios with a moving background. One example case can be seen in \figref{fig:teaser-slamp}. In the high-speed scenarios, our model can preserve the shape of the car (top example in \figref{fig:teaser-slamp}), and in dense environments, our model can conserve the details of the car, such as the license plate.

In this chapter, we first explain our approach in detail in \secref{sec:chap2-method}, then we validate our claims on motion history with the experiments presented in \secref{sec:chap2-exp}.

\section{Methodology}
\label{sec:chap2-method}

\begin{figure}[ht!]
\centering
\includegraphics[width=\textwidth]{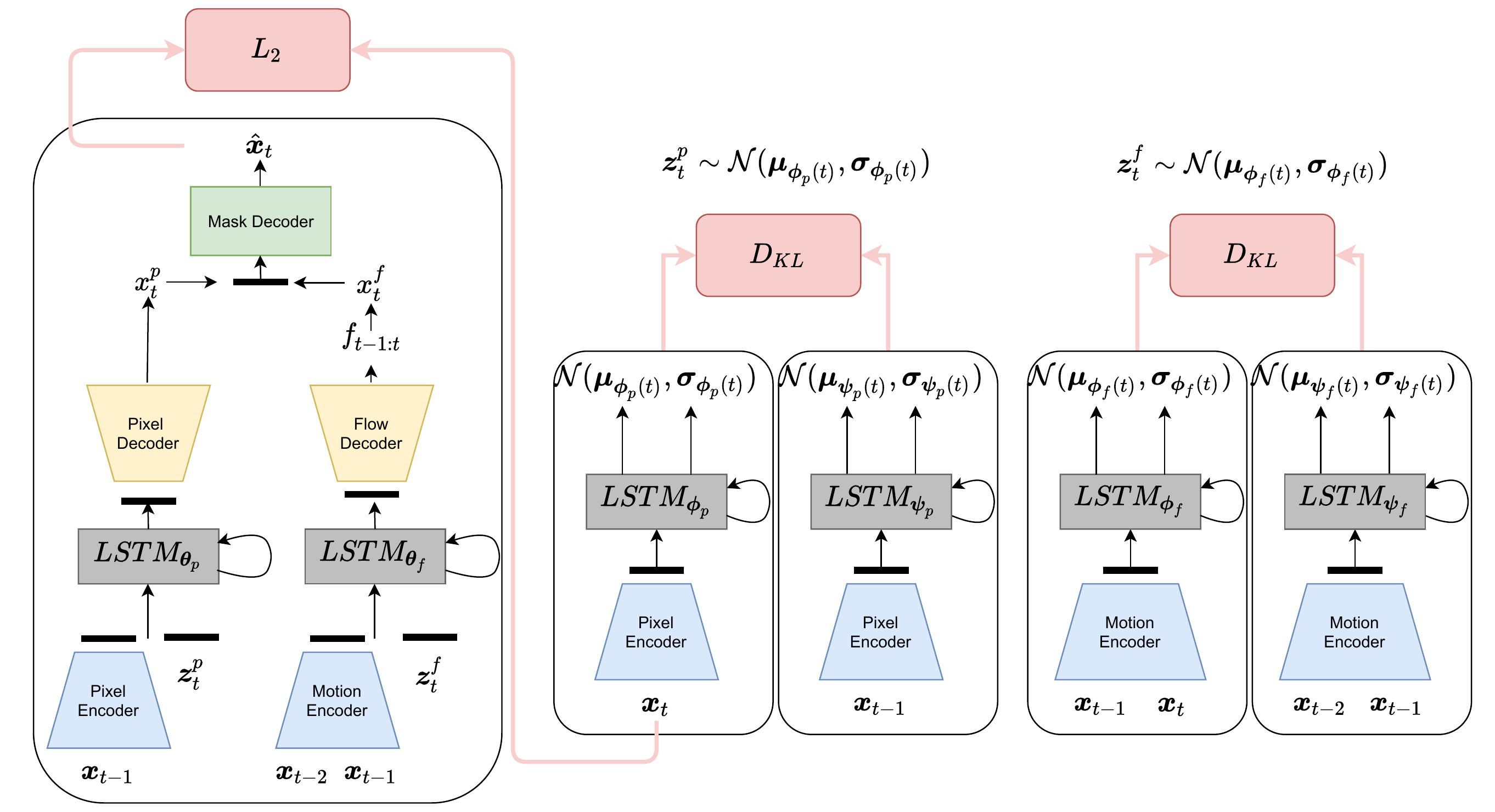}
\caption[SLAMP Architecture]{\textbf{SLAMP.} This figure shows the components of our SLAMP model including the prediction model, inference and learned prior models for pixel and then flow from left to right. Observations $\bx_t$ are mapped to the latent space by using a pixel encoder for appearance on each frame and and a motion encoder for motion between consecutive frames. The blue boxes show encoders, yellow and green ones decoders, gray ones recurrent posterior, prior, and predictor models, and lastly red ones show loss functions during training. Note that $L_2$ loss is applied three times for appearance prediction $\bx_t^p$, motion prediction $\bx_t^f$, and the combination of the two $\hat{\bx}_t$ according to the mask prediction $\bm(\bx_t^p,\bx_t^f)$. We only show L2 loss between the actual frame $\bx_t$ and the final predicted frame $\hat{\bx}_{t}$ in the figure. For inference, only the prediction model and learned prior models are used.}
\label{fig:model_motion_prior}
\end{figure}

\subsection{Stochastic Video Prediction}
\label{sec:svg}
Given the previous frames $\bx_{1:t-1}$ until time $t$, our goal is to predict the target frame $\bx_t$. For that purpose, we assume that we have access to the target frame $\bx_t$ during training and use it to capture the dynamics of the video in stochastic latent variables $\bz_t$. By learning to approximate the distribution over $\bz_t$, we can decode the future frame $\bx_t$ from $\bz_t$ and the previous frames $\bx_{1:t-1}$ at test time.

Using all the frames including the target frame, we compute a posterior distribution $q_{\bphi}(\bz_t \vert \bx_{1:t})$ and sample a latent variable $\bz_t$ from this distribution at each time step. 
The stochastic process of the video is captured by the latent variable $\bz_t$. In other words, it should contain information accumulated over the previous frames rather than only condensing the information on the current frame. This is achieved by encouraging $q_{\bphi}(\bz_t \vert \bx_{1:t})$ to be close to a prior distribution $p(\bz)$ in terms of KL-divergence. 
The prior can be sampled from a fixed Gaussian 

at each time step or can be learned from previous frames up to the target frame $p_{\bpsi}(\bz_t \vert \bx_{1:t-1})$. We prefer the latter one as it is shown to work better by learning a prior that varies across time \cite{Denton2018ICML}.

The target frame $\bx_t$ is predicted based on the previous frames $\bx_{1:t-1}$ and the latent vectors $\bz_{1:t}$.

In practice, we only use the latest frame $\bx_{t-1}$ and the latent vector $\bz_t$ as input and dependencies from further previous frames are propagated with a recurrent model. The output of the frame predictor $\bg_t$
 
contains the information required to decode $\bx_t$. 

Typically, $\bg_t$ is decoded to a fixed-variance Gaussian distribution whose mean is the predicted target~frame $\hat{\bx}_t$~\cite{Denton2018ICML}.

\subsection{SLAMP}
\label{sec:slamp}

We call the predicted target frame, appearance prediction $\bx_t^p$ in the pixel space.
In addition to $\bx_t^p$, we also estimate optical flow $\bff_{t-1:t}$ from the previous frame $t-1$ to the target frame $t$. The flow $\bff_{t-1:t}$ represents the motion of the pixels from the previous frame to the target frame. We reconstruct the target frame $\bx_t^f$ from the estimated optical flow via differentiable warping \cite{Jaderberg2015NeurIPS}. Finally, we estimate a mask $\bm(\bx_t^p,\bx_t^f)$ from the two frame estimations to combine them into the final estimation $\hat{\bx}_t$:

\begin{ceqn}
\begin{align}
    \label{eq:combined_x}
   \hat{\bx}_t = \bm(\bx_t^p,\bx_t^f) \odot \bx_t^p + (\mathbf{1} - \bm(\bx_t^p,\bx_t^f)) \odot \bx_t^f
\end{align}
\end{ceqn}
where $\odot$ denotes element-wise Hadamard product and $\bx_t^f$ is the result of warping the source frame to the target frame according to the estimated flow field $\bff_{t-1:t}$.
Especially in the dynamic parts with moving objects, the target frame can be reconstructed accurately using motion information. In the occluded regions where motion is unreliable, the model learns to rely on the appearance prediction. The mask prediction learns a weighting between the appearance and the motion predictions for combining them. 

We call this model SLAMP-Baseline because it is limited in the sense that it only considers the motion with respect to the previous frame while decoding the output. In SLAMP, we extend the stochasticity in the appearance space to the motion space as well. This way, we can model appearance changes and motion patterns in the video explicitly and make better predictions of future. \figref{fig:model_motion_prior} shows an illustration of SLAMP (see Appendix for \mbox{SLAMP-Baseline}).

In order to represent appearance and motion, we compute two separate posterior distributions $q_{\bphi_p}(\bz_t^p \vert \bx_{1:t})$ and $q_{\bphi_f}(\bz_t^f \vert \bx_{1:t})$, respectively. We sample two latent variables $\bz_t^p$ and $\bz_t^f$ from these distributions in the pixel space and the flow space. This allows a decomposition of the video into static and dynamic components. Intuitively, we expect the dynamic component to focus on changes and the static to what remains constant from the previous frames to the target frame. 
If the background is moving according to a camera motion, the static component can model the change in the background assuming that it remains constant throughout the video, \eg ego-motion of a car.

\boldparagraph{The Motion History} The latent variable $\bz_t^f$ should contain motion information accumulated over the previous frames rather than local temporal changes between the last frame and the target frame. We achieve this by encouraging $q_{\bphi_f}(\bz_t^f \vert \bx_{1:t})$ to be close to a prior distribution in terms of KL-divergence. 
Similar to \cite{Denton2018ICML}, we learn the motion prior conditioned on previous frames up to the target frame: $p_{\bpsi_f}(\bz_t^f \vert \bx_{1:t-1})$. We repeat the same for the static part represented by $\bz_t^p$ with posterior $q_{\bphi_p}(\bz_t^p \vert \bx_{1:t})$
and the learned prior $p_{\bpsi_p}(\bz_t^p \vert \bx_{1:t-1})$.

\begin{figure}[ht]
\centering
\includegraphics[width=\textwidth]{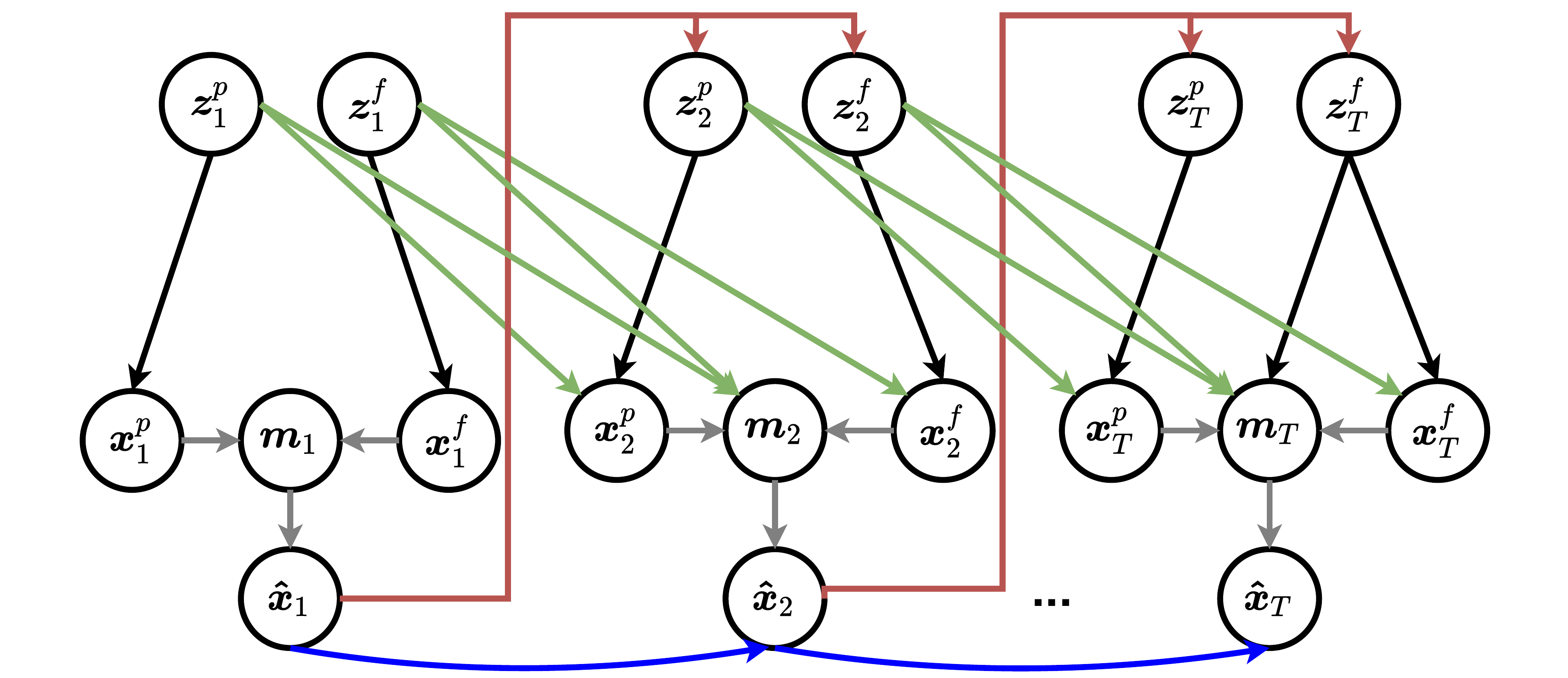}
\caption[Generative Model of SLAMP]{\textbf{Generative Model of SLAMP.} The graphical model shows the generation process of SLAMP with motion history. There are two separate latent variables for appearance $\bz^p_{t}$ and motion $\bz^f_{t}$ generating frames $\bx^p_{t}$ and $\bx^f_{t}$ (black).
Information is propagated between time-steps through the recurrence between frame predictions (\textcolor{blue}{blue}), corresponding latent variables (\textcolor{LimeGreen}{green}), and from frame predictions to latent variables (\textcolor{OrangeRed}{red}).
The final prediction $\hat{\bx}_{t}$ is a weighted combination of the $\bx^p_{t}$ and $\bx^f_{t}$ according to the mask $\bm(\bx^p_{t},\bx^f_{t})$.
Note that predictions at a time-step depend on all of the previous time-steps recurrently, but only the connections between consecutive ones are shown for clarity.}
\label{fig:graphical_model}
\end{figure}

\subsection{Variational Inference}
\label{sec:inf}
For our basic formulation (SLAMP-Baseline), the derivation of the loss function is straightforward and provided in Appendix. For SLAMP, the conditional joint probability corresponding to the graphical model in \figref{fig:graphical_model} is:

\begin{ceqn}
\begin{align}
    p(\bx_{1:T}) = \prod_{t=1}^{T} &~~p(\bx_t \lvert \bx_{1:t-1}, \bz_t^p, \bz_t^f)  \\
    &~~p(\bz_{t}^{p} \lvert \bx_{1:t-1}, \bz_{t-1}^{p})
    ~~p(\bz_{t}^{f} \lvert \bx_{1:t-1}, \bz_{t-1}^{f}) \nonumber
    \vspace{-3mm}
\end{align}
\end{ceqn}
The true distribution over the latent variables $\bz_{t}^{p}$ and $\bz_{t}^{f}$ is intractable. We train time-dependent inference networks $q_{\bphi_p}(\bz_{t}^{p} \lvert \bx_{1:T})$ and $q_{\bphi_f}(\bz_{t}^{f} \lvert \bx_{1:T})$ to approximate the true distribution with conditional Gaussian distributions.
In order to optimize the likelihood of $p(\bx_{1:T})$, we need to infer latent variables $\bz_{t}^{p}$ and $\bz_{t}^{f}$, which correspond to uncertainty of static and dynamic parts in future frames, respectively. We use a variational inference model to infer the latent variables.

Since $\bz_{t}^{p}$ and $\bz_{t}^{f}$ are independent across time, we can decompose Kullback-Leibler terms into individual time steps. We train the model by optimizing the variational lower bound (see Appendix for the derivation):

\begin{align}
    \label{eq:elbo}
    & \mathrm{log}\ p_{\btheta}(\bx) \geq \mathcal{L}_{\btheta,\bphi_{p}, \bphi_{f},\bpsi_{p},\bpsi_{f}}
    (\bx_{1:T}) \\
    & ={} 
    \begin{aligned}[t]
        \sum\limits_{t}
            \mathbbm{E}_{\substack{\bz_{1:t}^p \sim q_{\bphi_p} \\
                                  \bz_{1:t}^f \sim q_{\bphi_f}}}
                &~\mathrm{log}\ p_{\btheta}(\bx_t \lvert \bx_{1:t-1}, \bz_{1:t}^{p}, \bz_{1:t}^{f}) \nonumber \\
                & - \beta \Big[ D_{\mathrm{KL}}\big(q(\bz^{p}_t \lvert \bx_{1:t}) \mid \mid  p(\bz^{p}_t \lvert \bx_{1:t-1})  \big) +  D_{\mathrm{KL}}\big(q(\bz^{f}_t \lvert \bx_{1:t}) \mid \mid p(\bz^{f}_t \lvert \bx_{1:t-1}) \big)  \Big] \nonumber
    \end{aligned}
\end{align}

The likelihood $p_{\btheta}$, can be interpreted as an $L_2$ penalty between the actual frame $\bx_t$ and the estimation $\hat{\bx}_t$ as defined in \eqref{eq:combined_x}. We apply the $L_2$ loss to the predictions of appearance and motion components as well.

The posterior terms for uncertainty are estimated as an expectation over $q_{\bphi_p}(\mathbf{z}_t^p \lvert \mathbf{x}_{1:t})$, $q_{\bphi_f}(\mathbf{z}_t^f \lvert \mathbf{x}_{1:t})$.
As in \cite{Denton2018ICML}, we also learn the prior distributions from the previous frames up to the target frame as 
$p_{\bpsi_p}(\mathbf{z}_t^p \lvert \mathbf{x}_{1:t-1})$, $p_{\bpsi_f}(\mathbf{z}_t^f \lvert \mathbf{x}_{1:t-1})$.
We train the model using the re-parameterization trick \cite{Kingma2014ICLR}.
We classically choose the posteriors to be factorized Gaussian so that all the KL divergences can be computed analytically.

\subsection{Architecture}
\label{sec:arch}
We encode the frames with a feed-forward convolutional architecture to obtain appearance features at each time-step. In SLAMP, we also encode consecutive frame pairs into a feature vector representing the motion between them. 
We then train linear LSTMs to infer posterior and prior distributions 
at each time-step from encoded appearance and motion features.

Stochastic video prediction model with a learned prior \cite{Denton2018ICML} is a special case of our baseline model with a single pixel decoder, we also add motion and mask decoders.
Next, we describe the steps of the generation process for the dynamic part.

At each time step, we encode $\bx_{t-1}$ and $\bx_{t}$ into $\bh_{t}^f$, representing the motion from the previous frame to the target frame.
The posterior LSTM is updated based on the~$\bh_{t}^f$:

\begin{ceqn}
\begin{eqnarray}
    \bh_t^f &=& \mathrm{MotionEnc}(\bx_{t-1}, \bx_t) \\
    \bmu_{\bphi_f(t)}, \bsigma_{\bphi_f(t)} &=& \mathrm{LSTM}_{\bphi_f}(\bh_t^f) \nonumber
    \vspace{-5mm}
\end{eqnarray}
\end{ceqn}
For the prior, we use the motion representation $\bh_{t-1}^f$ from the previous time step, \ie the motion from the frame $t-2$ to the frame $t-1$, to update the prior LSTM: 
\begin{ceqn}
\begin{eqnarray}
    \bh_{t-1}^f &=& \mathrm{MotionEnc}(\bx_{t-2}, \bx_{t-1}) \\
    \bmu_{\bpsi_f(t)}, \bsigma_{\bpsi_f(t)} &=& \mathrm{LSTM}_{\bpsi_f}(\bh_{t-1}^f) \nonumber
    \vspace{-5mm}
\end{eqnarray}
\end{ceqn}
At the first time-step where there is no previous motion, we assume zero-motion by estimating the motion from the previous frame to itself.

The predictor LSTMs are updated according to encoded features and sampled latent variables:
\begin{ceqn}
\begin{eqnarray}
    \bg_t^f &=& \mathrm{LSTM}_{\btheta_f}(\bh_{t-1}^f, \bz_t^f) \\
    \bmu_{\btheta_f} &=& \mathrm{FlowDec}(\bg_t^f) \nonumber
\end{eqnarray}
\end{ceqn}
There is a difference between the train time and inference time in terms of the distribution the latent variables are sampled from. At train time, latent variables are sampled from the posterior distribution. At test time, they are sampled from the posterior for the conditioning frames and from the prior for the following frames.
The output of the predictor LSTMs are decoded into appearance and motion predictions separately and combined into the final prediction using the mask prediction (\eqnref{eq:combined_x}).

\section{Experiments}
\label{sec:chap2-exp}

\begin{figure}[ht]
    \centering
    \includegraphics[width=0.7\textwidth]{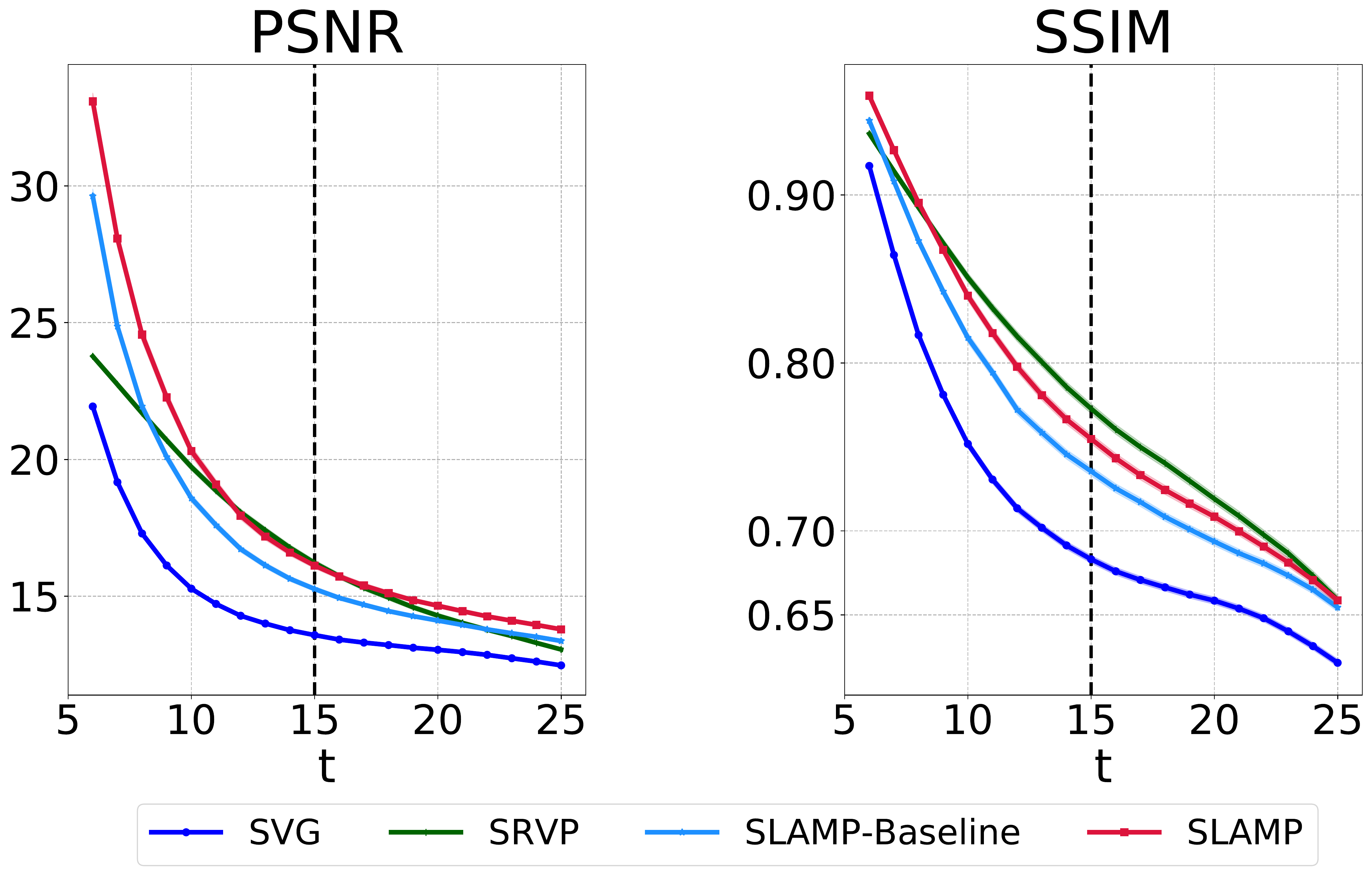}
    \caption[Quantitative Results on MNIST]{\textbf{Quantitative Results on MNIST.} This figure compares SLAMP to SLAMP-Baseline, SVG~\cite{Denton2018ICML}, and SRVP~\cite{Franceschi2020ICML} on MNIST in terms of PSNR~(\textbf{left}) and SSIM~(\textbf{right}). SLAMP clearly outperforms our baseline model and SVG, and performs comparably to SRVP. Vertical bars mark the length of the training sequences.}
    \label{fig:results_mnist}
    \vspace{-3.3mm}
\end{figure}
\begin{figure*}[ht]
    \centering
    \includegraphics[width=\textwidth]{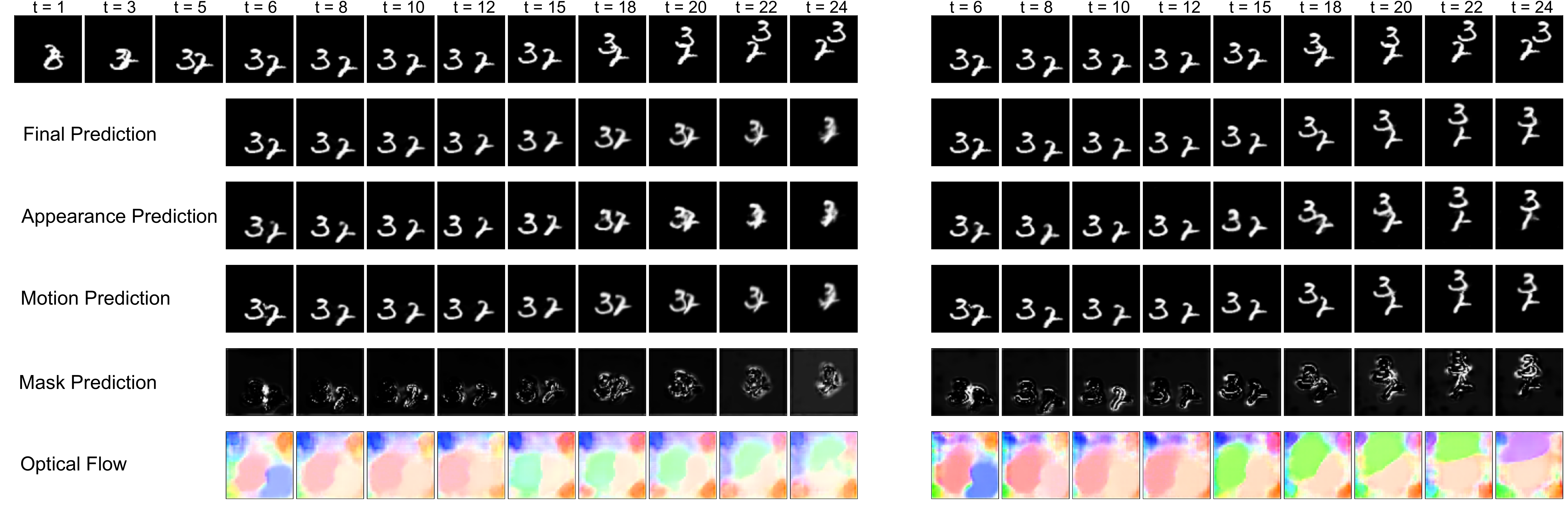}
    \caption[Comparison of SLAMP-Baseline and SLAMP]{\textbf{SLAMP-Baseline (left) \vs SLAMP (right) on MNIST.} The top row shows the ground truth, followed by the frame predictions by the final, the appearance, the motion, and the last two rows show the mask and the optical flow predictions with false coloring. In this challenging case with bouncing and collisions, the baseline confuses the digits and cannot predict last frames correctly whereas SLAMP can generate predictions very close to the ground truth by learning smooth transitions in the motion history, as can be seen from optical flow predictions.}
    \label{fig:qual_mnist_comp}
    \vspace{-5mm}
\end{figure*}
We evaluate the performance of the proposed approach and compare it to the previous methods on three standard video prediction datasets including Stochastic Moving MNIST, KTH Actions \cite{Schuldt2004CVPR} and BAIR Robot Hand \cite{EbertFLL17}. 
We specifically compare our baseline model (SLAMP-Baseline) and our model (SLAMP) to SVG \cite{Denton2018ICML} which is a special case of our baseline with a single pixel decoder, SAVP \cite{Lee2018ARXIV}, SV2P \cite{Babaeizadeh2018ICLR}, and lastly to SRVP \cite{Franceschi2020ICML}. 
We also compare our model to SVG \cite{Denton2018ICML} and SRVP \cite{Franceschi2020ICML} on two different challenging real world datasets, KITTI \cite{Geiger2012CVPR, Geiger2013IJRR} and Cityscapes \cite{Cordts2016CVPR}, with moving background and complex object motion.
We follow the evaluation setting introduced in \cite{Denton2018ICML} 
by generating 100 samples for each test sequence and report the results according to the best one in terms of average performance over the frames.
Our experimental setup including training details and parameter settings can be found in Appendix. We also share the code for reproducibility.

\begin{figure}[ht]
    \centering
    \includegraphics[width=\linewidth]{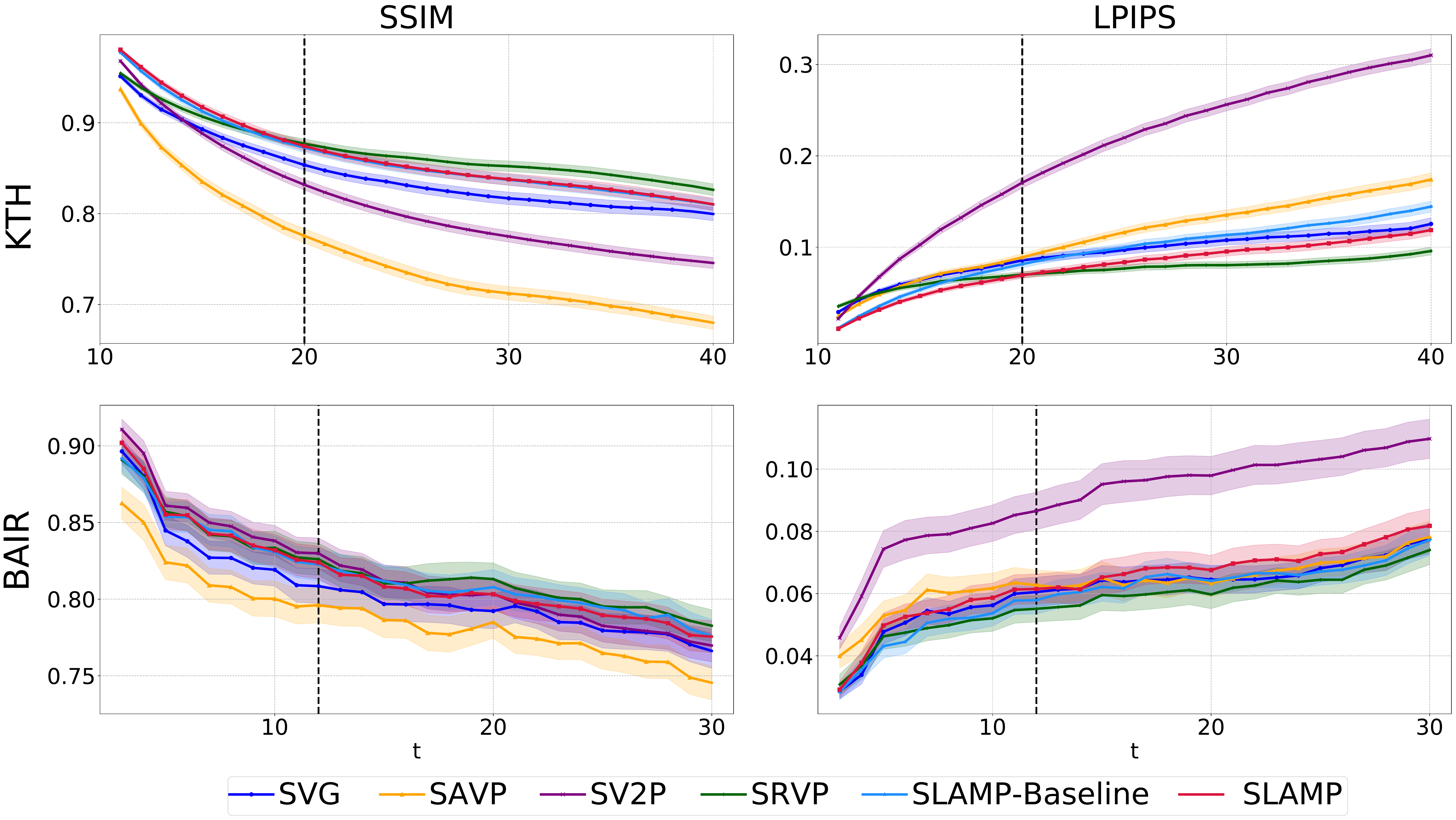}
    \caption[Quantitative Results on KTH and BAIR]{\textbf{Quantitative Results on KTH and BAIR.} We compare our results to previous work in terms of PSNR, SSIM, and LPIPS metrics with respect to the time steps on KTH (\textbf{top}), and BAIR (\textbf{bottom}) datasets, with 95\%-confidence intervals. Vertical bars mark the length of training sequences. SLAMP outperforms previous work including SVG \cite{Denton2018ICML}, SAVP \cite{Lee2018ARXIV}, SV2P \cite{Babaeizadeh2018ICLR} and performs comparably to the state of the art method SRVP \cite{Franceschi2020ICML} on both datasets.}
    \label{fig:results_kth_bair}
    \vspace{-4mm}
\end{figure}

\begin{table*}[h!]
    \captionsetup[table]{justification=centering}
    \sisetup{detect-weight, table-align-uncertainty=true, mode=text}
    \renewrobustcmd{\bfseries}{\fontseries{b}\selectfont}
    \renewrobustcmd{\boldmath}{}
    \caption[SLAMP's FVD Scores on the KTH and BAIR]{
        \label{tab:fvd}
        \textbf{FVD Scores on the KTH and BAIR.}This table compares all the methods in terms of FVD scores with their $95\%$-confidence intervals over five different samples from the models. Our model is the second best on KTH and among top three methods on BAIR.}
    \centering
    \vspace{-0.05in}
    \small
    \begin{tabular}{lccccccc}
        \toprule
        Dataset & {SV2P} & {SAVP} & {SVG} & {SRVP} & {SLAMP - Baseline} & {SLAMP} \tabularnewline
        \midrule
        KTH & 636 $\pm$ 1 & 374 $\pm$ 3 & 377 $\pm$ 6 & \bfseries 222 $\pm$ 3 & 236 $\pm$ 2 & \underline{228} $\pm$ 5 \tabularnewline
        BAIR & 965 $\pm$ 17 & \bfseries 152 $\pm$ 9 & 255 $\pm$ 4 & \underline{163} $\pm$ 4 & 245 $\pm$ 5 & {\textemdash} \tabularnewline
        \bottomrule
    \end{tabular}
    \vspace{-2mm}
\end{table*}

\boldparagraph{Evaluation Metrics} We compare the performance using three frame-wise metrics and a video-level one.
Peak Signal-to-Noise Ratio (PSNR), \emph{higher better}, based on $L_2$ distance between the frames penalizes differences in dynamics but also favors blur predictions.
Structured Similarity (SSIM), \emph{higher better}, compares local patches to measure similarity in structure spatially. %
Learned Perceptual Image Patch Similarity (LPIPS) \cite{Zhang2018CVPR}, \emph{lower better}, measures the distance between learned features extracted by a CNN trained for image classification.
Frechet Video Distance (FVD) \cite{Unterthiner2019ARXIV}, \emph{lower better}, compares temporal dynamics of generated videos to the ground truth in terms of representations computed for action recognition.

\boldparagraph{Stochastic Moving MNIST} This dataset contains up to two MNIST digits moving linearly and bouncing from walls with a random velocity as introduced in \cite{Denton2018ICML}.
Following the same training and evaluation settings as in the previous work, we condition on the first 5 frames during training and learn to predict the next 10 frames. During testing, we again condition on the first 5 frames but predict the next 20 frames.

\figref{fig:results_mnist} shows quantitative results on MNIST in comparison to SVG \cite{Denton2018ICML} and SRVP \cite{Franceschi2020ICML} in terms of PSNR and SSIM, omitting LPIPS as in SRVP. Our baseline model with a motion decoder (SLAMP-Baseline) already outperforms SVG on both metrics. SLAMP further improves the results by utilizing the motion history and reaches a comparable performance to the state of the art model SRVP. This shows the benefit of separating the video into static and dynamic parts in both state-space models (SRVP) and auto-regressive models (ours, SLAMP). This way, models can better handle challenging cases such as crossing digits as shown next.

\begin{table}[h!]
    \caption[SLAMP's Results on MNIST]{
        \label{tab:res-mnist}
        \textbf{Results on MNIST.} This table compares the results of SLAMP and SLAMP-Baseline to the previous work on MNIST dataset.
        Following the previous work, we report the results as the mean and the $95\%$-confidence interval in terms of PSNR and SSIM.
        Bold and underlined scores indicate the best and the second best performing method, respectively.
    }
    \sisetup{detect-weight, table-align-uncertainty=true, mode=text}
    \renewrobustcmd{\bfseries}{\fontseries{b}\selectfont}
    \renewrobustcmd{\boldmath}{}
    \centering
    \vspace{0.1in}
    \begin{tabular}{lcc}
        \toprule
        Models & {PSNR} & {SSIM} \tabularnewline
        \midrule
        SVG \cite{Denton2018ICML} & 14.50 $\pm$ 0.04 & 0.7090 $\pm$ 0.0015 \tabularnewline
        SRVP \cite{Franceschi2020ICML}& \underline{$16.93 \pm 0.07$} & \bfseries 0.7799 $\pm$ 0.0020 \tabularnewline
        SLAMP-Baseline & 16.83 $\pm$ 0.06 & 0.7537 $\pm$ 0.0018 \tabularnewline
        SLAMP & \bfseries 18.07 $\pm$ 0.08 &   \underline{$0.7736 \pm 0.0019$} \tabularnewline
        \bottomrule
    \end{tabular}
\end{table}

We qualitatively compare SLAMP to SLAMP-Baseline on MNIST in \figref{fig:qual_mnist_comp}. The figure shows predictions of static and dynamic parts as appearance and motion predictions, as well the final prediction as the combination of the two. According to the mask prediction, the final prediction mostly relies on the dynamic part shown as black on the mask and uses the static component only near the motion boundaries. Moreover, optical flow prediction does not fit the shape of the digits but expands as a region until touching the motion region of the other digit.
This is due to the uniform black background. Moving a black pixel in the background randomly is very likely to result in another black pixel in the background, which means zero-loss for the warping result.
Both models can predict optical flow correctly for the most part and resort to the appearance result in the occluded regions. However, continuity in motion is better captured by SLAMP with the colliding digits whereas the baseline model cannot recover from it, leading to blur results, far from the ground truth. Note that we pick the best sample for both models among 100 samples according to LPIPS.

\boldparagraph{KTH Action Dataset}
\begin{figure}[!t]
    \centering
    \includegraphics[width=\linewidth]{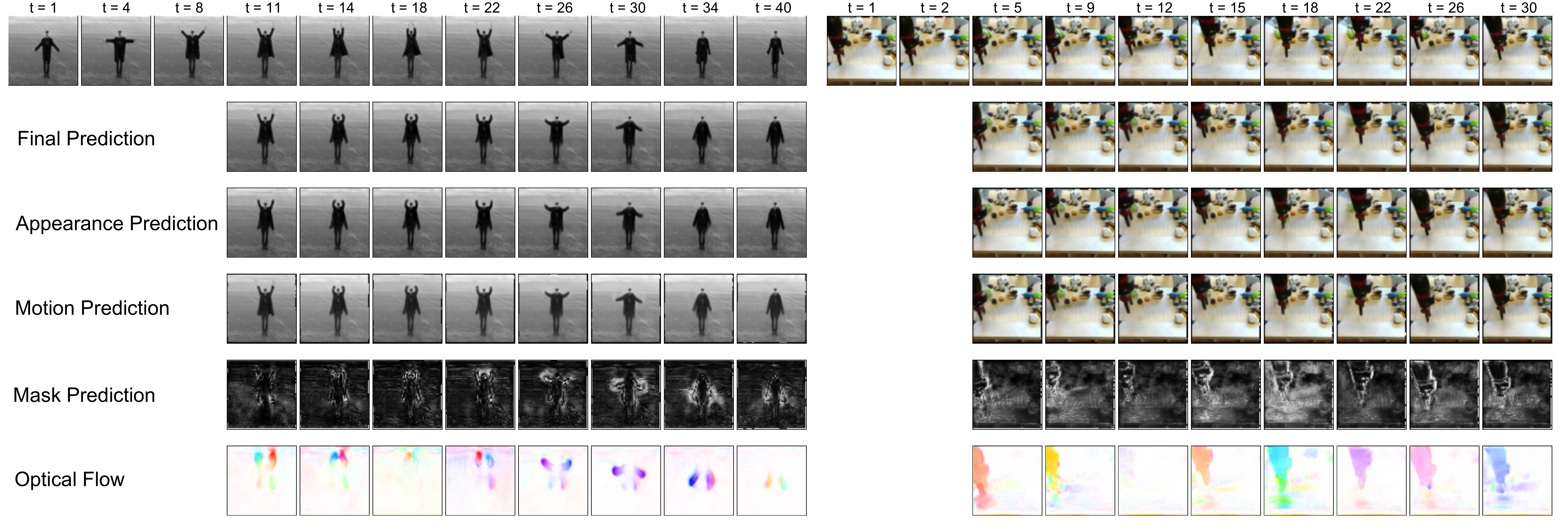}
    \caption[Qualitative Results on KTH and BAIR]{\textbf{Qualitative Results on KTH and BAIR.} We visualize the results of SLAMP on KTH (\textbf{left}) and SLAMP-Baseline on BAIR (\textbf{right}) datasets. The top row shows the ground truth, followed by the frame predictions by the final, the appearance, the motion, and the last two rows show the mask and the optical flow predictions. The mask prediction combines the appearance prediction (white) and the motion prediction (black) into the final prediction. Note that results on BAIR are without motion history because there are only two conditioning frames, \ie one flow field to condition on.}
    \label{fig:qual_kth_hist_bair_motion}
    \vspace{-5mm}
\end{figure}
\begin{figure*}[ht]
    \centering
    \includegraphics[width=\textwidth]{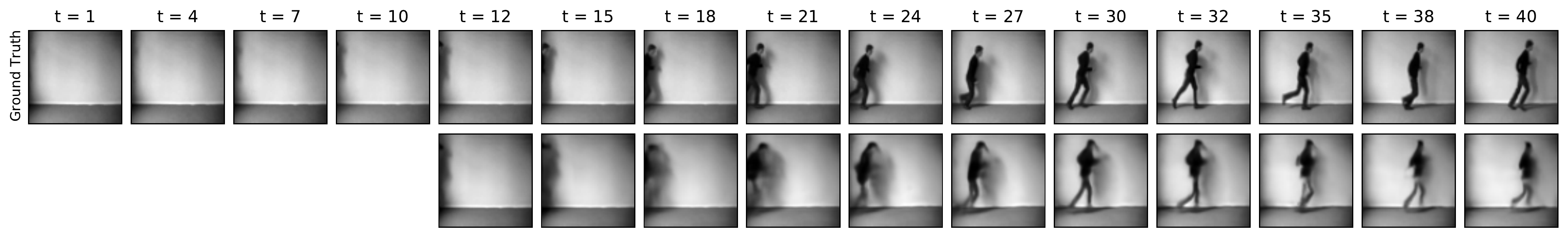}
    \caption[Example where Subject Appearing after the Conditioning Frames]{\textbf{Subject Appearing after the Conditioning Frames.} This figure shows a case where the subject appears after conditioning frames on KTH with ground truth (\textbf{top}) and a generated sample by our model (\textbf{bottom}). This shows our model's ability to capture dynamics of the dataset by generating samples close to the ground truth, even conditioned on empty frames.}
    \label{fig:qual_kth_hist_no_person}
    \vspace{-2mm}
\end{figure*}
\begin{table}[h!]
    \caption[SLAMP's Results on KTH]{
        \label{tab:res-kth}
        \textbf{Results on KTH.} This table compares the results of SLAMP and SLAMP-Baseline to the previous work on KTH dataset.
        Following the previous work, we report the results as the mean and the $95\%$-confidence interval in terms of PSNR, SSIM, and LPIPS.
        Bold and underlined scores indicate the best and the second best performing method, respectively.
    }
    \sisetup{detect-weight, table-align-uncertainty=true, mode=text}
    \renewrobustcmd{\bfseries}{\fontseries{b}\selectfont}
    \renewrobustcmd{\boldmath}{}
    \centering
    \vspace{0.1in}
    \begin{tabular}{lccc}
        \toprule
        Models & {PSNR} & {SSIM} & {LPIPS} \tabularnewline
        \midrule
        SV2P \cite{Babaeizadeh2018ICLR} & 28.19 $\pm$ 0.31 & 0.8141 $\pm$ 0.0050 & 0.2049 $\pm$ 0.0053 \tabularnewline
        SAVP \cite{Lee2018ARXIV} & 26.51 $\pm$ 0.29 & 0.7564 $\pm$ 0.0062 & 0.1120 $\pm$ 0.0039 \tabularnewline
        SVG \cite{Denton2018ICML} & 28.06 $\pm$ 0.29 & 0.8438 $\pm$ 0.0054 & 0.0923 $\pm$ 0.0038 \tabularnewline
        SRVP \cite{Franceschi2020ICML} & \bfseries 29.69 $\pm$ 0.32 & \bfseries 0.8697 $\pm$ 0.0046 & \bfseries 0.0736 $\pm$ 0.0029 \tabularnewline
        SLAMP-Baseline &  29.20 $\pm$ 0.28 & 0.8633 $\pm$ 0.0048 & 0.0951 $\pm$ 0.0036 \tabularnewline
        SLAMP & \underline{$29.39 \pm 0.30$} & \underline{$0.8646 \pm 0.0050$} & \underline{$0.0795 \pm 0.0034$} \tabularnewline
        \bottomrule
    \end{tabular}
\end{table}

KTH dataset contains real videos where people perform a single action such as walking, running, boxing, \etc in front of a static camera~\cite{Schuldt2004CVPR}. We expect our model with motion history to perform very well by exploiting regularity in human actions on KTH.
Following the same training and evaluation settings used in the previous work, we condition on the first 10 frames and learn to predict the next 10 frames. During testing, we again condition on the first 10 frames but predict the next 30 frames. 

\figref{fig:results_kth_bair} and \tabref{tab:fvd} show quantitative results on KTH in comparison to previous approaches. Both our baseline and SLAMP models outperform previous approaches and perform comparably to SRVP, in all metrics including FVD. A detailed visualization of all three frame predictions as well as flow and mask are shown in \figref{fig:qual_kth_hist_bair_motion}. Flow predictions are much more fine-grained than MNIST by capturing fast motion of small objects such as hands or thin objects such as legs (see Appendix). The mask decoder learns to identify regions around the motion boundaries which cannot be matched with flow due to occlusions and assigns more weight to the appearance prediction in these regions.

On KTH, the subject might appear after the conditioning frames. These challenging cases can be problematic for some previous work as shown in SRVP~\cite{Franceschi2020ICML}. Our model can generate samples close to the ground truth despite very little information on the conditioning frames as shown in \figref{fig:qual_kth_hist_no_person}. The figure shows the best sample in terms of LPIPS, please see Appendix for a diverse set of samples with subjects of various poses appearing at different time steps.

\boldparagraph{BAIR Robot Hand}
This dataset contains videos of a robot hand moving and pushing objects on a table \cite{EbertFLL17}. Due to uncertainty in the movements of the robot arm, 
BAIR is a standard dataset for evaluating stochastic video prediction models.
Following the training and evaluation settings used in the previous work, we condition on the first 2 frames and learn to predict the next 10 frames. During testing, we again condition on the first 2 frames but predict the next 28 frames.

\begin{figure*}[ht!]
\centering
   \includegraphics[width=1\linewidth]{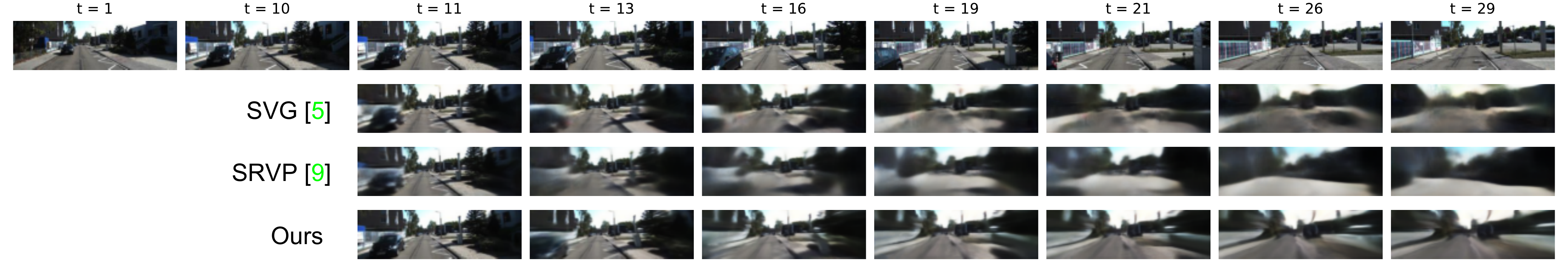}
   \includegraphics[width=1\linewidth]{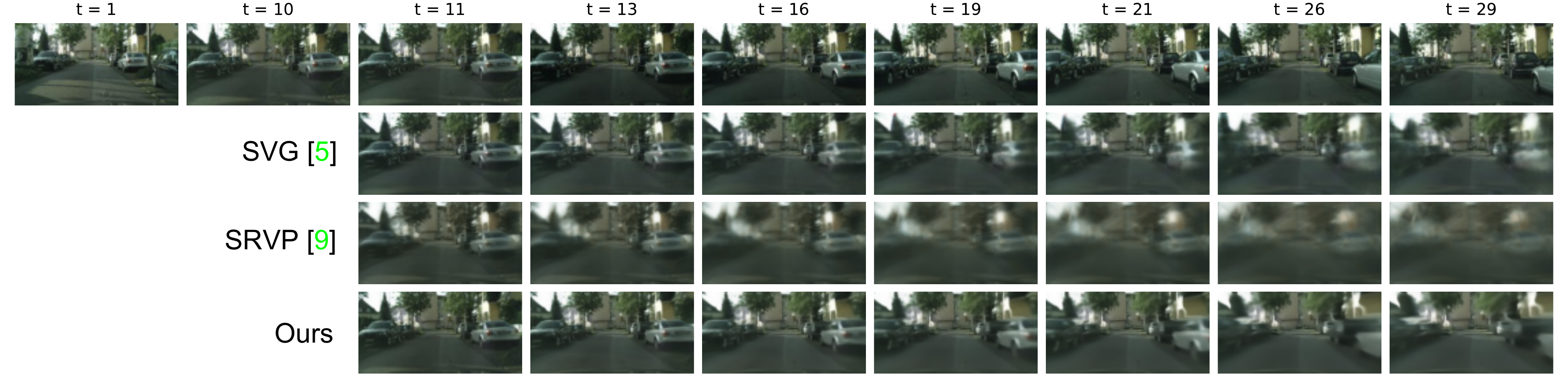}
\vspace{-2.5mm}
\caption[Qualitative Comparison on KITTI and Cityscapes]{\textbf{Qualitative Comparison on KITTI and Cityscapes.} We compare SLAMP to SVG~\cite{Denton2018ICML} and SRVP~\cite{Franceschi2020ICML} on KITTI~(\textbf{top}) and Cityscapes~(\textbf{bottom}). Our model can better capture the changes due to ego-motion thanks to explicit modeling of motion history.}
\label{fig:kitti_city_qual}
\vspace{-4.5mm}
\end{figure*}

\begin{table}[h!]
    \caption[SLAMP's Results on BAIR]{
        \label{tab:res-bair}
        \textbf{Results on BAIR.} This table compares the results of SLAMP and SLAMP-Baseline to the previous work on BAIR dataset.
        Following the previous work, we report the results as the mean and the $95\%$-confidence interval in terms of PSNR, SSIM, and LPIPS.
        Bold and underlined scores indicate the best and the second best performing method, respectively.
    }
    \sisetup{detect-weight, table-align-uncertainty=true, mode=text}
    \renewrobustcmd{\bfseries}{\fontseries{b}\selectfont}
    \renewrobustcmd{\boldmath}{}
    \centering
    \vspace{0.1in}
    \begin{tabular}{lccc}
        \toprule
        Models & {PSNR} & {SSIM} & {LPIPS} \tabularnewline
        \midrule
        SV2P \cite{Babaeizadeh2018ICLR}& \bfseries 20.39 $\pm$ 0.27 &  0.8169 $\pm$ 0.0086 & 0.0912 $\pm$ 0.0053 \tabularnewline
        SAVP \cite{Lee2018ARXIV} & 18.44 $\pm$ 0.25 & 0.7887 $\pm$ 0.0092 & 0.0634 $\pm$ 0.0026 \tabularnewline
        SVG \cite{Denton2018ICML}& 18.95 $\pm$ 0.26 & 0.8058 $\pm$ 0.0088 & 0.0609 $\pm$ 0.0034 \tabularnewline
        SRVP \cite{Franceschi2020ICML} & 19.59 $\pm$ 0.27 & \bfseries 0.8196 $\pm$ 0.0084 & \bfseries 0.0574 $\pm$ 0.0032 \tabularnewline
        SLAMP-Baseline & 19.60 $\pm$ 0.26 &  \underline{$0.8175 \pm 0.0084$} &  \underline{$0.0596 \pm 0.0032$} \tabularnewline
        SLAMP & \underline{$19.67 \pm 0.26$} &  0.8161 $\pm$ 0.0086 & 0.0639 $\pm$ 0.0037 \tabularnewline
        \bottomrule
    \end{tabular}
\end{table}

We show quantitative results on BAIR in \figref{fig:results_kth_bair} and \tabref{tab:fvd}. Our baseline model achieves comparable results to SRVP, outperforming other methods in all metrics except SV2P~\cite{Babaeizadeh2018ICLR} in PSNR and SAVP~\cite{Lee2018ARXIV} in FVD. With 2 conditioning frames only, SLAMP cannot utilize the motion history and performs similarly to the baseline model on BAIR (see Appendix). This is simply due to the fact that there is only one flow field to condition on, in other words, no motion history. Therefore, we only show the results of the baseline model on this dataset.

\begin{table}[t!]
    \sisetup{detect-weight, table-align-uncertainty=true, mode=text}
    \renewrobustcmd{\bfseries}{\fontseries{b}\selectfont}
    \renewrobustcmd{\boldmath}{}
    \centering
    \label{tab:kitti_city_results2}
    \caption[SLAMP's Results with a Moving Background]{\textbf{Results with a Moving Background.} We evaluate our model SLAMP in comparison to SVG and SRVP on %
    KITTI~\cite{Geiger2013IJRR} and Cityscapes~\cite{Cordts2016CVPR} datasets by conditioning on 10 frames and predicting 20 frames into the future.}
    \label{tab:kitti_city_results2}
    \begin{tabular}{lccc}%
        \toprule
        Models &  PSNR ($\uparrow$) & SSIM ($\uparrow$) & LPIPS ($\downarrow$) \tabularnewline 
        \midrule
        SVG~\cite{Denton2018ICML} & 12.70 $\pm$ 0.70 & 0.329 $\pm$ 0.030 &  \underline{0.594} $\pm$ 0.034 
        \tabularnewline
        SRVP~\cite{Franceschi2020ICML}  & \underline{13.41} $\pm$ 0.42 & \underline{0.336} $\pm$ 0.034 & 0.635 $\pm$ 0.021 \tabularnewline
        SLAMP & \bfseries 13.46 $\pm$ 0.74 & \bfseries 0.337 $\pm$ 0.034 & \bfseries 0.537 $\pm$ 0.042
        \tabularnewline       \bottomrule
    \end{tabular}
    \caption*{KITTI~\cite{Geiger2012CVPR, Geiger2013IJRR} %
    }
    \begin{tabular}{lccc}
        \toprule
        Models &  PSNR ($\uparrow$) & SSIM ($\uparrow$) & LPIPS ($\downarrow$) \tabularnewline 
        \midrule
        SVG~\cite{Denton2018ICML} & 20.42 $\pm$ 0.63 & \underline{0.606} $\pm$ 0.023 & \underline{0.340} $\pm$ 0.022
        \tabularnewline
        SRVP~\cite{Franceschi2020ICML}  & \underline{20.97} $\pm$ 0.43 & 0.603 $\pm$ 0.016 & 0.447 $\pm$ 0.014 \tabularnewline
        SLAMP & \bfseries 21.73 $\pm$ 0.76 & \bfseries 0.649 $\pm$ 0.025 & \bfseries 0.2941 $\pm$ 0.022
        \tabularnewline       \bottomrule
    \end{tabular}
    \caption*{Cityscapes~\cite{Cordts2016CVPR}}
\end{table}

\boldparagraph{Real-World Driving Datasets} We perform experiments on two challenging autonomous driving datasets: KITTI~\cite{Geiger2012CVPR, Geiger2013IJRR} and Cityscapes~\cite{Cordts2016CVPR} with various challenges.
Both datasets contain everyday real-world scenes with complex dynamics due to both background and foreground motion. KITTI is recorded in one town in Germany, while Cityscapes is recorded in 50 European cities, leading to higher diversity.

Cityscapes primarily focuses on semantic understanding of urban street scenes, therefore, contains a larger number of dynamic foreground objects compared to KITTI. However, motion lengths are larger on KITTI due to lower frame rate. On both datasets, we condition on 10 frames and predict 10 frames into the future to train our models. Then at test time, we predict 20 frames conditioned on 10 frames.

As shown in \tabref{tab:kitti_city_results2}, SLAMP outperforms both methods on all of the metrics on both datasets, which shows its ability to generalize to the sequences with moving background. Even SVG \cite{Denton2018ICML} performs better than the state of the art SRVP~\cite{Franceschi2020ICML} in the LPIPS metric for KITTI and on both SSIM and LPIPS for Cityscapes, which shows the limitations of SRVP on scenes with dynamic backgrounds. We also perform a qualitative comparison to these methods in \figref{fig:kitti_city_qual}. SLAMP can better preserve the scene structure thanks to explicit modeling of ego-motion history in the background.

\begin{figure}[h]
\centering
\includegraphics[width=\textwidth]{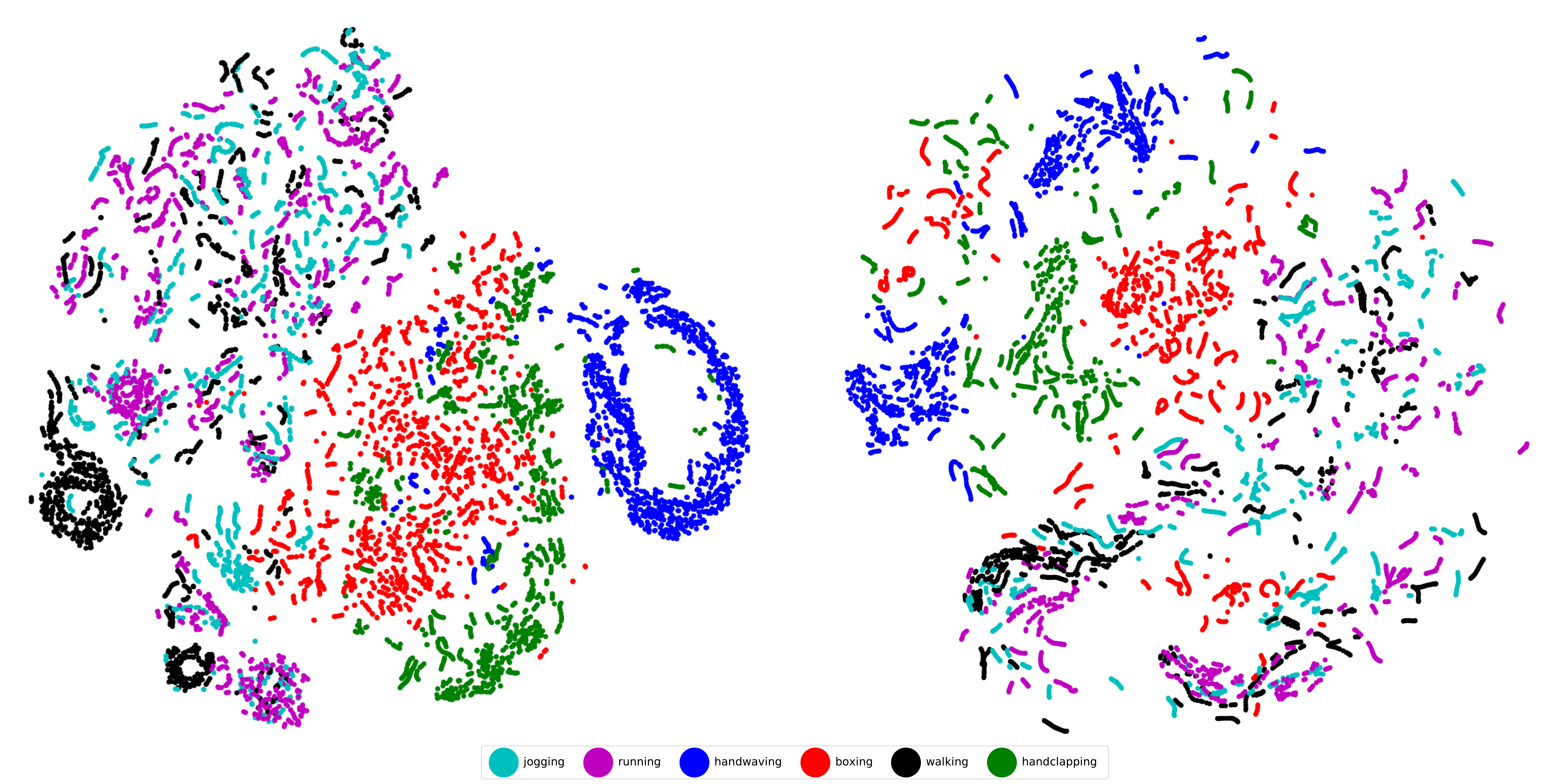}
\caption[Visualization of Latent Variables]{\textbf{Dynamic (left) vs. Static (right) Latent Variables}. This figure shows the T-SNE visualization of dynamic and static latent variables on 300 test videos from the KTH dataset. In dynamic latent variables, different classes with similar repetitive movements such as walking, running, and jogging are clustered together. However, in static latent variables, points are more scattered and do not form clusters according to semantic actions.}
\label{fig:tsne-both}
\end{figure}

\boldparagraph{Visualization of Latent Space}
We visualize stochastic latent variables of the dynamic component on KTH compared to the static latent variables. We provide both the static and the dynamic components for comparison.  As can be seen from \figref{fig:tsne-both}, static variables on the right are more scattered and do not form clusters according to semantic classes as in the dynamic variables on the left. This shows that our model can learn video dynamics according to semantic classes with separate modeling of the dynamic component.

\section{Using a different architecture}

In this section, we show the superiority of SLAMP's decoding scheme by using a different method, namely SRVP~\cite{Franceschi2020ICML}. We integrate SLAMP's extra decoder into SRVP's state-space model in order to find out whether SLAMP's decoding can be extended into other architecture or not.

\boldparagraph{SRVP Architecture}
Franceschi~\etal~\cite{Franceschi2020ICML} introduced SRVP, a state-space model for stochastic video prediction model. SRVP encodes the scene into a stochastic latent variable and uses time-independent residual updates to create a different state representation for each time step. The SRVP separates the content and the dynamic by using a separate deterministic content network to encode the scene details and using it for the decoding of future frames. In this way, the latent state will learn the dynamics, \eg motion in the scene, and the content part will learn the details of the scene. 

In each time step, SRVP~\cite{Franceschi2020ICML} creates a state variable, $by_t$, by using a stochastic residual network to update previous states into future ones. 

\begin{ceqn}
\begin{equation}
    \by_{t + \Delta t} = \by_{t} + \Delta t \cdot f_{\theta} \big (\by_{t}, \bz_{\lfloor t \rfloor + 1} \big ).
\end{equation}
\end{ceqn}

where $by_{t}$ is the state representation at time $t$, $\Delta t$ is a hyperparameter that defines the size of the updates, $\bz$ is the latent variable sampled from prior or posterior distribution according to train or inference time. After creating state representation for each time step, SRVP decodes all of them into video frames, \eg images, which can be seen in \eqref{eq:srvp_decode}.

\begin{ceqn}
\begin{equation}
    \label{eq:srvp_decode}
    \bx_{t} =  g_{\theta} \big (\by_{t}, \bw \big ).
\end{equation}
\end{ceqn}

where $bx_t$ is the image frame at time $t$, $g_{\theta}$ is the decoder parameterized by $\theta$, $\by_t$ is the state representation at time $t$, $\bw$ is the content feature extracted by deterministic content network.

\boldparagraph{Addition of SLAMP's decoder}

By following SLAMP, we add another decoder, \emph{motion decoder} $g_{\theta}^f$, into SRVP to decode the motion from the previous frame to the next frame. After the motion, \eg optical flow is predicted, we use it to warp the previous image into the next ones. In motion decoder, we use two consecutive state representations, \eg to decode the motion from $t-1$ to $t$, we use $\by_{t-1}$ and $\by_{t}$. Moreover, we use a content vector similar to the original decoder of SRVP.

\begin{table}[ht!]
    \caption[SRVP++ Results on generic datasets]{
         \textbf{SRVP++ Comparisons on generic video prediction datasets} This table shows the quantitative results comparing the SRVP++ with SVG~\cite{Denton2018ICML}, the original SRVP~\cite{Franceschi2020ICML} and SLAMP. \textbf{Direct} refers to direct masking strategy whereas \textbf{Mask} refer to binary mask strategy.
         Following the previous work, we report the results as the mean and the $95\%$-confidence interval in terms of PSNR, SSIM, and LPIPS on all the datasets except LPIPS on MNIST.
    }
    \label{tab:srvp++-generic}
    \sisetup{detect-weight, table-align-uncertainty=true, mode=text}
    \renewrobustcmd{\bfseries}{\fontseries{b}\selectfont}
    \renewrobustcmd{\boldmath}{}
    \centering
    \vspace{0.1in}
    \begin{tabular}{clccc}
        \toprule
        {} & Models & {PSNR ($\uparrow$)} & {SSIM ($\uparrow$)} & {LPIPS ($\downarrow$)} \tabularnewline
        \midrule
        \multirow{5}{*}{\rotatebox[origin=c]{90}{\textbf{MNIST}}}
        {} & SVG  & 14.50 $\pm$ 0.04 & 0.7090 $\pm$ 0.0015 & \textemdash \tabularnewline
        {} & SRVP & 16.93 $\pm$ 0.07 & \bfseries 0.7799 $\pm$ 0.0020 & \bfseries 0.1129 $\pm$ 0.0010 \tabularnewline
        {} & SLAMP &  \underline{$18.07 \pm 0.08$} &   \underline{$0.7736 \pm 0.0019$} & \underline{0.1282 $\pm$ 0.0006} \tabularnewline
        {} & SRVP++ - Direct & \bfseries 18.14 $\pm$ 0.08 & 0.7634 $\pm$ 0.0019 & 0.1323 $\pm$ 0.0012 \tabularnewline
        {} & SRVP++ - Mask & 16.57 $\pm$ 0.05 & 0.7367 $\pm$ 0.0018 & 0.1597 $\pm$ 0.0011 \tabularnewline
        \midrule \midrule
        \multirow{5}{*}{\rotatebox[origin=c]{90}{\textbf{KTH}}}
        {} & SVG  & 28.06 $\pm$ 0.29 & 0.8438 $\pm$ 0.0054 & 0.0923 $\pm$ 0.0038 \tabularnewline
        {} & SRVP & \bfseries 29.69 $\pm$ 0.32 & \bfseries 0.8697 $\pm$ 0.0046 & \bfseries 0.0736 $\pm$ 0.0029 \tabularnewline
        {} & SLAMP & \underline{$29.39 \pm 0.30$} & \underline{$0.8646 \pm 0.0050$} & \underline{$0.0795 \pm 0.0034$} \tabularnewline
        {} & SRVP++ - Direct & 28.04 $\pm$ 0.26 & 0.8541 $\pm$ 0.0053 & 0.0948 $\pm$ 0.0034 \tabularnewline
        {} & SRVP++ - Mask & 28.76 $\pm$ 0.30 & 0.8612 $\pm$ 0.0053 & 0.0836 $\pm$ 0.0034 \tabularnewline
        \midrule \midrule
        \multirow{5}{*}{\rotatebox[origin=c]{90}{\textbf{BAIR}}}
        {} & SVG & 18.95 $\pm$ 0.26 & 0.8058 $\pm$ 0.0088 & 0.0609 $\pm$ 0.0034 \tabularnewline
        {} & SRVP  & 19.59 $\pm$ 0.27 &  \underline{0.8196 $\pm$ 0.0084} & \bfseries 0.0574 $\pm$ 0.0032 \tabularnewline
        {} & SLAMP & $19.67 \pm 0.26$ &  0.8161 $\pm$ 0.0086 & 0.0639 $\pm$ 0.0037 \tabularnewline
        {} & SRVP++ - Direct & \underline{19.74 $\pm$ 0.26} & \bfseries 0.8206 $\pm$ 0.0084 & \underline{0.0598 $\pm$ 0.0033} \tabularnewline
        {} & SRVP++ - Mask & 19.01 $\pm$ 0.26 & 0.8089 $\pm$ 0.0087 & 0.0663 $\pm$ 0.0037 \tabularnewline
        \bottomrule
    \end{tabular}
\end{table}

So, the motion decoder of SRVP works as follows:

\begin{ceqn}
\begin{eqnarray}
    \bff_{t-1:t} &=&  g^f_{\theta} \big (\by_{t-1}, \by_{t}, \bw \big ). \\
    \bx_{t} &=& \mathrm{Warp}(\bx_{t-1}, \bff_{t-1:t}) \nonumber
\end{eqnarray}
\end{ceqn}

where $\bff_{t-1:t}$ is the optical flow from time $t-1$ to $t$, $g^f_{\theta}$ is the motion decoder, $\by$'s are the representations, Warp is a function for differentiable warping \cite{Jaderberg2015NeurIPS}. 

The goal is to learn the motion dynamics of the scene by explicitly decoding them. In the end, \emph{new} SRVP architecture will have two predictions coming from two different branches. We employ the same strategy as SLAMP's for the combination, see \eqref{eq:combined_x}. Moreover, we propose a different combination strategy. Instead of predicting a binary mask to combine predictions of each branch, the mask decoder will directly output the final image. The intuition is to let the network decide the pixels instead of explicitly selecting a pixel. We ablate this combination strategy in our experiments. See the appendix for the overall architectural figure of SRVP++.

\boldparagraph{Experiments}
In order to analyze the proposed decoding scheme, we compare the new SRVP, which we name SRVP++, with SVG, SRVP, and SLAMP.

We first show results on generic video prediction datasets, namely MNIST, KTH~\cite{Schuldt2004CVPR} and BAIR~\cite{Ebert2017CORL}. According to the results on \tabref{tab:srvp++-generic}, SRVP++ decoding strategy does not improve the results. We suspect that this is due to the static background of the generic video prediction datasets. SRVP's assumption of static background holds for these datasets. However, it fails for the real-world driving datasets with moving background, as it is explained before in \secref{sec:chap2-exp}.

We further evaluate the models on challenging real-world datasets, KITTI~\cite{Geiger2012CVPR, Geiger2013IJRR} and Cityscapes~\cite{Cordts2016CVPR} following SLAMP. Since the static background assumption of SRVP fails on the real-world driving datasets with moving background, SRVP++ performs better than SRVP. The additional motion decoder can learn the motion in the scene and break the assumption of SRVP. By warping the previous frames, SRVP++ can generate higher-quality images than SRVP. The results in \tabref{tab:srvp++-real} show that SLAMP's decoding scheme can be applied to different methods and improve their performance on challenging real-world datasets.

\begin{table}[h!]
    \caption[SRVP++ Results on real-world datasets]{
         \textbf{SRVP++ Comparisons on real-world driving datasets} This table shows the quantitative results comparing the SRVP++ with SVG~\cite{Denton2018ICML}, the original SRVP~\cite{Franceschi2020ICML} and SLAMP on KITTI~\cite{Geiger2012CVPR, Geiger2013IJRR} and Cityscapes~\cite{Cordts2016CVPR}. \textbf{Direct} refers to direct masking strategy whereas \textbf{Mask} refer to binary mask strategy.
         Following the previous work, we report the results as the mean and the $95\%$-confidence interval in terms of PSNR, SSIM, and LPIPS on all the datasets.
    }
    \label{tab:srvp++-real}
    \sisetup{detect-weight, table-align-uncertainty=true, mode=text}
    \renewrobustcmd{\bfseries}{\fontseries{b}\selectfont}
    \renewrobustcmd{\boldmath}{}
    \centering
    \vspace{0.1in}
    \begin{tabular}{clccc}
        \toprule
        {} & Models & {PSNR ($\uparrow$)} & {SSIM ($\uparrow$)} & {LPIPS ($\downarrow$)} \tabularnewline
        \midrule
        \multirow{5}{*}{\rotatebox[origin=c]{90}{\textbf{KITTI}}}
        {} & SVG & 12.70 $\pm$ 0.70 & 0.329 $\pm$ 0.030 &  0.594 $\pm$ 0.034 
        \tabularnewline
        {} & SRVP  & 13.41 $\pm$ 0.42 & {0.336} $\pm$ 0.034 & 0.635 $\pm$ 0.021 \tabularnewline
        {} & SLAMP &  {13.46 $\pm$ 0.74} &  \underline{0.337 $\pm$ 0.034} &  0.537 $\pm$ 0.042 \tabularnewline
        {} & SRVP++ - Direct & \bfseries 14.09 $\pm$ 0.49 &  \bfseries 0.345 $\pm$ 0.023 &  \underline{0.505 $\pm$ 0.020} \tabularnewline
        {} & SRVP++ - Mask &  \underline{14.01 $\pm$ 0.50} & 0.331 $\pm$ 0.024 & \bfseries 0.455 $\pm$ 0.020 \tabularnewline
        \midrule \midrule
        \multirow{5}{*}{\rotatebox[origin=c]{90}{\textbf{Cityscapes}}}
        {} & SVG & 20.42 $\pm$ 0.63 & \underline{0.606} $\pm$ 0.023 & 0.340 $\pm$ 0.022
        \tabularnewline
        {} & SRVP  & \underline{20.97} $\pm$ 0.43 & 0.603 $\pm$ 0.016 & 0.447 $\pm$ 0.014 \tabularnewline
        {} & SLAMP & \bfseries 21.73 $\pm$ 0.76 & \bfseries 0.649 $\pm$ 0.025 & \bfseries 0.2941 $\pm$ 0.022 \tabularnewline
        {} & SRVP++ - Direct & 20.11 $\pm$ 0.42 &  0.596 $\pm$ 0.017 & 0.331 $\pm$ 0.011 \tabularnewline
        {} & SRVP++ - Mask & 20.35 $\pm$ 0.39 &  0.587 $\pm$ 0.015 & \underline{0.314 $\pm$ 0.008} \tabularnewline
        \bottomrule
    \end{tabular}
\end{table}

\section{Conclusion}
\label{sec:chap2-conc}

We presented a stochastic video prediction framework to decompose video content into appearance and dynamic components. 
Our baseline model with deterministic motion and mask decoders outperforms SVG, which is a special case of our baseline model. 
Our model with motion history, SLAMP, further improves the results and reaches the performance of the state-of-the-art method SRVP on the previously used datasets. Moreover, it outperforms both SVG and SRVP on two real-world autonomous driving datasets with dynamic background and complex motion. We show that motion history enriches the model's capacity to predict the future, leading to better predictions in challenging cases. 

Moreover, we show that the decoding scheme of SLAMP can be used in other video prediction methods, \eg SRVP~\cite{Franceschi2020ICML}, and improve the performance on real-world datasets with a moving background.

Our model with motion history cannot realize its full potential in standard settings of stochastic video prediction datasets. A fair comparison is not possible on BAIR due to the small number of conditioning frames. BAIR holds a great promise with changing background but infrequent, small changes are not reflected in current evaluation metrics.

\clearpage
\thispagestyle{plain}
~
\newpage
\chapter{Stochastic Video Prediction with Structure and Motion}
\label{chapter:chap3}

\section{Introduction}
\label{sec:slamp3d-intro}

The world observed from a moving vehicle can be decomposed into a static part that moves only according to the vehicle's motion, or the ego-motion, and a dynamic part that contains independently moving objects. With this separation, the decomposition of the static and the dynamic parts is inevitable. The static part of the video does not have any motion; however, they move in the video according to the motion of the vehicle. This part can easily be explained by the domain knowledge following the previous work on structure and motion~\cite{Zhou2017CVPR, Godard2019ICCV}. The static part is modeled by both the 3D structure and the ego-motion. On the other hand, the motion of the objects that move independently cannot be explained by ego-motion. There should be a separate mechanism to model the remaining motion in the scene after applying the ego-motion. The most suitable option is to use optical flow on top of the ego-motion for modeling the independently moving objects. 

The key factor in this design is that the optical flow should correspond to residual or remaining motion in the scene after applying the ego-motion. To do this, the dynamic part, which corresponds to the independently moving objects, should be conditioned on the static part. In this way, the dynamic part will be aware of the static part, \eg ego-motion, and can be modeled as a residual motion in the scene. Following previous work on video decomposition~\cite{Detlefsen2019NeurIPS}, which conditions the body pose on the shape, we condition the motion of the dynamic part on the static part to achieve a disentanglement between the two types of motion in driving.

In this chapter, we propose SLAMP-3D. SLAMP-3D decomposes the driving video into static and dynamic parts. The static part is handled by the 3D structure and the ego-motion. On top of this, the dynamic part is covered with the residual flow using the conditioning of the dynamic on top of the static part. With this separation, the performance of future prediction in real-world scenes with moving background and independently moving foreground objects on two real-world driving datasets, KITTI~\cite{Geiger2012CVPR, Geiger2013IJRR} and Cityscapes~\cite{Cordts2016CVPR}. Furthermore, conditioning the object motion on the ego-motion improves the results, especially for foreground objects in dynamic scenes of Cityscapes.

\begin{figure}[ht!]
    \centering
    \includegraphics[width=\linewidth]{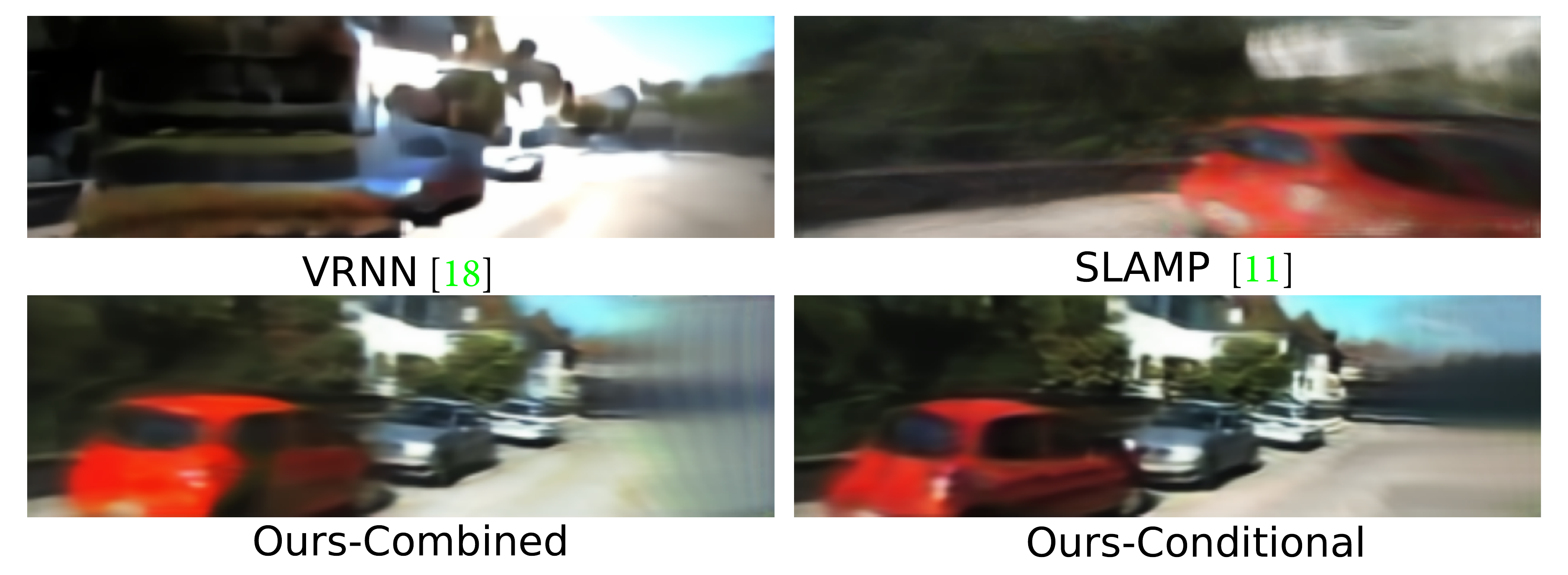}
    \caption[Future Prediction while Turning]{\textbf{Future Prediction while Turning.} We compare the future prediction results of previous methods (\textbf{top}) to ours (\textbf{bottom}) while the vehicle makes a right turn. While previous methods fail due to a less frequent scenario in the dataset, our method can generate a better future prediction due to explicit modeling of the structure and ego-motion. This figure shows a single frame after the conditioning frames, please see \figref{fig:qual_comp_kitti} for the whole sequence.}
    \label{fig:slamp3d-teaser}
\end{figure}

With this conditioned decomposition, SLAMP-3D can predict the future frames even in the most challenging cases, such as turning. In \figref{fig:slamp3d-teaser}, SLAMP-3D can preserve the structure of the red car easily while the other methods fail to do so.

\begin{figure*}[t!]
\centering
\includegraphics[width=\textwidth]{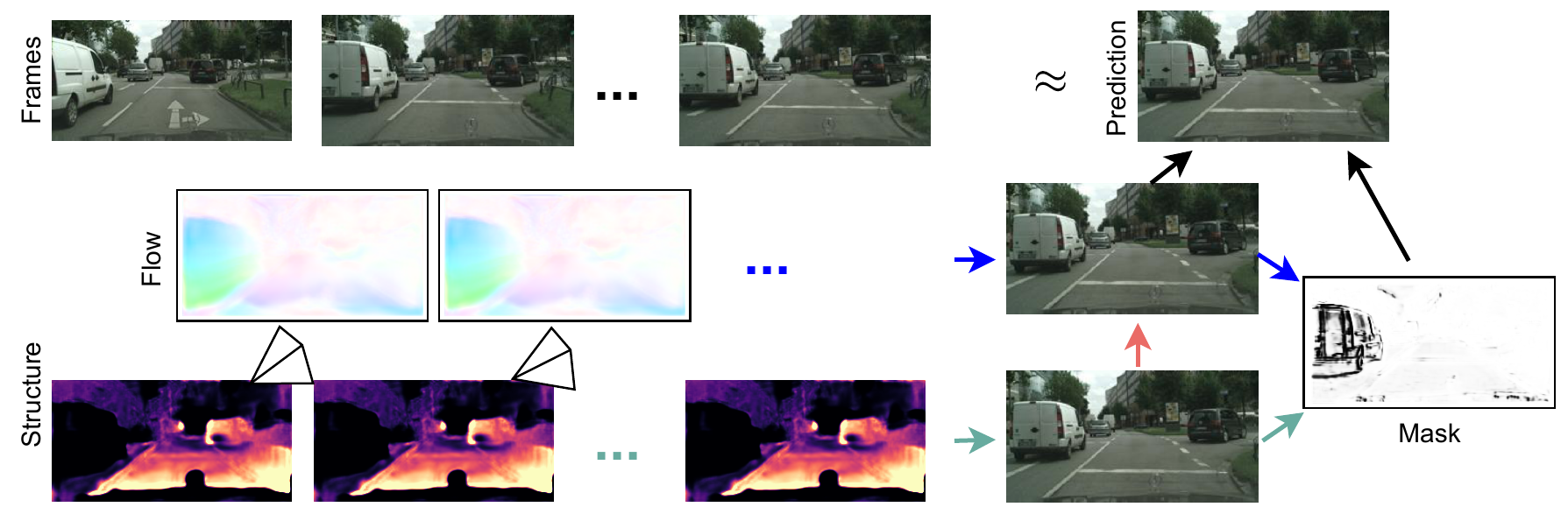}
    \caption[Model Overview]{\textbf{Model Overview.} For predicting future frames in a video~(\textbf{top}), we predict depth on each frame and camera pose, and optical flow between consecutive frames. We model the static part of the scene with depth and pose~(\textbf{bottom}). Conditioned on the static part, we model the remaining motion due to moving foreground objects with the flow in the dynamic parts of the scene (\textbf{middle}). We obtain the final prediction~(\textbf{top-right}) by combining the static and the dynamic predictions based on the predicted mask~(\textbf{right}, dynamic in black). The moving car is missing in depth because it is predicted by the dynamic. The arrow from the static to the final prediction is omitted for clarity.}
\label{fig:slamp3d-overview}
\end{figure*}

The overall architecture of SLAMP-3D can be seen in the \figref{fig:slamp3d-overview}.

\section{Methodology}
\label{sec:chap3-method}

We investigate the effects of decomposing the scene into static and dynamic parts for stochastic video prediction, inspired by and built on top of SLAMP.
We assume that the background or the static part of the scene, moves only according to the motion of the ego-vehicle and can be explained by the ego-motion and the scene structure.
We first predict future depth and ego-motion conditioned on a few given frames. This is different than SLAMP which predicts the static part in the pixel-space. By using the predictions of the structure and the ego-motion, we synthesize the static part of the scene. We model the remaining motion due to independently moving objects, the dynamic part of the scene, as the residual flow on top of the ego-motion. At the end, we combine the static and the dynamic predictions with a learned mask to generate the final predictions.

\subsection{Stochastic Video Prediction (SVP)}

The same notation given in \secref{sec:svg} is preserved throughout this chapter.

\subsection{From SLAMP to SLAMP-3D}

In this chapter, we propose to extend SLAMP, explained in Chapter~\ref{chapter:slamp}, to SLAMP-3D by incorporating scene structure into the estimation. Similar to SLAMP, we decompose the scene into static and dynamic components where the static component focus on the changes in the background due to camera motion and the dynamic component on the remaining motion in the scene due to independently moving objects. There are two major differences with respect to SLAMP. First, in this work, we model the static parts of the scene in the motion space as well to better represent the changes in the background due to the ego-motion of the vehicle.
Second, we condition the prediction of dynamic component on the static to predict the residual motion in the scene, \ie the remaining motion after the scene moves according to ego-motion due to the independently moving objects. 

\begin{figure}[t!]
\centering
\includegraphics[width=\linewidth]{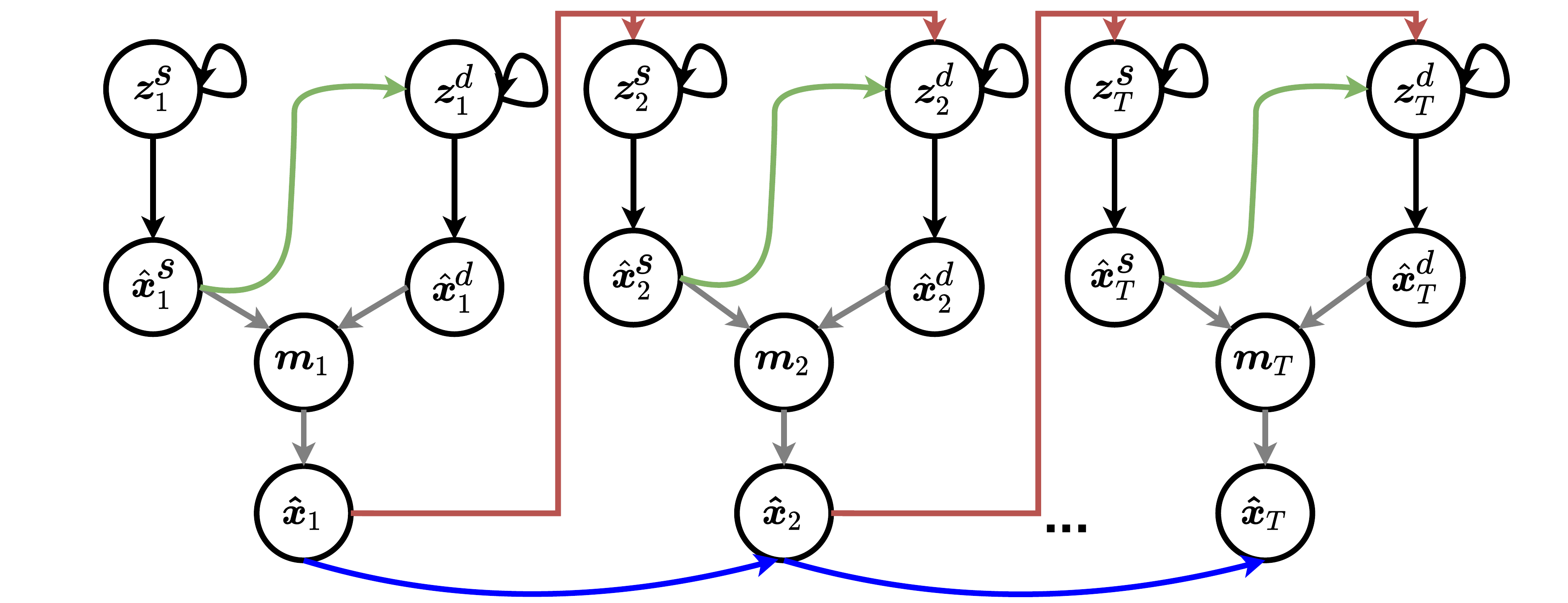}
\caption[Graphical Model of SLAMP-3D]{\textbf{Graphical Model of SLAMP-3D.} The graphical model shows the generation process of our model with static and dynamic latent variables $\bz^s_{t}$ and $\bz^d_{t}$ generating frames $\hat{\bx}^s_{t}$ and $\hat{\bx}^d_{t}$.
Information is propagated between time-steps through the recurrence between frame predictions (\textcolor{blue}{blue}), corresponding latent variables, and from frame predictions to latent variables (\textcolor{OrangeRed}{red}). The dependency of dynamic latent variables on static is shown in \textcolor{LimeGreen}{green}.}
\label{fig:graphical}
\end{figure}

Similar to SLAMP, we compute two separate distributions, $q_{\bphi_s}\left(\bz_t^s \mid \bx_{1:t}\right)$ and $q_{\bphi_d}\left(\bz_t^d \mid \bx_{1:t}\right)$, for static and dynamic components, respectively. This way, both distributions focus on different parts of the scene which decomposes scene into static and dynamic parts. Since the camera motion applies both to the background and to the objects, dynamic component only needs to learn the residual motion. We achieve this by introducing an explicit conditioning of the dynamic latent variable $\bz_t^d$ on the static latent variable $\bz_t^s$:
\begin{ceqn}

\begin{align}
    \label{eq:slamp3d_prior_post}
    p\left(\bz_t^s \mid \bx_{1:t}\right) &\approx q_{\bphi_s}\left(\bz_t^s \mid \bx_{1:t}\right) \quad \text{and} \\
    p\left(\bz_t^d \mid \bx_{1:t}\right) &\approx q_{\bphi_d}\left(\bz_t^d \mid \bx_{1:t}, \bz_t^s\right) \nonumber
\end{align}
\end{ceqn}
With two latent variables, we extend the stochasticity to the motion space, separately for the ego-motion and the object motion. Furthermore, we condition the dynamic component on the static to define the object motion as residual motion that remains after explaining the scene according to camera motion, similar to the disentanglement of the body pose and the shape in \cite{Detlefsen2019NeurIPS}.

\boldparagraph{Two Types of Motion History} Similar to SLAMP, we model the motion history in SLAMP-3D but by disentangling it to ego-motion and residual motion. Essentially, we learn two separate motion histories for the static and the dynamic components. The latent variables $\bz_t^s$ and $\bz_t^d$ contain motion information accumulated over the previous frames rather than local temporal changes between the last frame and the target frame. This is achieved by encouraging each posterior, $q_{\bphi_s}\left(\bz_t^s \mid \bx_{1:t}\right)$ and $q_{\bphi_d}\left(\bz_t^d \mid \bx_{1:t}, \bz_t^s\right)$, to be close to a prior distribution in terms of KL-divergence which can be seen in \eqref{eq:slamp3d_prior_post}. 
Similar to SVG~\cite{Denton2018ICML} and SLAMP, we learn each prior distribution from the previous frames up to the target frame, $p_{\bpsi_s}\left(\bz_t^s \mid \bx_{1:t-1}\right)$ and $p_{\bpsi_d}\left(\bz_t^d \mid \bx_{1:t-1}\right)$. Note that we learn separate posterior and prior distributions for the static and the dynamic components. The static contains depth and pose encoding, while the dynamic contains the flow encoding conditioned on the static as explained next.

\subsection{SLAMP-3D}
\begin{figure}[t!]
\includegraphics[width=\linewidth]{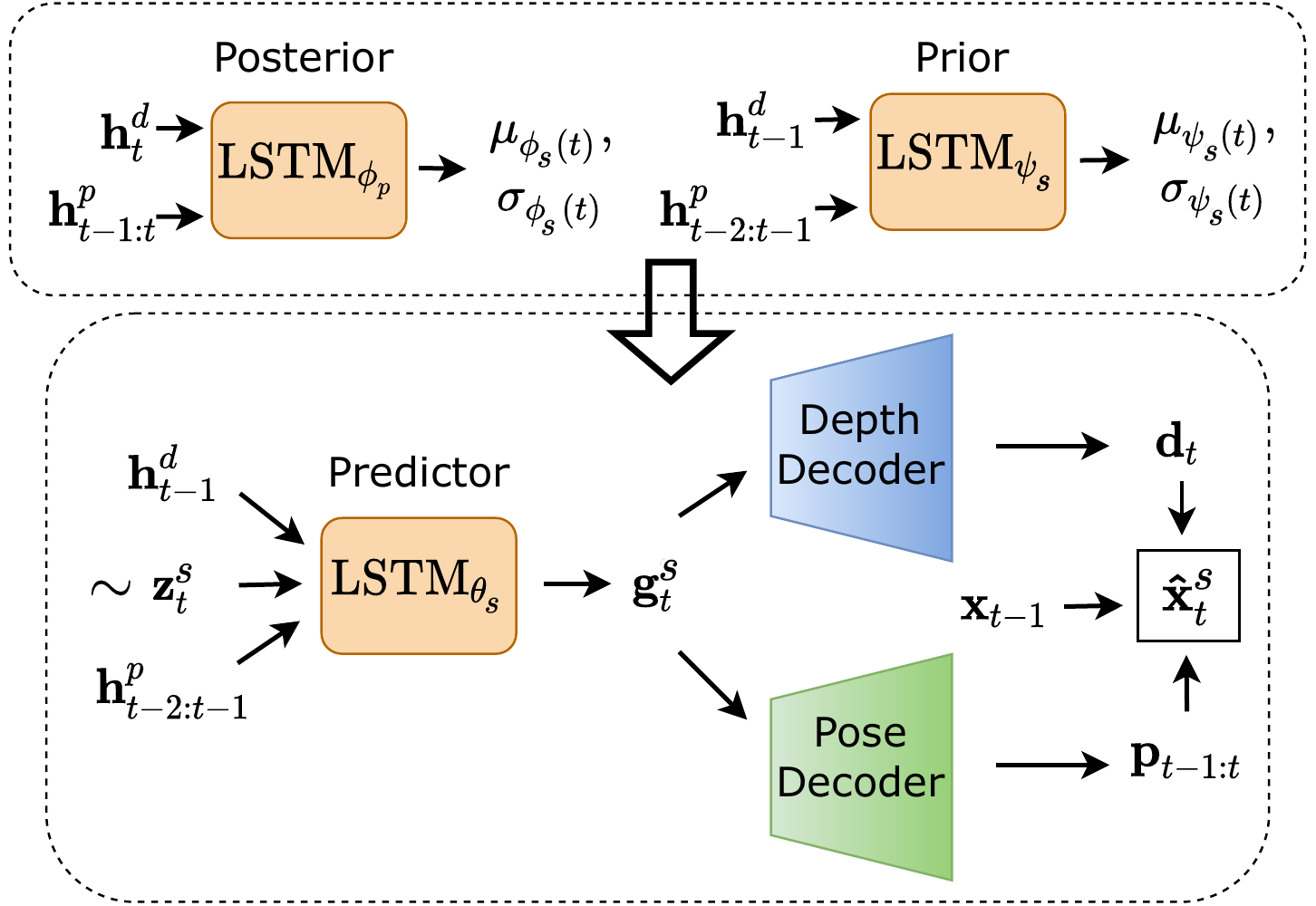}
\caption[SLAMP-3D Static Architecture]{{\bf Static Architecture.} 
The architecture shows the sampling of the static latent variable $\bz_t^s$ from the posterior $\bphi_s$ or the prior $\bpsi_s$, which is then used to predict $\bg_t^s$, the future representation of the static part. The future depth $\bd_t$ and pose $\bp_{t-1:t}$ are decoded and used to generate the static prediction $\hat{\bx}_t^s$. The architecture shows the similar procedure for the dynamic part using optical flow with the flow decoder instead of depth and pose.}
\label{fig:static_model_details}
\end{figure}
\begin{figure}[t!]
\includegraphics[width=\linewidth]{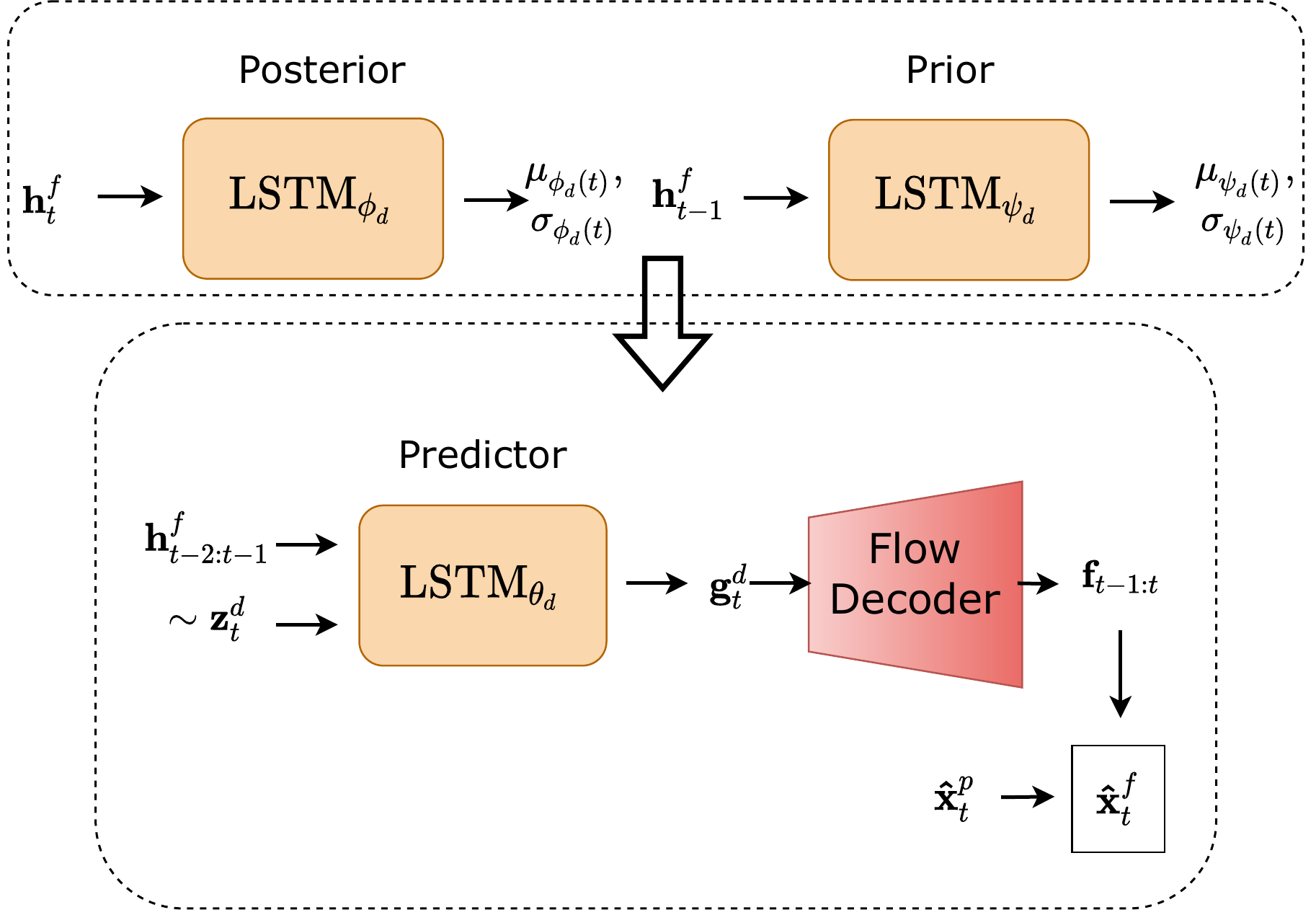}
\caption[SLAMP-3D Dynamic Architecture]{{\bf Dynamic Architecture.} 
The architecture in shows the sampling of the static latent variable $\bz_t^d$ from the posterior $\bphi_d$ or the prior $\bpsi_d$, which is then used to predict $\bg_t^d$, the future representation of the dynamic part. The future optical flow $\bff_{t-1:t}$ is decoded and used to generate the dynamic prediction $\hat{\bx}_t^d$.}
\label{fig:dynamic_model_details}
\end{figure}
Inspired by the previous work on unsupervised learning of depth and ego-motion~\cite{Zhou2017CVPR, Godard2019ICCV, Casser2019AAAI, Guizilini2020CVPR, Safadoust2021THREEDV}, flow~\cite{Jason2016ECCV, Ren2017AAAI, Meister2018AAAI}, as well as the relating of the two~\cite{Zou2018ECCV, Yin2018CVPR, Ranjan2019CVPR, Chen2019ICCV}, we reconstruct the target frame from the static and the dynamic components at each time-step.

\boldparagraph{Depth and Pose} For the static component which moves only according to the camera motion, we estimate the relative camera motion or the camera pose $\bp_{t-1:t}$ from the previous frame $t-1$ to the target frame $t$. The pose $\bp_{t-1:t}$ represents the 6-degrees of freedom rigid motion of the camera from the previous frame to the target frame. We also estimate the depth of the target frame, $\bd_{t}$, and reconstruct the target frame as $\hat{\bx}_t^s$ from the previous frame $\bx_{t-1}$ using the estimated pose $\bp_{t-1:t}$ and depth $\bd_{t}$ via differentiable warping~\cite{Jaderberg2015NeurIPS}. Note that we predict the depth and pose a priori, without actually seeing the target frame in contrast to the previous unsupervised monocular depth approaches that use the target frame to predict depth and pose~\cite{Zhou2017CVPR, Godard2019ICCV, Casser2019AAAI, Guizilini2020CVPR}.

\boldparagraph{Residual Flow} We estimate optical flow $\bff_{t-1:t}$ as the remaining motion from the reconstruction of the target frame $\hat{\bx}_t^s$ to the target frame $\bx_t$. The flow $\bff_{t-1:t}$ represents the motion of the pixels belonging to objects that move independently from the previous frame to the target frame. We reconstruct the target frame as $\hat{\bx}_t^d$ from the static prediction, $\hat{\bx}^s_t$, by using the estimated optical flow $\bff_{t-1:t}$.
Explicit conditioning of the optical flow on the static prediction allows the network to learn the remaining motion in the scene from the source frame to the target frame which corresponds to the motion that cannot be explained by the camera pose. As in the case of depth and pose prediction, we predict the flow from the reference frame to the target frame without actually seeing the target frame. In other words, we predict future motion based on the motion history, which is different than regular optical flow approaches that use the target frame for prediction~\cite{Jason2016ECCV, Ren2017AAAI, Meister2018AAAI}.

Note that the role of optical flow in SLAMP-3D is different than SLAMP. In SLAMP-3D, the optical flow is predicted conditioned on the static component. This way, the optical flow in SLAMP-3D corresponds to the residual flow as opposed to full flow as in the case of SLAMP. Therefore, we warp the \emph{static component's prediction} instead of the previous frame as in the case of SLAMP.

\boldparagraph{Combining Static and Dynamic Predictions} Similar to SLAMP, we use the Hadamard product to combine the static prediction $\hat{\bx}_t^s$ and the dynamic prediction $\hat{\bx}_t^d$ into the final prediction $\hat{\bx}_t$, showed in \eqref{eq:slamp3d_mask}.
The static parts in the background can be reconstructed accurately using the depth and camera pose estimation. For the dynamic parts with moving objects, the target frame can be reconstructed accurately using the residual motion prediction.
The mask prediction learns a weighting between the static and the dynamic predictions for combining them into the final prediction. 

\begin{ceqn}
\begin{align}
    \label{eq:slamp3d_mask}
   \hat{\bx}_t = \bm(\bx_t^s,\bx_t^d) \odot \bx_t^s + (\mathbf{1} - \bm(\bx_t^s,\bx_t^d)) \odot \bx_t^d
\end{align}
\end{ceqn}

\subsection{Variational Inference}
\label{sec:inf}
Removing time dependence for clarity, \eqnref{eq:cond_prob} expresses the conditional joint probability corresponding to the graphical model shown in \figref{fig:graphical}:
\begin{ceqn}
\begin{equation}
    \label{eq:cond_prob}
    p_{\btheta}(\bx) = \iint p(\bx \mid \bz^s, \bz^d)~p(\bz^s)~p(\bz^d) ~\mathrm{d}\bz^s ~\mathrm{d}\bz^d
\end{equation}
\end{ceqn}
The true distribution over the latent variables $\bz_{t}^{s}$ and $\bz_{t}^{d}$ is intractable. We assume that the dynamic component $\bz_{t}^{d}$ depends on the static component $\bz_{t}^{s}$ and train time-dependent inference networks $q_{\bphi_s}\left(\bz_{t}^{s} \mid \bx_{1:T}\right)$ and $q_{\bphi_d}\left(\bz_{t}^{d} \mid \bx_{1:T}, \bz_{t}^{s}\right)$ to approximate the true distribution with conditional Gaussian distributions. 
In order to optimize the likelihood of $p\left(\bx_{1:T}\right)$, we need to infer latent variables $\bz_{t}^{s}$ and $\bz_{t}^{d}$, which correspond to uncertainty in static and dynamic parts of future frames, respectively. We use a variational inference model to infer the latent variables.

At each time-step, $\bz_{t}^{d}$ depends on $\bz_{t}^s$ but each is independent across time. Therefore, we can decompose Kullback-Leibler terms into individual time steps. We train the model by optimizing the variational lower bound as shown in \eqref{eq:elbo} where each $\bz_{1:t}^\cdot$ is sampled from the respective posterior distribution $q_{\bphi_{\cdot}}$~(see Appendix for the derivation). 

\begin{align}
\label{eq:elbo}
\log p_{\btheta}(\bx) \geq 
&\mathbb{E}_{q_s(\bz^s \mid \bx)} \left[ \mathbb{E}_{q_d(\bz^d \mid \bz^s, \bx)}\left[ \log p(\bx \mid \bz^s, \bz^d)\right] \right] \\ 
&- \mathbb{E}_{q_s(\bz^s \mid \bx)} \left[D_{KL}(q_d(\bz^d \mid \bz^s, \bx) \mid\mid p(\bz^d)) \right] - D_{KL}(q_s(\bz^s \mid \bx) \mid\mid p(\bz^s)) \nonumber
\end{align}

The likelihood $p_{\btheta}$ can be interpreted as minimizing the difference between the actual frame $\bx_t$ and the prediction $\hat{\bx}_t$ as defined in \eqnref{eq:combined_x} by changing $p$ and $f$ to $s$ and $d$. We apply the reconstruction loss to the predictions of static and dynamic components as well.
The posterior terms for uncertainty are estimated as an expectation over $q_{\bphi_s}\left(\bz_t^s \mid \bx_{1:t}\right)$ and $q_{\bphi_d}\left(\bz_t^d \mid \bx_{1:t}, \bz_t^s\right)$.
For the second term, we use sampling to approximate the expectation as proposed in \cite{Detlefsen2019NeurIPS}.
As in \cite{Denton2018ICML}, we also learn the prior distributions from the previous frames up to the target frame as 
$p_{\bpsi_s}\left(\bz_t^s \mid \bx_{1:t-1}\right)$ and $p_{\bpsi_d}\left(\bz_t^d \mid \bx_{1:t-1}\right)$.
We train the model using the re-parameterization trick \cite{Kingma2014ICLR} and choose the posteriors to be factorized Gaussian so that all the KL divergences can be computed analytically.
We apply the optimal variance estimate as proposed in $\sigma$-VAE~\cite{Rybkin2021ICML} to learn an optimal value for the hyper-parameter corresponding to the weight of the KL term. 

\subsection{Architecture}
\label{sec:arch}
We encode the frames with a feed-forward convolutional architecture to obtain frame-wise features at each time-step. Our goal is to reduce the spatial resolution of the frames in order to reduce the complexity of the model.  
On top of this shared representation, we have a depth head and a pose head to learn an encoding of depth at each frame, $\bh_{t}^d$, and an encoding of pose for each consecutive frame pair, $\bh_{t-1:t}^p$.
As shown in \figref{fig:static_model_details}, we train convolutional LSTMs to infer the static distribution at each time-step from the encoding of depth and pose:

\begin{ceqn}
\begin{align}
    \label{eq:static_dist}
    &\be_{t-1},~\be_t = \mathrm{ImgEnc}\left(\bx_{t-1}\right),~ \mathrm{ImgEnc}\left(\bx_t\right) \\
    &\bh_t^d,~\bh_{t-1:t}^p = \mathrm{DepthEnc}\left(\be_t\right),~ ~\mathrm{PoseEnc}\left(\be_{t-1}, \be_t\right) \nonumber \\
    &\bz_t^s \sim \mathcal{N}\left(\bmu_{\bphi_s(t)}, \bsigma_{\bphi_s(t)}\right) \quad \text{where} \nonumber \\ &\bmu_{\bphi_s(t)},~ \bsigma_{\bphi_s(t)} = \mathrm{LSTM}_{\bphi_s}\left(\bh_t^d,~ \bh_{t-1:t}^p\right) \nonumber
     \nonumber
\end{align}
\end{ceqn}

\eqnref{eq:static_dist} only shows the posterior distribution, similar steps are followed for the static prior by using the depth and pose representations from the previous time-step.

At the first time-step where there is no previous pose estimation, we assume zero-motion by estimating the pose from the previous frame to itself.
We learn a static frame predictor to generate the target frame based on the encoding of depth and pose and the static latent variable $\bz_t^s$: 
\begin{ceqn}
\begin{equation}
    \label{eq:static_frame_pred}
    \bg_t^s = \mathrm{LSTM}_{\btheta_s}\left(\bh_{t-1}^d, \bh_{t-2:t-1}^p, \bz_t^s\right)
\end{equation}
\end{ceqn}
Then, the depth and pose estimations are decoded from $\bg_t^s$. 
Based on the depth and pose estimations, the static prediction $\hat{\bx}_t^s$ is reconstructed by inverse warping the previous frame $\bx_{t-1}$. 

We learn an encoding of the remaining motion, $\bh_{t}^f$, in the dynamic parts from the output of the static frame predictor $\bg_{t}^s$ to the encoding of the target frame $\bx_t$.
We train convolutional LSTMs to infer dynamic posterior and prior distributions at each time-step from the encoded residual motion.
The posterior LSTM is updated based on the $\bh_{t}^f$ and the latent variable $\bz_t^d$ is sampled from the posterior:
\begin{ceqn}
\begin{align}
    &\bh_t^f = \mathrm{MotionEnc}\left(\bg_t^s, \be_t\right) \\
    &\bz_t^d \sim \mathcal{N}\left(\bmu_{\bphi_d(t)}, \bsigma_{\bphi_d(t)}\right) \quad \text{where} \nonumber \\ &\bmu_{\bphi_d(t)},~ \bsigma_{\bphi_d(t)} = \mathrm{LSTM}_{\bphi_d}\left(\bh_t^f\right) \nonumber
\end{align}
\end{ceqn}

For the dynamic prior, we use the motion representation from the previous time step to update the prior LSTM and sample the $\bz_t^d$ from it.
Similar to the pose above, we assume zero-motion at the first time-step where there is no previous motion.
The dynamic predictor LSTM is updated according to encoded features and sampled latent variables:
\begin{ceqn}
\begin{equation}
    \bg_t^d = \mathrm{LSTM}_{\btheta_d}\left(\bh_{t-1}^f, \bz_t^d\right)
\end{equation}
\end{ceqn}
During training, the latent variables are sampled from the posterior distribution. At test time, they are sampled from the posterior for the conditioning frames and from the prior for the following frames. 

After we generate target time's features for static and dynamic components, $\bg_t^s$ and $\bg_t^d$, respectively, we decode them into depth and pose for the static component, optical flow for the dynamic component. We first predict the depth values of the target frame without seeing the frame and the pose from the previous frame to the target frame using the static features, $\bg_t^s$. Then, we warp the previous frame $\bx_{t-1}$ using the predicted depth and pose to obtain the static prediction for the target frame, $\hat{\bx}_t^s$. For the dynamic part, we predict the residual flow from the dynamic features, $\bg_t^d$. Then, we warp the static frame prediction $\hat{\bx}_t^s$, using the predicted residual flow to obtain the dynamic prediction, $\hat{\bx}_t^d$. Finally, static and dynamic predictions are combined into the final prediction, $\hat{\bx}_t$, using the predicted mask as shown in \eqref{eq:slamp3d_mask}.

\section{Experiments}

\boldparagraph{Implementation Details}
We first process each image with a shared backbone~\cite{Simonyan2015ICLR, He2016CVPR}
to reduce the spatial resolution.
We add separate heads for extracting features for depth, pose, and flow. We learn the static distribution based on depth and pose features, and dynamic distribution based on flow features with two separate ConvLSTMs. Based on the corresponding latent variables, we learn to predict the next frame's static and dynamic representation with another pair of ConvLSTMs. 
We decode depth and pose from the static representation and flow from the dynamic. 
We obtain the static frame prediction by warping the previous frame according to depth and pose, and dynamic prediction by warping the static prediction according to flow. We train the model using the negative log likelihood loss. 

\boldparagraph{Datasets} 

We perform experiments on two challenging autonomous driving datasets, KITTI~\cite{Geiger2012CVPR, Geiger2013IJRR} and Cityscapes~\cite{Cordts2016CVPR}. 
Both datasets contain everyday real-world scenes with complex dynamics due to both background and foreground motion. 
We train our model on the training set of the Eigen split on KITTI~\cite{Eigen2014NeurIPS}. We apply the pre-processing followed by monocular depth approaches and remove the static scenes~\cite{Zhou2017CVPR}. In addition, we use the depth ground-truth on KITTI to validate our depth predictions.
Cityscapes primarily focuses on semantic understanding of urban street scenes, therefore contains a larger number of dynamic foreground objects compared to KITTI. However, motion lengths are larger on KITTI due to lower frame-rate. 
On both datasets, we condition on 10 frames and predict 10 frames into the future to train our models. Then, at test time, we predict 20 frames conditioned on 10 frames.

\boldparagraph{Evaluation Metrics}
We compare the video prediction performance using four different metrics:
Peak Signal-to-Noise Ratio (PSNR) based on the $L_2$ distance between the frames penalizes differences in dynamics but also favors blurry predictions.
Structured Similarity (SSIM) compares local patches to measure similarity in structure spatially. %
Learned Perceptual Image Patch Similarity (LPIPS)~\cite{Zhang2018CVPR} measures the distance between learned features extracted by a CNN trained for image classification.
Frechet Video Distance (FVD) \cite{Unterthiner2019ARXIV}, lower better, compares temporal dynamics of generated videos to the ground truth in terms of representations computed for action recognition.

\subsection{Ablation Study}
\begin{table}[h!]
    \sisetup{detect-weight, table-align-uncertainty=true, mode=text}
    \renewrobustcmd{\bfseries}{\fontseries{b}\selectfont}
    \renewrobustcmd{\boldmath}{}
    \centering
    \caption[SLAMP-3D Baselines]{
        \textbf{Baselines.} Results of our method~(\emph{Conditional}) by removing the dynamic part~(\emph{Depth-Only}) and by removing the conditioning of the dynamic part on the static~(\emph{Combined}).}
    \label{tab:ablation}
    \resizebox{0.75\linewidth}{!}{\begin{tabular}{lccc}
        \toprule
        Models &  PSNR ($\uparrow$) & SSIM ($\uparrow$) & LPIPS ($\downarrow$) \tabularnewline 
        \midrule
        Depth-Only & 13.02 $\pm$ 0.44 & 0.301 $\pm$ 0.021 & 0.523 $\pm$ 0.014 
        \tabularnewline
        Combined  & \bfseries 14.45 $\pm$ 0.35 &   0.378 $\pm$ 0.019 & 0.533 $\pm$ 0.016 \tabularnewline
        Conditional &  14.32 $\pm$ 0.33 & \bfseries 0.383 $\pm$ 0.020 & \bfseries 0.501 $\pm$ 0.016
        \tabularnewline
        \bottomrule
    \end{tabular}
    }
    \vspace{1mm}
    \caption*{KITTI~\cite{Geiger2012CVPR, Geiger2013IJRR}}
    \vspace{2mm}
    \resizebox{0.75\linewidth}{!}{\begin{tabular}{lccc}
        \toprule
        Models &  PSNR ($\uparrow$) & SSIM ($\uparrow$) & LPIPS ($\downarrow$) \tabularnewline 
        \midrule
        Depth-Only & 19.97 $\pm$ 0.48 & 0.580 $\pm$ 0.016 & 0.445 $\pm$ 0.014
        \tabularnewline
        Combined  & 21.00 $\pm$ 0.41 & 0.631 $\pm$ 0.014 & 0.309 $\pm$ 0.009 \tabularnewline
        Conditional & \bfseries 21.43 $\pm$ 0.43 & \bfseries 0.643 $\pm$ 0.014 & \bfseries 0.306 $\pm$ 0.009
        \tabularnewline       \bottomrule
    \end{tabular}
    }
    \vspace{1mm}
    \caption*{Cityscapes~\cite{Cordts2016CVPR}}
\end{table}

We first examine the importance of each contribution on KITTI and Cityscapes in \tabref{tab:ablation}. We start with a simplified version of our model called \emph{Depth-Only} by removing the dynamic latent variables and the flow decoder. This model inherits the weakness of depth and ego-motion estimation methods by assuming a completely static scene. Therefore, it cannot model the motion of the dynamic objects but it still performs reasonably well, especially on KITTI, since the background typically covers a large portion of the image. In the next two models, we include dynamic latent variables. 
We evaluate the importance of conditioning of dynamic variables on the static, \emph{Combined} versus \emph{Conditional}. For the \emph{Combined}, we independently model static and dynamic latent variables and simply combine them in the end with the predicted mask. For the \emph{Conditional}, we conditioned the dynamic latent variable on the static latent variable. First, both models improve the results compared to the \emph{Depth-Only} case. This confirms our intuition about modelling the two types of motion in the scene separately. The two perform similarly on KITTI but the \emph{Conditional} outperforms the \emph{Combined} on Cityscapes due to larger number of moving foreground objects.

\subsection{Quantitative Results}
We trained and evaluated stochastic video prediction methods on KITTI and Cityscapes by using the same number of conditioning frames including SVG~\cite{Denton2018ICML}, SRVP~\cite{Franceschi2020ICML}, SLAMP, and Improved-VRNN~\cite{Castrejon2019ICCV}. We optimized their architectures for a fair comparison~(see Appendix).

\begin{table*}[t!]
    \caption[Quantitative Results on KITTI.]{
        \label{tab:results_kitti}
        \textbf{Quantitative Results on KITTI.} We compare our \emph{Combined} and \emph{Conditional} models to the other video prediction approaches on KITTI~\cite{Geiger2012CVPR, Geiger2013IJRR}. The best results are shown in bold and the second best underlined.
    }
    \sisetup{detect-weight, table-align-uncertainty=true, mode=text}
    \renewrobustcmd{\bfseries}{\fontseries{b}\selectfont}
    \renewrobustcmd{\boldmath}{}
    \centering
    \resizebox{0.9\textwidth}{!}{\begin{tabular}{lccc}%
        \toprule
        Models & PSNR ($\uparrow$) & SSIM ($\uparrow$) & LPIPS ($\downarrow$) \tabularnewline
        \midrule
        SVG~\cite{Denton2018ICML} & 12.70 $\pm$ 0.70 & 0.329 $\pm$ 0.030 &  0.594 $\pm$ 0.034 \tabularnewline
        SRVP~\cite{Franceschi2020ICML}  & 13.41 $\pm$ 0.42 & 0.336 $\pm$ 0.034 & 0.635 $\pm$ 0.021 \tabularnewline
        SLAMP & 13.46 $\pm$ 0.74 & 0.337 $\pm$ 0.034 & 0.537 $\pm$ 0.042 \tabularnewline
        Improved-VRNN~\cite{Castrejon2019ICCV} &  14.15 $\pm$ 0.47 & \underline{0.379} $\pm$ 0.023 & \bfseries 0.372 $\pm$ 0.020 \tabularnewline
        Ours-Combined & \bfseries 14.45 $\pm$ 0.35 &  0.378 $\pm$ 0.019 &  0.533 $\pm$ 0.016 \tabularnewline
        Ours-Conditional &  \underline{14.32} $\pm$ 0.33 & \bfseries 0.383 $\pm$ 0.020 &  0.501 $\pm$ 0.016 \tabularnewline
        \bottomrule
    \end{tabular}}
\end{table*}

\begin{table*}[t!]
    \caption[Quantitative Results on Cityscapes.]{
        \label{tab:results_city}
        \textbf{Quantitative Results on Cityscapes.} We compare our \emph{Combined} and \emph{Conditional} models to the other video prediction approaches on Cityscapes~\cite{Cordts2016CVPR}. The best results are shown in bold and the second best underlined.
    }
    \sisetup{detect-weight, table-align-uncertainty=true, mode=text}
    \renewrobustcmd{\bfseries}{\fontseries{b}\selectfont}
    \renewrobustcmd{\boldmath}{}
    \centering
    \resizebox{0.9\textwidth}{!}{\begin{tabular}{lccc}%
        \toprule
        Models & PSNR ($\uparrow$) & SSIM ($\uparrow$) & LPIPS ($\downarrow$) \tabularnewline
        \midrule
        SVG~\cite{Denton2018ICML} & 20.42 $\pm$ 0.63 & 0.606 $\pm$ 0.023 & 0.340 $\pm$ 0.022 \tabularnewline
        SRVP~\cite{Franceschi2020ICML}& 20.97 $\pm$ 0.43 & 0.603 $\pm$ 0.016 & 0.447 $\pm$ 0.014 \tabularnewline
        SLAMP~\cite{Akan2021ICCV}& \bfseries 21.73 $\pm$ 0.76 & \bfseries 0.649 $\pm$ 0.025 & \underline{0.294} $\pm$ 0.022 \tabularnewline
        Improved-VRNN~\cite{Castrejon2019ICCV} & 21.42 $\pm$ 0.67 &  {0.618} $\pm$ 0.020 & \bfseries 0.260 $\pm$ 0.014 \tabularnewline
        Ours-Combined & 21.00 $\pm$ 0.41 &  0.631 $\pm$ 0.014 & 0.309 $\pm$ 0.009 \tabularnewline
        Ours-Conditional &  \underline{21.43} $\pm$ 0.43 & \underline{0.643} $\pm$ 0.014 & {0.306} $\pm$ 0.009 \tabularnewline
        \bottomrule
    \end{tabular}}
\end{table*}

\boldparagraph{Frame-Level Evaluations}
We report the frame-level evaluation results on \tabref{tab:results_kitti} and \tabref{tab:results_city}.Recent methods including SLAMP, Improved-VRNN~\cite{Castrejon2019ICCV}, and our models clearly outperform SVG~\cite{Denton2018ICML} and SRVP~\cite{Franceschi2020ICML} in terms of all three metrics. Improved-VRNN \cite{Castrejon2019ICCV} \footnote{The results of Improved-VRNN on Cityscapes is different from the ones in the original paper since we retrained their model on a similar resolution to ours with the same number of conditioning frames as ours.} achieves that by using five times more parameters compared to our models~(57M vs. 308M). A recent study shows the importance of attacking the under-fitting issue in video prediction~\cite{Babaeizadeh2021Arxiv}. The results can be improved by over-parameterizing the model and using data augmentation to prevent over-fitting. This finding
is complementary to other approaches including ours, however, it comes at a cost in terms of run-time. The time required to generate 10 samples is 40~seconds for Improved-RNN compared to 1~second, or significantly less in case of SRVP, for other methods. %

SLAMP achieves the top-performing results on Cityscapes in terms of PSNR and SSIM by explicitly modelling the motion history. The separation between the pixel and the motion space is achieved by learning a separate distribution for each with a slight increase in complexity compared to SVG. We follow a similar decomposition as static and dynamic but we differentiate between the two types of motion in driving. As a result, our models clearly outperform SLAMP on KITTI where the camera motion is large due to low frame-rates. In addition, our models can produce reliable depth predictions for the static part of the scene (\tabref{tab:depth_eval}).
In SLAMP, we report results on generic video prediction datasets such as MNIST, KTH~\cite{Schuldt2004CVPR} and BAIR~\cite{EbertFLL17} as well. 
In SLAMP-3D, we focus on driving scenarios by utilizing the domain knowledge for a better modelling of the static part. In case of SLAMP, the pixel decoder only focuses on cases that cannot be handled by the flow decoder, \eg occlusions. Whereas in our case, the static models the whole background, leading not only to a better segmentation of the background in the mask but also to better results in the background on KITTI, as shown in \tabref{tab:fg_bg_results_kitti}.

\begin{table}[h!]
    \captionsetup[table]{justification=centering}
    \renewrobustcmd{\bfseries}{\fontseries{b}\selectfont}
    \renewrobustcmd{\boldmath}{}
    \caption[SLAMP-3D FVD Scores]{
        \label{tab:fvd}
        \textbf{FVD Scores on KITTI and Cityscapes.} }
    \centering
    \small
    \vspace{-2mm}
    \begin{tabular}{lcc}
        \toprule
        Dataset & KITTI & Cityscapes \tabularnewline
        \midrule
        SVG~\cite{Denton2018ICML} & 1733 $\pm$ 198 & 870 $\pm$ 95 \tabularnewline
        SRVP~\cite{Franceschi2020ICML} & 1792 $\pm$ 190 &  1409 $\pm$ 138 \tabularnewline
        SLAMP & 1585 $\pm$ 154 & 796 $\pm$ 89 \tabularnewline
        VRNN~\cite{Castrejon2019ICCV} & \bfseries 1022 $\pm$ 145 & \bfseries 658 $\pm$ 80 \tabularnewline
        Ours-Combined & 1463 $\pm$ 186 & 793 $\pm$ 86 \tabularnewline
        Ours-Conditional. & \underline{1297 $\pm$ 142} & \underline{789 $\pm$ 84} \tabularnewline
    \end{tabular}
    \vspace{-3mm}
\end{table}

\boldparagraph{Video-Level Evaluations} We also use a video level evaluation metric, FVD~\cite{Unterthiner2019ARXIV}, for a video-level comparison. According to the results in \tabref{tab:fvd}, VRNN performs the best in terms of FVD. Our models, both combined and conditional, outperform all the other methods and approach the performance of VRNN with a nearly $40$ times shorter inference time. %

\begin{sidewaysfigure}[h!]
\centering
\includegraphics[width=\textwidth]{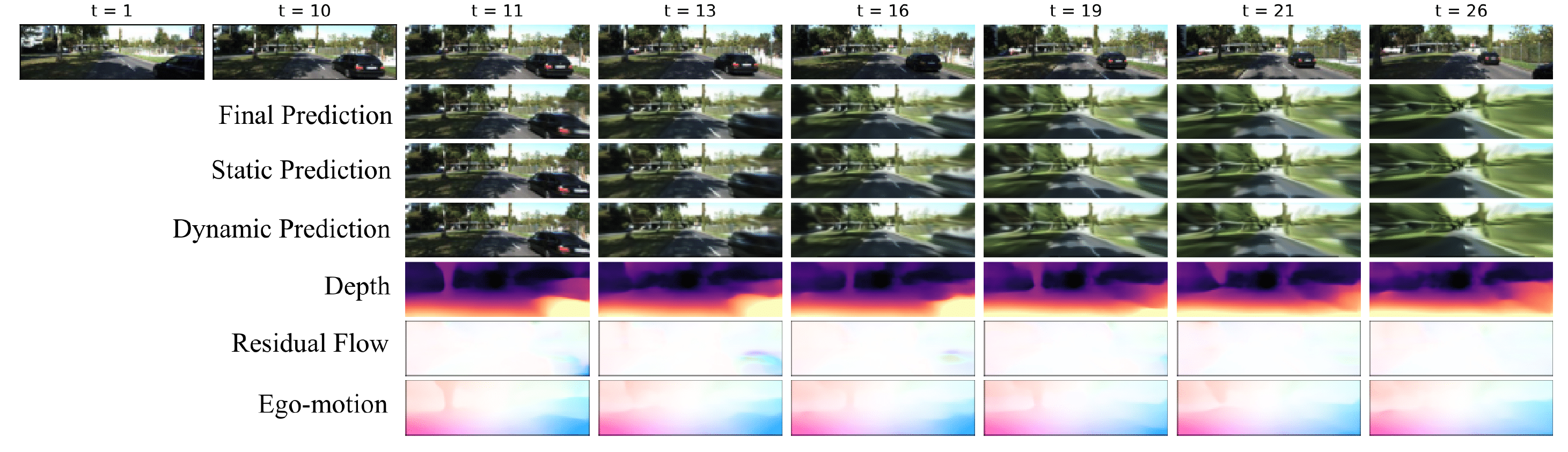}
\caption[Conditional Model on KITTI]{\textbf{Conditional Model on KITTI.}
The top row shows the ground truth, followed by the frame predictions by the final, the static, and the dynamic. The last three rows show depth, residual flow, and flow due to ego-motion.
The structure and camera motion can explain the static part of the scene and the independent motion of the moving car on the left is captured by the residual flow.
}
\label{fig:qual_detailed}
\end{sidewaysfigure}

\begin{table*}[h!]
    \caption[Foreground and Background Evaluations on KITTI.]{
        \label{tab:fg_bg_results_kitti}
        \textbf{Foreground and Background Evaluations on KITTI.} We use off-the-shelf semantic segmentation network to identify the foreground regions, mask all the other regions and compute the metrics separately in the foreground and the background.}
    \sisetup{detect-weight, table-align-uncertainty=true, mode=text}
    \renewrobustcmd{\bfseries}{\fontseries{b}\selectfont}
    \renewrobustcmd{\boldmath}{}
    \centering
    \vspace{0.1in}
    \resizebox{\textwidth}{!}{\begin{tabular}{clccc}%
        \toprule
        {} &  Models & PSNR ($\uparrow$) & SSIM ($\uparrow$) & LPIPS ($\downarrow$)  \tabularnewline 
        \midrule
        \multirow{6}{*}{\rotatebox[origin=c]{90}{\textbf{Foreground}}}
        {} & SVG \cite{Denton2018ICML}  & 26.47 $\pm$ 0.31 & 0.9147 $\pm$ 0.0026 &  0.1183 $\pm$ 0.0027 \tabularnewline
        {} & SRVP \cite{Franceschi2020ICML} & \underline{27.28} $\pm$ 0.31 & 0.9162 $\pm$ 0.0026 & 0.1244 $\pm$ 0.0029\tabularnewline
        {} & SLAMP &  26.93 $\pm$ 0.32 &  0.9154 $\pm$ 0.0026 &  0.1125 $\pm$ 0.0026  \tabularnewline
        {} & Improved-VRNN \cite{Castrejon2019ICCV} & 26.86 $\pm$ 0.27 & \underline{0.9192} $\pm$ 0.0025 & \bfseries 0.0877 $\pm$ 0.0020  \tabularnewline
        {} & Ours-Combined & \bfseries 27.44 $\pm$ 0.29 &  \bfseries 0.9193 $\pm$ 0.0025 &  0.1093 $\pm$ 0.0024  \tabularnewline
        {} & Ours-Conditional & 26.80 $\pm$ 0.29 & 0.9161 $\pm$ 0.0026 & \underline{0.1046} $\pm$ 0.0023 \tabularnewline
        \midrule \midrule
        \multirow{6}{*}{\rotatebox[origin=c]{90}{\textbf{Background}}}
        {} & SVG \cite{Denton2018ICML} & 13.30 $\pm$ 0.07 & 0.4003 $\pm$ 0.0034 & 0.5227 $\pm$ 0.0031\tabularnewline
        {} & SRVP \cite{Franceschi2020ICML} & 13.97 $\pm$ 0.06 & 0.4080 $\pm$ 0.0033 & 0.5498 $\pm$ 0.0028 \tabularnewline
        {} & SLAMP  & 14.09 $\pm$ 0.08 & 0.4089 $\pm$ 0.0032 & 0.4707 $\pm$ 0.0029 \tabularnewline
        {} & Improved-VRNN \cite{Castrejon2019ICCV} &  14.82 $\pm$ 0.07 &  0.4417 $\pm$ 0.0037 &  \bfseries 0.3380 $\pm$ 0.0031    \tabularnewline
        {} & Ours-Combined & \bfseries 15.11 $\pm$ 0.06 & \underline{0.4455} $\pm$ 0.0032 & 0.4712 $\pm$ 0.0028 \tabularnewline
        {} & Ours-Conditional & \underline{15.02} $\pm$ 0.05 & \bfseries 0.4507 $\pm$ 0.0031 &  \underline{0.4451} $\pm$ 0.0028 \tabularnewline
        \bottomrule
    \end{tabular}}
\end{table*}

\begin{table*}[h!]
    \caption[Foreground and Background Evaluations on Cityscapes.]{
        \label{tab:fg_bg_results_city}
        \textbf{Foreground and Background Evaluations on Cityscapes.} We use off-the-shelf semantic segmentation network to identify the foreground regions, mask all the other regions and compute the metrics separately in the foreground and the background.}
    \sisetup{detect-weight, table-align-uncertainty=true, mode=text}
    \renewrobustcmd{\bfseries}{\fontseries{b}\selectfont}
    \renewrobustcmd{\boldmath}{}
    \centering
    \vspace{0.1in}
    \resizebox{\textwidth}{!}{\begin{tabular}{clccc}%
        \toprule
        {} &  Models & PSNR ($\uparrow$) & SSIM ($\uparrow$) & LPIPS ($\downarrow$)  \tabularnewline 
        \midrule
        \multirow{6}{*}{\rotatebox[origin=c]{90}{\textbf{Foreground}}}
        {} & SVG \cite{Denton2018ICML} & 30.63 $\pm$ 0.14 & 0.9483 $\pm$ 0.0006 &  0.0676 $\pm$ 0.0005 \tabularnewline
        {} & SRVP \cite{Franceschi2020ICML}  & 30.85 $\pm$ 0.14 & 0.9474 $\pm$ 0.0006 & 0.0763 $\pm$ 0.0006 \tabularnewline
        {} & SLAMP &  \underline{31.71} $\pm$ 0.15 &  \underline{0.9536} $\pm$ 0.0005 &  0.0577 $\pm$ 0.0005 \tabularnewline
        {} & Improved-VRNN \cite{Castrejon2019ICCV} & 30.65 $\pm$ 0.14 & 0.9495 $\pm$ 0.0006 & \bfseries 0.0554 $\pm$ 0.0004   \tabularnewline
        {} & Ours-Combined & 31.52 $\pm$ 0.14 & \underline{0.9536} $\pm$ 0.0005 &  \underline{0.0568} $\pm$ 0.0004 \tabularnewline
        {} & Ours-Conditional & \bfseries 31.77 $\pm$ 0.14 &  \bfseries 0.9542 $\pm$ 0.0005 & 0.0579 $\pm$ 0.0005 \tabularnewline
        \midrule \midrule
        \multirow{6}{*}{\rotatebox[origin=c]{90}{\textbf{Background}}}
        {} & SVG \cite{Denton2018ICML} & 21.24 $\pm$ 0.04 & 0.6699 $\pm$ 0.0014 & 0.2620 $\pm$ 0.0011 \tabularnewline
        {} & SRVP \cite{Franceschi2020ICML} & 21.96 $\pm$ 0.04 & 0.6757 $\pm$ 0.0014 & 0.3358 $\pm$ 0.0012 \tabularnewline
        {} & SLAMP & \bfseries 22.66 $\pm$ 0.05 & \bfseries 0.7087 $\pm$ 0.0014 & \underline{0.2341} $\pm$ 0.0011\tabularnewline
        {} & Improved-VRNN \cite{Castrejon2019ICCV} & \underline{22.52} $\pm$ 0.07 &  0.6811 $\pm$ 0.0017 &  \bfseries 0.2155 $\pm$ 0.0013   \tabularnewline
        {} & Ours-Combined & 21.87 $\pm$ 0.04 &  0.6935 $\pm$ 0.0013 & 0.2496 $\pm$ 0.0009 \tabularnewline
        {} & Ours-Conditional &  22.36 $\pm$ 0.05 & \underline{0.7026} $\pm$ 0.0013 &  0.2426 $\pm$ 0.0009 \tabularnewline
        \bottomrule
    \end{tabular}}
\end{table*}

\boldparagraph{Foreground and Background Evaluations}
We compare the prediction performance separately in the foreground and the background regions of the scene in \tabref{tab:fg_bg_results_kitti} and \tabref{tab:fg_bg_results_city} on KITTI and Cityscapes, respectively. 
We use off-the-shelf semantic segmentation models to extract the objects in the scene and assume some of the semantic classes as the foreground. Specifically, we use the pre-trained model by \cite{Zhu2019CVPR, Reda2018ECCV}, which is the best model on KITTI semantic segmentation leaderboard with code available (the second-best overall), to obtain the masks on this dataset. On Cityscapes, we use the pre-trained model by \cite{Tao2020Arxiv} for similar reasons. We choose the following classes as the foreground objects: ``person, rider, car, truck, bus, train, motorcycle, bicycle" and consider the remaining pixels as the background.
For foreground results, we extract the foreground regions based on the segmentation result and ignore all the other pixels in the background by assigning the mean color, gray. 
We do the opposite masking for the background region by setting all of the foreground regions to gray, and calculate the metrics again. 

On Cityscapes, our \textit{Combined} and \textit{Conditional} models achieve the best results in terms of PSNR and SSIM metrics in the foreground regions. The \textit{Conditional} outperforms the \textit{Combined}, showing the importance of conditioning of dynamic latent variables on the static for modelling the independent motion of foreground objects. However, SLAMP and Improved-VRNN outperform our models in the background. Improved-VRNN's performance is especially impressive in terms of LPIPS, consistently in all regions on both datasets.
On KITTI, our \textit{Combined} model performs the best in foreground regions in terms of PSNR and SSIM. For background regions, our \textit{Combined} and \textit{Conditional} models are the two best-performing models in terms of PSNR and SSIM.
In summary, we validate our two claims; first, to better model the motion history of foreground objects on the more dynamic Cityscapes and second, to better model the larger motion in the background on KITTI due to a smaller frame rate.

\begin{sidewaysfigure}[h!]
\centering
\includegraphics[width=\textwidth]{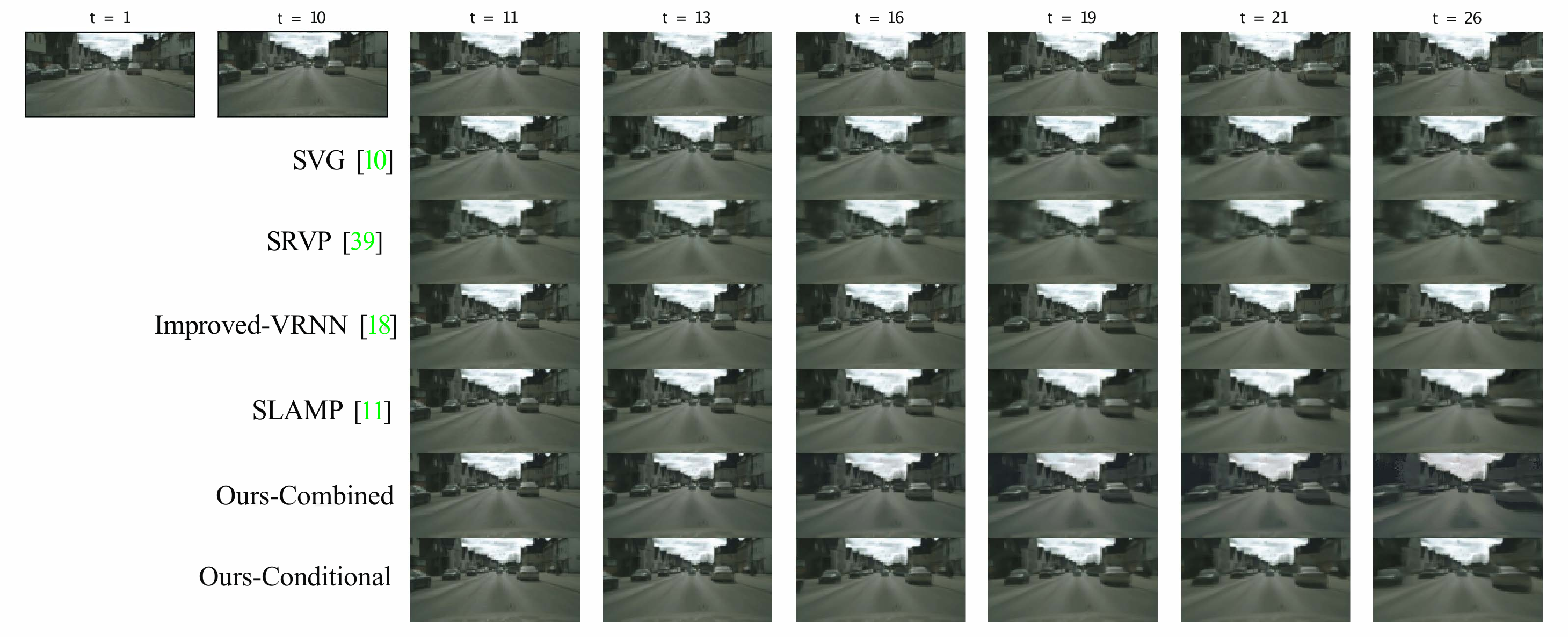}
\caption[Comparison on Cityscapes]{\textbf{Comparison to Other Methods on Cityscapes.} 
Our \emph{Conditional} model can better predict the motion of the vehicles moving on the left and on the right thanks to the separate modeling of the motion history of background and foreground regions.
}
\label{fig:qual_comp}
\end{sidewaysfigure}

\begin{sidewaysfigure}[h!]
\centering
\includegraphics[width=\textwidth]{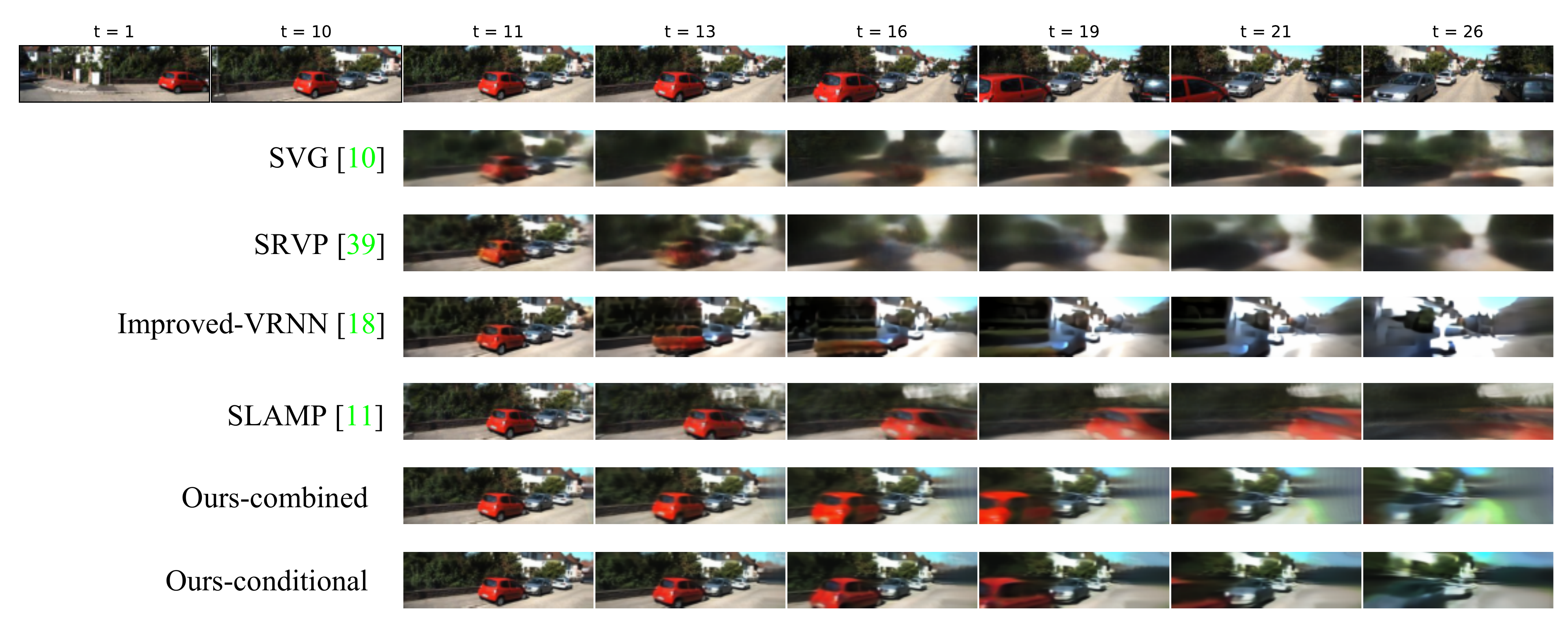}
\caption[Difficult Turning Case on KITTI]{\textbf{Difficult Turning Case on KITTI.} 
Our \emph{Conditional} and \emph{Combined} model can better predict future after the conditioning frames. Our models, especially the conditional model, can better preserve the red car in their predictions for a longer time. All of the other methods fail in the later time steps due to the difficulty of predicting future in the turning case which is not frequent in the dataset.
}
\label{fig:qual_comp_kitti}
\end{sidewaysfigure}

\boldparagraph{Evaluation of Depth Predictions}
\begin{table*}[h!]
    \caption[Evaluation of Depth Predictions]{
        \label{tab:depth_eval}
        \textbf{Evaluation of Depth Predictions.} We compare depth predictions from our \emph{Combined} and \emph{Conditional} models to the state-of-the-art unsupervised monocular depth prediction method Monodepth2~\cite{Godard2019ICCV} on KITTI. We predict future depth without seeing the target frame.
    }
    \sisetup{detect-weight, table-align-uncertainty=true, mode=text}
    \renewrobustcmd{\bfseries}{\fontseries{b}\selectfont}
    \renewrobustcmd{\boldmath}{}
    \centering
    \vspace{-0.1in}
    \resizebox{\textwidth}{!}{\begin{tabular}{l|ccccccc}%
        \toprule
        Models &  Abs Rel ($\downarrow$) & Sq Rel ($\downarrow$) & RMSE ($\downarrow$) & RMSE$_{log}$ ($\downarrow$) & $\delta < 1.25$ ($\uparrow$) & $\delta < 1.25^2$ ($\uparrow$) & $\delta < 1.25^3$ ($\uparrow$) \tabularnewline 
        \midrule
        Ours-Combined & 0.204 & 2.388 & 7.232 & 0.289 & 0.729 & 0.892 & 0.949
        \tabularnewline
        Ours-Conditional & 0.221 & 3.173 & 7.474 & 0.298 & 0.724 & 0.887 & 0.944
        
        \tabularnewline
        Monodepth2~\cite{Godard2019ICCV} & \bfseries 0.136 & \bfseries 1.095 & \bfseries 5.369 & \bfseries 0.213 & \bfseries 0.836 & \bfseries 0.946 & \bfseries 0.977 
        \tabularnewline
        \bottomrule
    \end{tabular}}
    \vspace{-0.35cm}
\end{table*}
In order to validate the quality of our depth predictions, we evaluate them on KITTI with respect to a state-of-the-art monocular depth estimation approach Monodepth2~\cite{Godard2019ICCV} by training it on a similar resolution to ours ($320 \times 96$). Monodepth2 serves as an upper-bound for the performance of our models because we predict future depth based on previous frame predictions without actually seeing the target frame. %
As can be seen from \tabref{tab:depth_eval}, both our models perform reasonably well in all metrics~\cite{Eigen2014NeurIPS} in comparison to Monodepth2.

\begin{figure*}[h!]
\centering
\includegraphics[width=\textwidth]{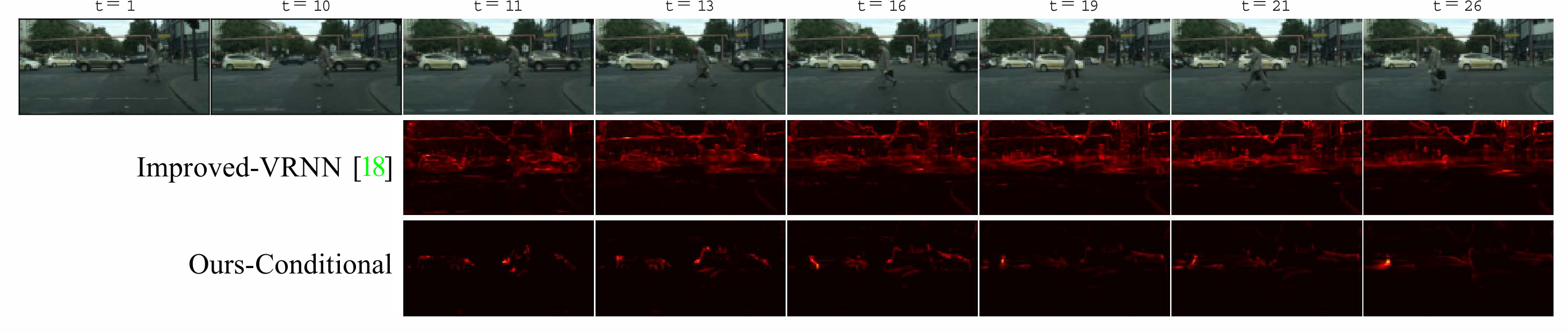}
\caption[Diversity of Samples]{\textbf{Diversity of Samples on Cityscapes.} We compare the standard deviations of 100 samples generated by Improved-VRNN~\cite{Castrejon2019ICCV} and our \emph{Conditional} model. Our model can pinpoint uncertain regions around the foreground objects in comparison to more uniform results of Improved-VRNN.}
\label{fig:diversity}
\end{figure*}

\begin{table}[t!]
    \caption[Run-time Comparison]{\textbf{Run-time Comparison of the Models.} We measure the time needed to generate 10 samples and report the results in seconds as the average over the dataset. We only measure the GPU processing time and discard the other operations.}
    \label{tab:speed}
    \sisetup{detect-weight, table-align-uncertainty=true, mode=text}
    \centering
    \vspace{0.1in}
    \resizebox{0.8\textwidth}{!}{\begin{tabular}{lrr}%
        \toprule
        \textbf{Models} &  \textbf{KITTI} & \textbf{Cityscapes} \tabularnewline 
        \midrule
        SVG~\cite{Denton2018ICML} & 1.04 & 0.90
        \tabularnewline
        SLAMP  & 2.03 & 1.44 \tabularnewline
        SRVP~\cite{Franceschi2020ICML}  & 0.11 & 0.12 \tabularnewline
        Improved VRNN~\cite{Castrejon2019ICCV}  & 39.25 & 41.59 \tabularnewline
        Ours-Combined & 1.02 & 1.16 \tabularnewline
        Ours-Conditional & 1.03 & 1.21 \tabularnewline
        \bottomrule
    \end{tabular}}
    \vspace{-5mm}
\end{table}
\boldparagraph{Run-time comparisons}
We compare the methods in terms of their run-time in \tabref{tab:speed}.
We measure the time needed to generate 10 samples and report the results in seconds as the average over the dataset. There is a trade off between the run-time and the prediction performance of the models. Improved-VRNN has the largest run-time by far due to its hierarchical latent space, with nearly 40 seconds of processing time for ten samples. The performance of Improved-VRNN is impressive, especially in terms of both LPIPS and FVD, however costly in terms of both memory and run-time, therefore not applicable to autonomous driving. Our models, both Combined and Conditional, have a similar run-time on KITTI and only a slight increase in run-time compared to vanilla stochastic video generation model SVG~\cite{Denton2018ICML}. The state-space model SRVP~\cite{Franceschi2020ICML} leads to the best run-time by removing the need for autoregressive predictions but at the cost of low performance on real-world datasets.

\subsection{Qualitative Results}
We visualize the results of our model including the intermediate predictions on KITTI (\figref{fig:qual_detailed}) as well as the final prediction in comparison to previous methods on Cityscapes (\figref{fig:qual_comp}). Please see Appendix for more results on both datasets. 
As can be seen from \figref{fig:qual_detailed}, our model can predict structure and differentiate between the two types of motion in the scene. First, the camera motion in the background is predicted~(\textit{Ego-motion}) and the motion of the foreground object is predicted as residual on top of it~(\textit{Residual Flow}).
The scene decomposition into static and dynamic and separate modeling of motion in each allow our model to generate better predictions of future in dynamic scenes. \figref{fig:qual_comp} shows a case where our model can correctly estimate the change in the shape and the size of the two vehicles moving in comparison to previous approaches.

We evaluate the diversity of predictions qualitatively in \figref{fig:diversity}. For this purpose, we visualize the standard deviations of generated frames over 100 samples for Improved-VRNN~\cite{Castrejon2019ICCV} and our \emph{Conditional} model. The higher the standard deviation at a pixel, the higher the uncertainty is at that pixel due to differences in samples. While the uncertainty is mostly uniform over the image for Improved-VRNN, our model can pinpoint it to the foreground regions thanks to our separate modeling of the motion history in the background and the foreground regions.

\section{Conclusion and Future Work}
\label{sec:chap3-conc}

We introduced a conditional stochastic video prediction model by decomposing the scene into static and dynamic in driving videos. We showed that separate modelling of foreground and background motion leads to better future predictions. Our method is among the top-performing methods overall while still being efficient. The modelling of the static part with domain knowledge is justified on KITTI with large camera motion. The separate modelling of foreground objects as residual motion leads to better results in dynamic scenes of Cityscapes. Our visualizations show that flow prediction can capture the residual motion of foreground objects on top of ego-motion thanks to our conditional model.

We found that modelling the stochasticity in terms of underlying factors is beneficial for autonomous driving. From this point forward, we see three important directions to improve stochastic video prediction in autonomous driving. Since the uncertainty is mostly due to foreground objects, stochastic methods can focus on foreground objects with more sophisticated models. Next, there is a limitation in our model due to the warping operation. Our model can predict the future motion of a car visible in the conditioning frames but it cannot predict a car appearing after that. For that, we need other ways of obtaining self-supervision without warping. 

{\clearpage
\thispagestyle{plain}
~
\newpage}
\chapter[Stretching Future Instance Prediction Spatially and Temporally]{StretchBEV: Stretching Future Instance Prediction Spatially and Temporally}
\label{chap:chapter4}

\section{Introduction}
\label{sec:chap4-intro}

Sequential visual data has been used before in autonomous driving tasks. However, the pixel space of the images or videos is noisy and contains a lot of redundant information. Instead, a more compact representation, which is much more suitable for the downstream tasks of autonomous driving, can be used. For example, understanding the 3D structure of the world and the motion of the agents around it can help the tasks in driving scenarios. The bird's-eye view~(BEV) representation meets these requirements by first fusing information from multiple cameras into a 3D point cloud and then projecting the points to the ground plane~\cite{Philion2020ECCV}. This leads to a compact representation where the agents, the lanes, and all the necessary information are preserved, and the redundant ones are dismissed.

In this chapter, we explore the potential of stochastic future prediction for self-driving in BEV representation. The aim is to generate admissible and diverse results in long sequences with an efficient and compact BEV representation.

Future prediction from the BEV representation has been recently proposed in FIERY~\cite{Hu2021ICCV}. Their method has several drawbacks. First, the lack of diversity might be a problem due to two distributions representing the present and the future. These two distributions may not be enough to model the diversity because the predictions of FIERY degrade over longer time spans due to the limited representation capability of a single distribution for increasing diversity in longer predictions. For example, for planning, long-term multiple future predictions are crucial.

Following FIERY, we start from the same BEV representation and predict the same output modalities to be comparable. Differently, instead of two distributions for the present and the future, we propose to learn time-dependent distributions by predicting a residual change at each time step to better capture long-term dynamics. Furthermore, we show that by sampling a random variable at each time step, we can increase the diversity of future predictions while still being accurate and efficient. For efficiency, we use a state-space model \cite{Murphy2023Prob} instead of costly auto-regressive models.

\section{Methodology}
\label{sec:method}
\subsection{A Compact Representation for Future Prediction}
\label{sec:bev}
Modern self-driving vehicles are typically equipped with multiple cameras observing the scene from multiple viewpoints. Placing cameras on the vehicle is cheap but processing information even from a single camera can be quite expensive. The traditional approach in computer vision is to extract low-level and semantic cues from these cameras and then fuse them into a holistic scene representation to perform prediction and planning. Recent success of end-to-end methods in driving has led to a rethinking of this approach. Furthermore, building and maintaining HD maps require a significant effort which is expensive and hard to scale. A better approach is to learn a geometrically consistent scene representation which can also mark the location, the motion, and even the semantics of the dynamic objects in the scene.

The bird's-eye view (BEV) representation initially proposed in \cite{Philion2020ECCV}, takes image $\bx_t^i$ at time $t$ from each camera $i \in \{1,\cdots,6\}$ and fuses them into a compact BEV representation $\bs_t$. This is achieved by encoding each image and also by predicting a distribution over the possible depth values. The BEV features are obtained by weighting the encoded image features according to depth probabilities predicted. These features are first lifted to 3D by using known camera intrinsic and extrinsic parameters and then the height dimension is pooled over to project the features into the bird's-eye view. This results in the state representation $\bs_t$ that we use for future prediction as explained next. 

\subsection{Learning Temporal Dynamics in BEV}
\label{sec:dynamics}
\begin{figure}[t]
\centering
    \includegraphics[width=\textwidth]{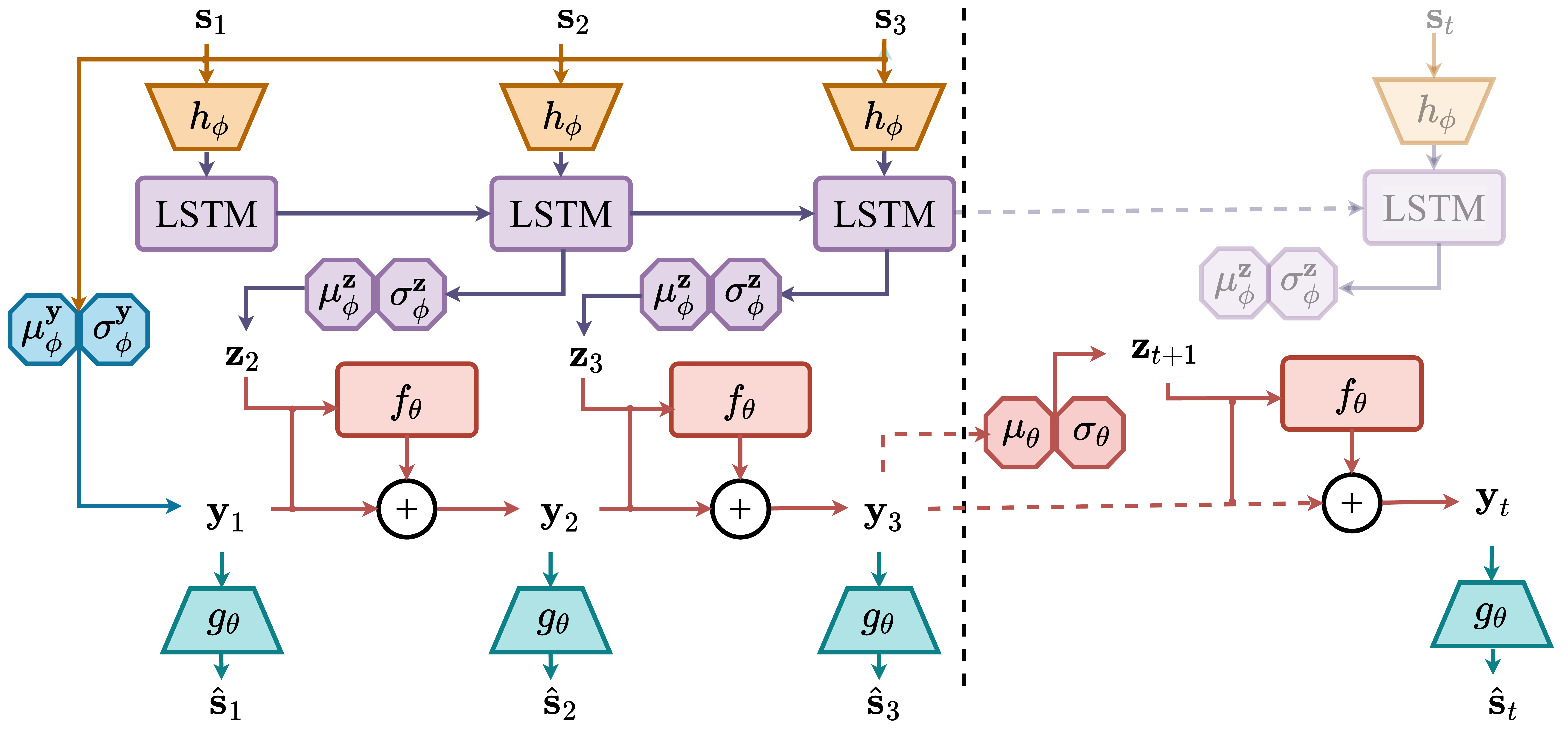}
    \caption[StretchBEV Architecture]{\textbf{Architecture for Learning Temporal Dynamics.} This figure shows the inference procedure of our model StretchBEV. We start with the first $k=3$ conditioning frames where we sample the stochastic latent variables from the posterior distribution~(purple). On the right, we show the prediction at a step $t$ after the conditioning frames where we sample from the learned future distribution~(red). The dashed vertical line marks the conditioning frames.}
    \label{fig:temporal_dynamics}
\end{figure}

\boldparagraph{Notation}
In our formulation, $\bs_{1:T}$ denotes a sequence of BEV feature maps representing the state of the vehicle and its environment for $T$ time steps. In stochastic future prediction, the goal is to predict possible future states $\hat{\bs}_{k+1:T}$ conditioned on the state in the first $k$ time steps. Precisely, we condition on the first $k=3$ steps and predict future in varying lengths from 4 to 12 steps ahead. 

Differently from previous work on stochastic video prediction \cite{Denton2018ICML,Franceschi2020ICML}, the BEV state $\bs_t$ that is input to our stochastic prediction framework is the intermediate representations in a high dimensional space rather than a video frame in the pixel space as explained in \secref{sec:bev}. Similarly, the predicted output $\hat{\bs}_t$ represents the predictions of future in the same high dimensional space. We decode these high dimensional future predictions into various output modalities $\hat{\bo}_t$ such as future instance segmentation and motion as explained in \secref{sec:decoding}. While these modalities need to be trained in a supervised manner in contrast to typical self-supervised stochastic video prediction frameworks \cite{Denton2018ICML,Franceschi2020ICML}, they provide interpretability which is critical in self-driving. Furthermore, using these modalities in the posterior in addition to the future state representations improves the results significantly as we show in our experiments. 

\boldparagraph{Stochastic Residual Dynamics}

Following \cite{Franceschi2020ICML}, we learn the changes in the states through time with stochastic residual updates to a sequence of latent variables. For each state $\bs_t$, there is a corresponding latent variable $\by_t$ generating it, independent of the previous states (\figref{fig:temporal_dynamics}). Each $\by_{t+1}$ only depends on the previous $\by_t$ and a stochastic variable $\bz_{t+1}$. The randomness is introduced by the stochastic latent variable $\bz_{t+1}$ which is sampled from a normal distribution learned from the previous state's latent variable only:
\begin{ceqn}
\begin{eqnarray}
    \bz_{t+1} \sim \cN\left(\mu_{\theta}(\by_t), \sigma_{\theta}(\by_t)~\bI \right)
\end{eqnarray}
\end{ceqn}

Given $\bz_{t+1}$, the dependency between the latent variables $\by_t$ and $\by_{t+1}$ is deterministic through the residual update:
\begin{ceqn}
\begin{eqnarray}
    \label{eq:y_res}
    \by_{t+1} = \by_t + f_{\theta}\left(\by_t, \bz_{t+1}\right)
\end{eqnarray}
\end{ceqn}
where $f_{\theta}$ is a small CNN to learn the residual updates to $\by_t$. We learn the distribution of future states from the corresponding latent variable as a normal distribution with constant diagonal variance: $\hat{\bs}_t \sim \cN(g_{\theta}(\by_t))$. The first latent variable is inferred from the conditioning frames by assuming a standard Gaussian prior: $\by_1 \sim \cN\left(\b0, \bI \right)$.

\boldparagraph{On the Content Variable}
In video prediction, a common practice is to represent the static parts of the scene with a content variable which allows the model to focus on the moving parts. On the contrary to the state of the art in video prediction~\cite{Franceschi2020ICML}, the content variable does not improve the results in our case (see Appendix), therefore we omit it in our formulation here. This can be attributed to the details in the background that are confusing for learning dynamics while operating in the pixel space, but in our case, most of these details are already suppressed in the BEV representation. 

\boldparagraph{On Diversity}
In contrast to a present and a future distribution in FIERY \cite{Hu2020ECCV,Hu2021ICCV}, there is a distribution learned at each time step in our model. This corresponds to sampling stochastic random variables at each time step as opposed to sampling once to represent all the future frames. This is the key property which allows our model to produce diverse predictions in long sequences. By sampling from a learned distribution at each time step, our model learns to represent the complex dynamics of future frames, even for predictions further away from the conditioning frames. Furthermore, our model does not need a separate temporal block for learning the dynamics prior to learning these distributions. The dynamics are learned through the temporal evolution of latent variables by also considering the randomness of the future predictions with stochastic random variables at each time step. This not only increases the diversity of predictions but also alleviates the need for a separate temporal block, \eg with 3D convolutions. Note that our formulation is still efficient, almost the same inference time as FIERY (see Appendix), because the latent variables are low dimensional and each state is generated independently.

Moreover, FIERY uses only a single vector of latent variables which is expanded to the spatial grid to generate futures states probabilistically. Therefore, it uses the same stochastic noise in all the coordinates of the grid. However, in our model, we have a separate random variable at each coordinate of the grid to model the uncertainty spatially as well. Training FIERY with a separate random variable at each location of the grid results in diverging loss values.

\subsection{Variational Inference and Architecture}
Following the generative process in \cite{Franceschi2020ICML}, the joint probability of the BEV states $\bs_{1:T}$, the output modalities $\bo_t$, and the latent variables $\bz_{1:T}$ and $\by_{1:T}$ is as follows: %
\begin{ceqn}
\begin{eqnarray}
    \label{eq:cond_prob}
    p\left(\bs_{1:T}, \bo_{1:T}, \bz_{2:T}, \by_{1:T} \right) &=& p\left(\by_1\right) \prod_{t=2}^T p\left(\bz_t,\by_t \vert \by_{t-1} \right) \prod_{t=1}^T p\left(\bo_t \vert \bs_t \right) p\left(\bs_t \vert \by_t \right) \\
    \label{eq:z_y}
    p\left(\bz_t, \by_t \vert \by_{t-1} \right) &=& p\left(\by_t \vert \by_{t-1}, \bz_t \right) p\left(\bz_t \vert \by_{t-1} \right) 
\end{eqnarray}
\end{ceqn}
The relationship between $\by_t$ and $\by_{t-1}$ in $p\left(\by_t \vert \by_{t-1}, \bz_t \right)$ \eqref{eq:z_y} is deterministic through the stochastic latent residual as formulated in \eqref{eq:y_res}. Similarly for the term $p\left(\bo_t \vert \bs_t \right)$ in \eqref{eq:cond_prob}, the output modalities $\bo_t$ is learned from $\bs_t$  with a deterministic decoder in a supervised manner (\secref{sec:decoding}).

Our goal is to maximize the likelihood of the BEV states extracted from the frames (\secref{sec:bev}) and the corresponding output modalities $p\left(\bs_{1:T}, \bo_{1:T}\right)$. For that purpose, we learn a deep variational inference model $q$ parametrized by $\phi$ which is factorized as follows: 
\begin{ceqn}
\begin{eqnarray}
    \label{eq:fact}
    q_{Z,Y} &\triangleq& q\left(\bz_{2:T}, \by_{1:T} \vert \bs_{1:T}, \bo_{1:T} \right) \\ &=& q\left(\by_1 \vert \bs_{1:k} \right) \prod_{t=2}^T q\left(\bz_t \vert \bs_{1:t}, \bo_{1:t} \right) q\left(\by_t \vert \by_{t-1}, \bz_t \right) \nonumber
\end{eqnarray}
\end{ceqn}
where $k = 3$ is the number of conditioning frames and $q\left(\by_t \vert \by_{t-1}, \bz_t \right)$ is equal to $p\left(\by_t \vert \by_{t-1}, \bz_t \right)$ with the residual update as explained above. We obtain two versions of our model by keeping (StretchBEV-P) or removing (StretchBEV) the dependency of $\bz_t$ on $\bo_{1:t}$ in the posterior in \eqref{eq:fact}.
We refer the reader to Appendix for the derivation of the following evidence lower bound~(ELBO):

\begin{eqnarray}
    \label{eq:elbo}
    \text{log}~p\left( \bs_{1:T}, \bo_{1:T} \right) &~&\ge \cL \left( \bs_{1:T}, \bo_{1:T}; \theta, \phi \right) \\
    \triangleq &-& D_{\text{KL}} \left( q\left( \by_1 \vert \bs_{1:k}\right) \mid\mid p\left( \by_1\right) \right) \nonumber \\
    &+& \nE_{\left(\tilde{\bz}_{2:T}, \tilde{\by}_{1:T}\right)\sim q_{Z,Y}}
    \Bigg[\sum_{t=1}^T \text{log}~p\left( \bs_t \vert \tilde{\by}_t\right) p\left( \bo_t \vert \bs_t\right) \nonumber \\ 
    &-& \sum_{t=2}^T D_{\text{KL}} \big( q\left( \bz_t \vert \bs_{1:t}, \bo_{1:t} \right) \mid\mid p\left( \bz_t \vert \tilde{\by}_{t-1} \right) \big)\bigg]
    \nonumber
\end{eqnarray}
where $D_{\text{KL}}$ denotes the Kullback-Leibler~(KL) divergence, $\theta$ and $\phi$ represent model and variational parameters, respectively. Following the common practice, we choose $q\left( \by_1 \vert \bs_{1:k}\right)$ and $q\left( \bz_t \vert \bs_{1:t}, \bo_{1:t} \right)$ to be factorized Gaussian for analytically computing the KL divergences and use the re-parametrization trick~\cite{Kingma2014ICLR} to compute gradients through the inferred variables. 
Then, the resulting objective function is maximizing the ELBO as defined in \eqref{eq:elbo} and minimizing the supervised losses for the output modalities (\secref{sec:decoding}).

We provide a summary of the steps in our temporal model as shown in \figref{fig:temporal_dynamics}. We start by fusing the images $\bx^i_t$ at time $t$ from each camera $i$ into the BEV state $\bs_t$ at each time step as explained in \secref{sec:bev}. 
\begin{enumerate}
    \item The resulting BEV states are still in high resolution, $\bs_t \in \nR^{C \times H \times W}$ where $(H,W) = (200,200)$. Therefore, we first process them with an encoder $h_{\phi}$ to reduce the spatial resolution to $50\times50$. 
    \item The first latent variable $\by_1$ is inferred using a convolutional neural network on the first three encoded states.
    \item  The stochastic latent variable $\bz_t$ is inferred at each time step from the respective encoded state, using a recurrent neural network which is a combination of a ConvGRU and convolutional blocks.
    \item The residual change in the dynamics is predicted with $f_{\theta}$ based on both the previous state dynamics $\by_t$ and the stochastic latent variable $\bz_{t+1}$ and added to $\by_t$ to obtain $\by_{t+1}$.
    \item From each $\by_t$, the state $\hat{\bs}_t$ is predicted in the original resolution with $g_{\theta}$.
    \item Finally, the output modalities $\hat{\bo}_t$ are decoded from the state prediction $\hat{\bs}_t$.
\end{enumerate}

\subsection{Decoding Future Predictions}
\label{sec:decoding}
Based on the predictions of the future states at each time step, we train a supervised decoder to output semantic segmentation, instance center, offsets, and future optical flow, similar to the previous work~\cite{Cheng2020CVPR,Hu2021ICCV}. The decoding function is a deterministic neural network that can be trained either jointly with the dynamics or independently, \eg later for interpretability. The output modalities show both the location and the motion of instances at each time step. The motion predicted as future flow is used to track instances. We use the same supervised loss functions for each modality as the FIERY~\cite{Hu2021ICCV}. Although the decoding of future predictions is not necessary for planning and control, for example when training a driving agent to act on predictions, these predictions provide interpretability and allow to compare the methods in terms of various metrics evaluating each modality.

\section{Experiments}
\label{sec:exp}
\subsection{Dataset and Evaluation Setting}
\label{sec:data_eval}
We evaluate the performance of the proposed approach and compare to the state of the art method, FIERY \cite{Hu2021ICCV}, on the nuScenes dataset~\cite{Caesar2020CVPR}. On the nuScenes, there are $6$ cameras with overlapping views which provide the ego vehicle with a complete view of its surroundings. The nuScenes dataset consists of $1000$ scenes with $20$ seconds long at $2$ frames per second.

We first follow the training and the evaluation setting proposed in FIERY~\cite{Hu2021ICCV} for comparison by using $1.0$ second of past context to predict $2.0$ seconds of future context. Given the sampling rate of $2$ frames per second, this setting corresponds to predicting $4$ future frames conditioned on $3$ past frames. We call this setting \emph{short} in terms of temporal length and define two more settings for longer temporal predictions. In the \emph{mid} and \emph{long} settings, we double and triple the number of future frames to predict, \ie $8$ and $12$, respectively, that corresponds to $4.0$ and $6.0$ seconds into the future. These settings are closer to the stochastic video prediction setup~\cite{Denton2018ICML,Franceschi2020ICML,Akan2021ICCV} where there are typically many more frames to predict than the conditioning frames for measuring diversity and the performance of the models further away from the conditioning frames. Note that \emph{short} and \emph{long} refer to temporal length in our evaluations as opposed spatial coverage as defined in the previous work~\cite{Hu2021ICCV}. We also evaluate in terms of spatial coverage but call it \emph{near}~($30$m$\times30$m) and \emph{far}~($100$m $\times 100$m) for clarity.

\subsection{Training Details}
Our models follow the input and output setting proposed in the previous work \cite{Hu2021ICCV}. We process 6 camera images at a resolution of $224 \times 480$ pixels for each frame and construct the BEV state of size $200 \times 200 \times 64$. We further process the states into a smaller spatial resolution ($50 \times 50$) for efficiency before learning the temporal dynamics but increase it back to the initial resolution afterwards. Given the predicted states, we use the same decoder architecture as the FIERY to decode the object centers, the segmentation masks, the instance offsets, and the future optical flow at a resolution of $200 \times 200$ pixels. We provide the details of the architecture in Appendix and we will release the code upon publication.

In our approach, learning temporal dynamics and decoding output modalities are separated from each other. Therefore, we can pre-train the temporal dynamics part without using the labels for the output modalities. In pre-training, our objective is to learn to match the future states that are extracted using a pre-trained BEV segmentation model~\cite{Philion2020ECCV}, conditioned on the past states. This approach is more similar to self-supervised stochastic video prediction methods \cite{Denton2018ICML,Franceschi2020ICML}. Furthermore, this way, learning of temporal dynamics can be improved by using camera sequences only as input which can be easily collected in large quantities. Then, we fine-tune the temporal dynamics with a smaller learning rate (see Appendix for details) while learning to decode the output modalities in a supervised manner. The alternative is to jointly train the temporal dynamics and supervised decoding without pre-training. We present the results of our model StretchBEV with and without pre-training.

StretchBEV does not use the labels ($\bo_t$) for learning the temporal dynamics, it only uses them in the supervised loss to decode the output modalities. In our full model StretchBEV-P, we encode output modalities following FIERY and use them in the posterior for learning the temporal dynamics. During training, we sample the stochastic latent variables from the posterior and learn to minimize the difference between the posterior and the future distribution. During inference, we sample from the posterior in the conditioning frames and sample from the learned future distribution in the following steps as shown in \figref{fig:temporal_dynamics}.

\subsection{Metrics}
We use two different metrics for evaluating the decoded modalities, one frame level and another video level, that are also used in the previous work~\cite{Hu2021ICCV}. The first is Intersection over Union~(IoU) to measure the quality of the segmentation at each frame. The second is Video Panoptic Quality~(VPQ) to measure the quality of the segmentation and consistency of the instances through the video:

\begin{ceqn}
\begin{eqnarray}
    \text{VPQ} = \sum_{t=0}^H \frac{\sum_{(p_t,q_t) \in TP_t} \text{IoU}(p_t,q_t)}{|TP_t| + \frac{1}{2}|FP_t| + \frac{1}{2}|FN_t|}
\end{eqnarray}
\end{ceqn}

where $H$ is the temporal horizon considered, $TP_t$ is the number of True Positives, $FP_t$ the number of False Positives, and $FN_t$ the number of False Negatives at a time step $t$. 
We report both metrics for both spatially near and far regions and temporally short, mid, and long spans as explained in \secref{sec:data_eval}.

We evaluate the diversity quantitatively in terms of Generalized Energy Distance ($D_\mathrm{GED}$) \cite{GED} by using $\left(1 - \mathrm{VPQ}\right)$ as the distance as proposed in FIERY~\cite{Hu2021ICCV}. 

\begin{figure}[t]
\centering
    \includegraphics[width=\textwidth]{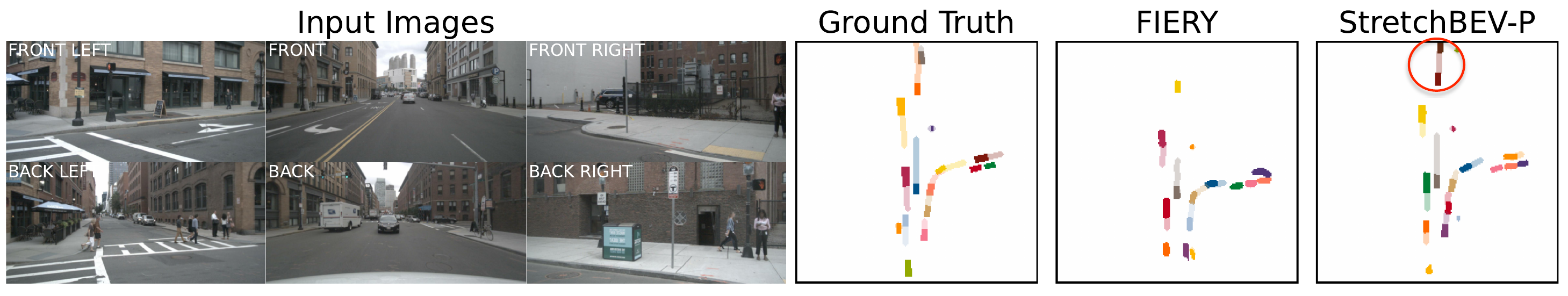} %
    \\
    \includegraphics[width=\textwidth]{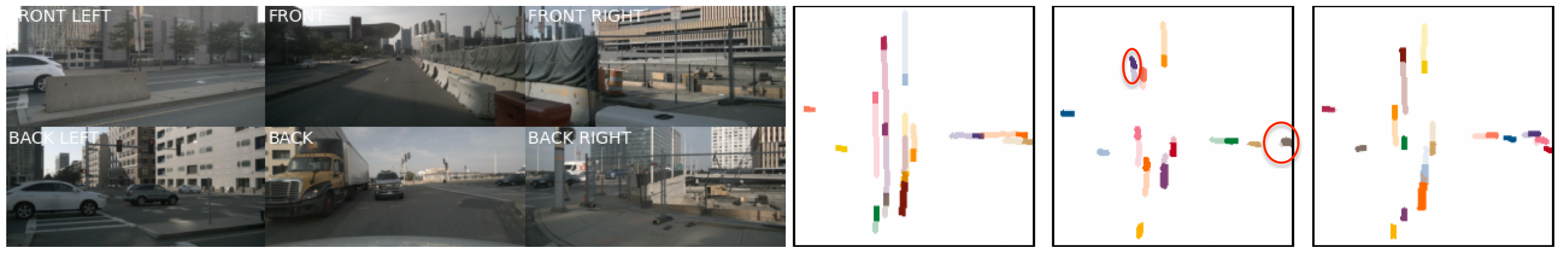} \\ %
    \includegraphics[width=\textwidth]{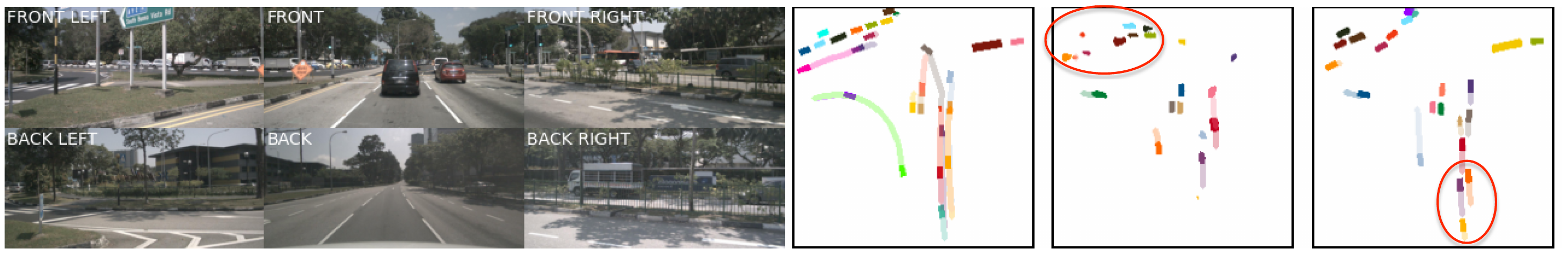} 
    \caption[Qualitative Comparison over Different Temporal Horizons.]{\textbf{Qualitative Comparison over Different Temporal Horizons.} In this figure, we qualitatively compare the results of our model StretchBEV-P \textbf{(right)} to the ground truth \textbf{(left)} and FIERY \cite{Hu2021ICCV} \textbf{(middle)} over short \textbf{(top)}, mid \textbf{(middle)}, and long \textbf{(bottom)} temporal horizons. Each color represents an instance of a vehicle with its trajectory trailing in the same color transparently.}
    \label{fig:comp_temp}
    \vspace{-0.5cm}
\end{figure}

\subsection{Ablation Study}
\begin{table}[ht!]
    \centering
    \setlength{\tabcolsep}{7pt} 
    \begin{tabular}{c | c | c | c c | c c}
     & Pre- & ~Posterior~ & \multicolumn{2}{c|}{IoU ($\uparrow$)} & \multicolumn{2}{c}{VPQ ($\uparrow$)} \\ 
    \textbf{} & ~training~ & w/labels & ~Near~ & ~Far~ & ~Near~ & ~Far~ \\ \toprule
    \multirow{2}{*}{~StretchBEV~} & \textemdash & \multirow{2}{*}{\textemdash} & 53.3 & 35.8 & 41.7 & 26.0 \\
     & \checkmark & & 55.5 & 37.1 & 46.0 & 29.0 \\ 
    \midrule
     FIERY~\cite{Hu2021ICCV} & \multirow{2}{*}{\textemdash} & \multirow{2}{*}{\checkmark} & \textbf{59.4} & 36.7 & 50.2 & 29.4 \\
    Reproduced & & & 58.8 & 35.8 & 50.5 & 29.0 \\
    \midrule
    StretchBEV-P & \textemdash &  \checkmark & 58.1 & \textbf{52.5} & \textbf{53.0} & \textbf{47.5} \\
    \bottomrule
    \end{tabular}
    \caption[StretchBEV Ablation Study]{\textbf{Ablation Study.} In this table, we present the results for the two versions of our model with (StretchBEV-P) and without (StretchBEV) using the labels for the output modalities in the posterior while learning the temporal dynamics and also show the effect of pre-training for the latter in comparison to FIERY \cite{Hu2021ICCV} and our reproduced version of their results (Reproduced).}
    \label{tab:ablation}
    \vspace{-1cm}
\end{table}
In \tabref{tab:ablation}, we evaluate the effect of different versions of our model using IoU and VPQ metrics in the short temporal setting to be comparable to the previous work FIERY~\cite{Hu2021ICCV}. We reproduced their results as shown in the row \emph{Reproduced}. In the first part of the table~(StretchBEV), we show the results without explicitly using the labels for future prediction. In that case, labels are only used for decoding the output modalities and back-propagated to future prediction through decoding. Although this introduces a two-stage training, we believe that reporting results using this separation is important for future work to focus on future prediction with more unlabelled data.
We measure the effect of pre-training by learning to match our future predictions to the results of a pre-trained model~\cite{Philion2020ECCV} in terms of the BEV state representation. Pre-training allows our model to learn the dynamics before decoding and improves the results significantly in each metric.

In the second half of the \tabref{tab:ablation}, we report the results using the labels in future prediction by explicitly feeding their encoding to the posterior distribution with the same encoding used in \cite{Hu2021ICCV} to learn the future distribution. The difference between StretchBEV and StretchBEV-P is that the first has access to the BEV encoding of future predictions while the latter has access to both the BEV encoding and the encoding of the output modalities to predict in the posterior distribution. As can be seen from the results, both FIERY and our model using the labels in the future distribution perform better. This shows the importance of using a more direct and accurate information about future while learning the posterior. Compared to FIERY, our model can use the labels in the conditioning frames during inference and improves the results, especially in spatially far regions and in terms of VPQ, which point to a higher quality in our predictions stretching spatially and temporally over the video.

\subsection{Temporally Long Predictions}
\begin{figure}[t]
\centering
    \includegraphics[width=\textwidth]{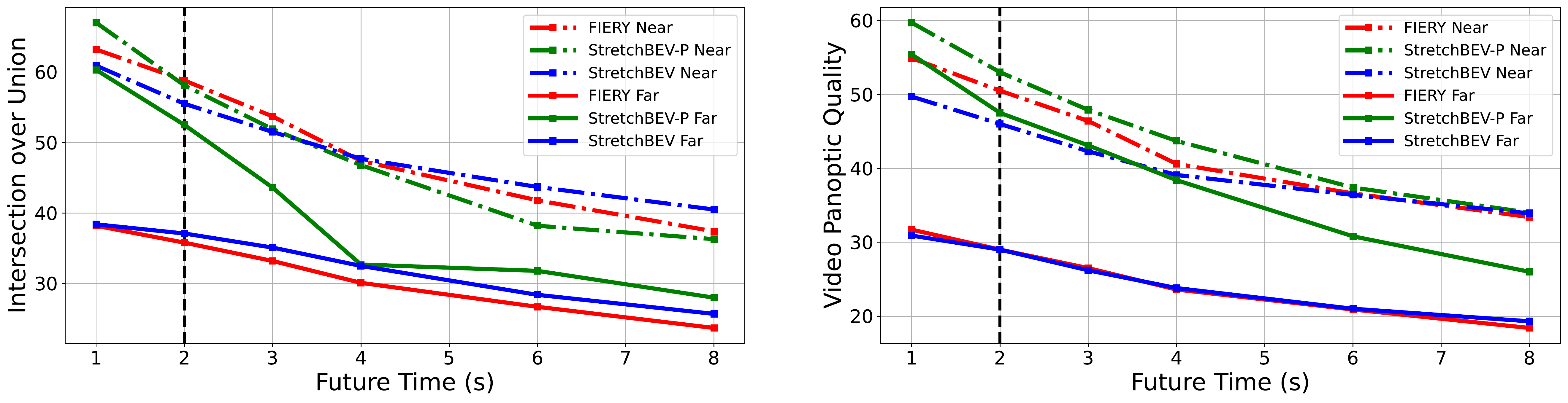}
    \caption[Evaluation over Different Temporal Horizons]{\textbf{Evaluation over Different Temporal Horizons.} We plot the performance of our models StretchBEV and StretchBEV-P in comparison to FIERY~\cite{Hu2021ICCV} over a range of temporal horizons from 1 second to 8 seconds in terms of IoU~\textbf{(left)} and VPQ~\textbf{(right)} for spatially far~\textbf{(solid)} and near~\textbf{(dashed)} regions separately. The vertical dashed line marks the training horizon.}
    \label{fig:temporal_horizon}
\end{figure}

In longer temporal horizons, future prediction becomes increasingly difficult. This is mainly due to increasing uncertainty of future further away from conditioning frames. %
In \figref{fig:temporal_horizon},
we present the results over different temporal horizons for our model with pre-training without using the labels in the posterior (StretchBEV), FIERY~\cite{Hu2021ICCV}, and our model by using the labels in the posterior (StretchBEV-P). There is a separate plot for IoU on the left and for VPQ on the right with respect to the future time steps predicted, ranging from $1$s to $8$s. The vertical line in $2$s marks the training horizon. 
In Appendix, we provide a table for the results over short, mid, and long temporal spans.

The negative effect of uncertain futures on each metric can be observed from the results of all the methods degrading from shorter to longer temporal spans. Our models perform better than FIERY in longer temporal spans. This is due to better handling of uncertainty with stochastic latent residual variables. Our method StretchBEV-P outperforms FIERY by significant margins, especially in terms of far VPQ in longer temporal horizons, showing consistent predictions in the overall scene throughout the video. This can be attributed to the difficulty of locating small vehicles in spatially far regions. Since StretchBEV-P has access to the labels via the posterior in the conditioning frames, it learns the temporal dynamics to correctly propagate them to the future frames, while StretchBEV and FIERY struggle to locate the instances in the first place. FIERY learns a single distribution for present and future each, therefore we cannot utilize the labels in the conditioning frames with FIERY. The results of StretchBEV outperforming the other two methods in terms of near IOU in longer temporal spans is promising for future prediction methods with less supervision. 

In \figref{fig:comp_temp}, we qualitatively compare the performance of our model StretchBEV-P on the right to FIERY in the middle over short, mid, and long temporal horizons in each row. In the first row, our model predicts the future trajectories that are more similar to the ground truth shown on the left. For example, FIERY fails to predict the trajectory of the vehicle in front (marked with red circle). In the second row, our model correctly segments the vehicles, whereas FIERY misses several vehicles far on the right and also, predicts a vehicle that does not exist (in purple on the top left). In the third row, our model predicts the future trajectories of the vehicles correctly while FIERY misses some of the vehicles (marked with red circles). The challenging case of a vehicle turning on the left (green in ground truth) is missed by both models. Some of the vehicles are not visible on the input images, \eg the back camera in the long temporal horizon. 

\subsection{Segmentation}
\begin{table}[h]
    \centering
    \setlength{\tabcolsep}{5pt} 
    \begin{tabular}{lc}
    Methods & Segmentation IoU \\ \toprule
    Fishing-Cam~\cite{Hendy2020CVPRW} & 30.0 \\
    Fishing-LiDAR~\cite{Hendy2020CVPRW} & 44.3 \\
    FIERY~\cite{Hu2021ICCV} & 57.3 \\
    StretchBEV & \underline{58.8} \\
    StretchBEV-P & \textbf{65.7} \\ \bottomrule
    \end{tabular}
    \caption[Comparison of Semantic Segmentation Prediction]{\textbf{Comparison of Semantic Segmentation Prediction.} In this table, we compare the predictions of our models, StretchBEV and StretchBEV-P for semantic segmentation to other BEV prediction methods in terms of IoU using the setting proposed in \cite{Hendy2020CVPRW}, \ie $32.0$m $\times$ $19.2$m at $10$cm resolution over $2$s future.}
    \label{tab:seg}
    \vspace{-0.7cm}
\end{table}
The previous work on bird's-eye view segmentation typically focuses on single image segmentation task with a couple of exceptions focusing on prediction. In \tabref{tab:seg}, we compare our methods, StretchBEV and StretchBEV-P, to two BEV segmentation prediction methods~\cite{Hendy2020CVPRW,Hu2021ICCV} using their setting with $32.0$m $\times$ $19.2$m at $10$cm resolution. Both methods predict $2$s into the future which corresponds to our short temporal setting. FIERY~\cite{Hu2021ICCV} outperforms the previous method~\cite{Hendy2020CVPRW} even when using LiDAR, and our method significantly outperforms both methods. %

\subsection{Diversity}
\begin{table}[h!]
    \centering
    \setlength{\tabcolsep}{7pt} 
    \begin{tabular}{c | c c | c c | c c}
    \multicolumn{1}{c}{} & \multicolumn{6}{c}{Generalized Energy Distance ($\downarrow$)} \\
    \cmidrule(){2-7}
    \multicolumn{1}{c}{} & \multicolumn{2}{c}{Short} & \multicolumn{2}{c}{Mid} & \multicolumn{2}{c}{Long} \\
    \cmidrule(r){2-3} \cmidrule(r){4-5} \cmidrule(r){6-7}
    \multicolumn{1}{c|}{} &  Near & Far & Near & Far & Near & Far  \\ 
    \toprule
    FIERY~\cite{Hu2021ICCV} & 106.09 & 140.36 & 118.74 & 147.26 & 127.18 & 152.38 \\  
    StretchBEV & \underline{103.97} & \underline{132.38} & \underline{114.11} & \underline{138.15} & \underline{119.01} & \underline{142.51} \\
    StretchBEV-P &  \textbf{82.04} & \textbf{85.51} & \textbf{94.02} & \textbf{98.45} & \textbf{101.90} & \textbf{109.12} \\
    \bottomrule
    \end{tabular}
    \caption[Quantitative Evaluation of Diversity]{\textbf{Quantitative Evaluation of Diversity.} This table compares the results of our models to the reproduced results of FIERY~\cite{Hu2021ICCV} in terms of Generalized Energy Distance based on VPQ~(lower better) for evaluating diversity.} %
    \label{tab:ged}
\end{table}
\begin{figure}[t]
\centering
    \includegraphics[width=.49\textwidth]{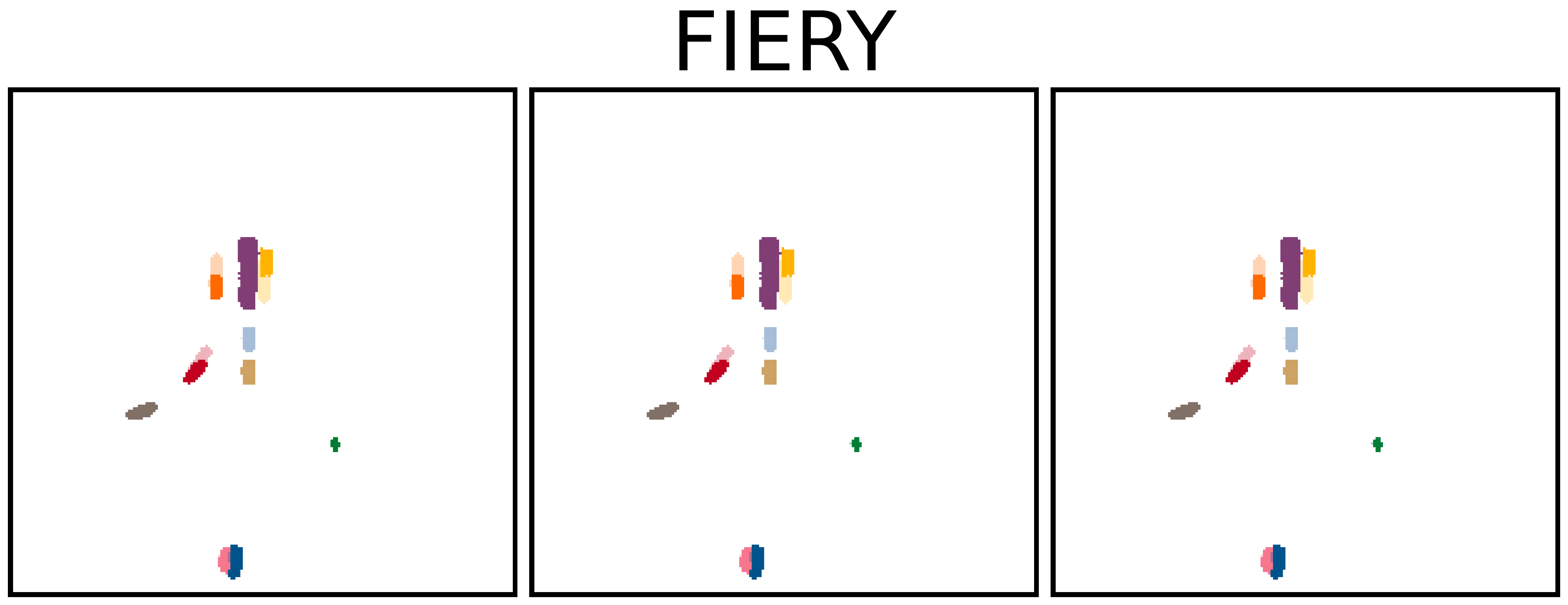}  \includegraphics[width=.49\textwidth]{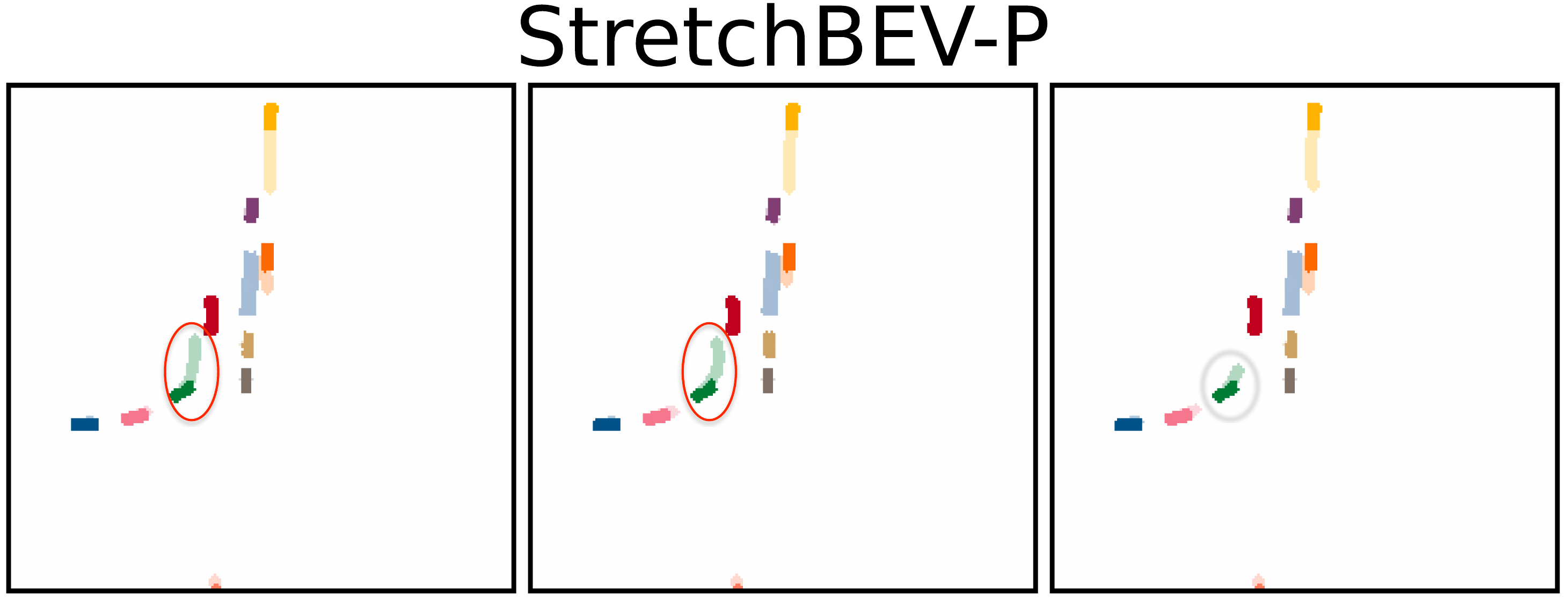} \\
    \includegraphics[width=.49\textwidth]{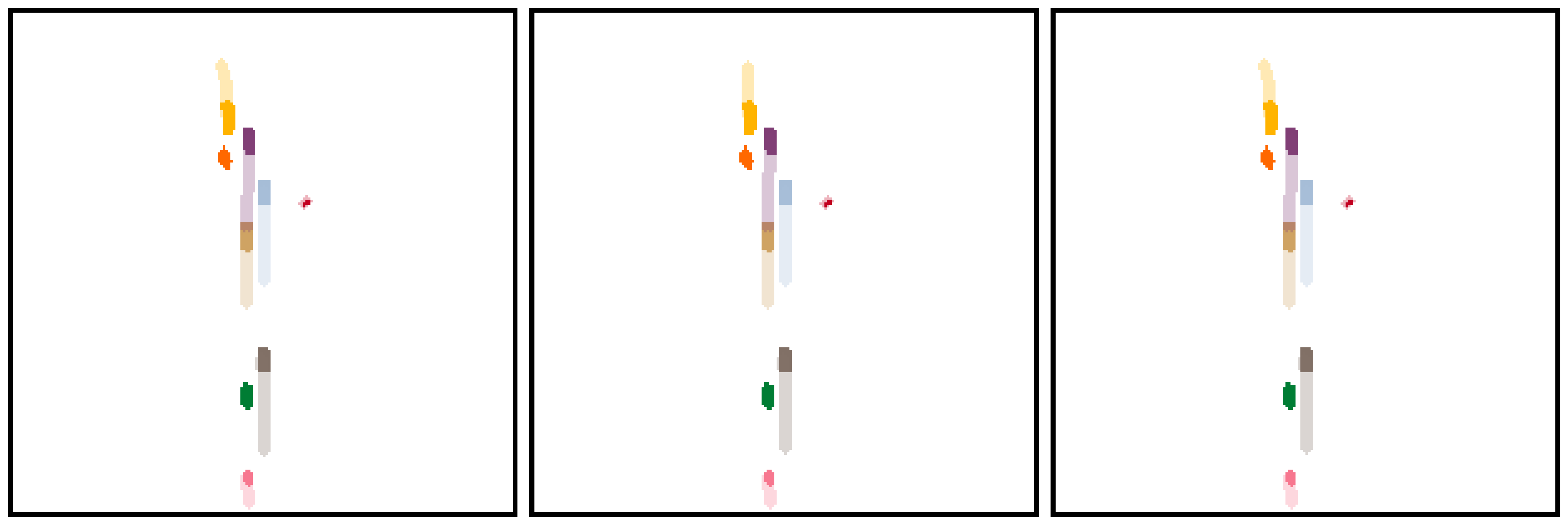} \includegraphics[width=.49\textwidth]{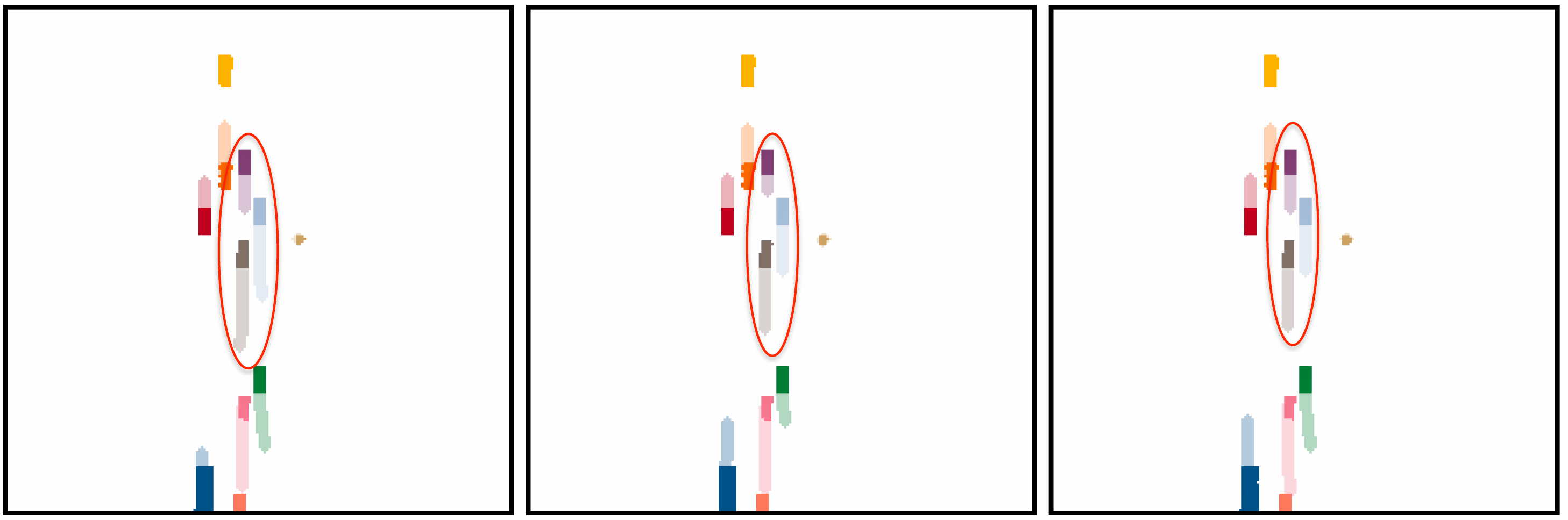} \\
    \includegraphics[width=.49\textwidth]{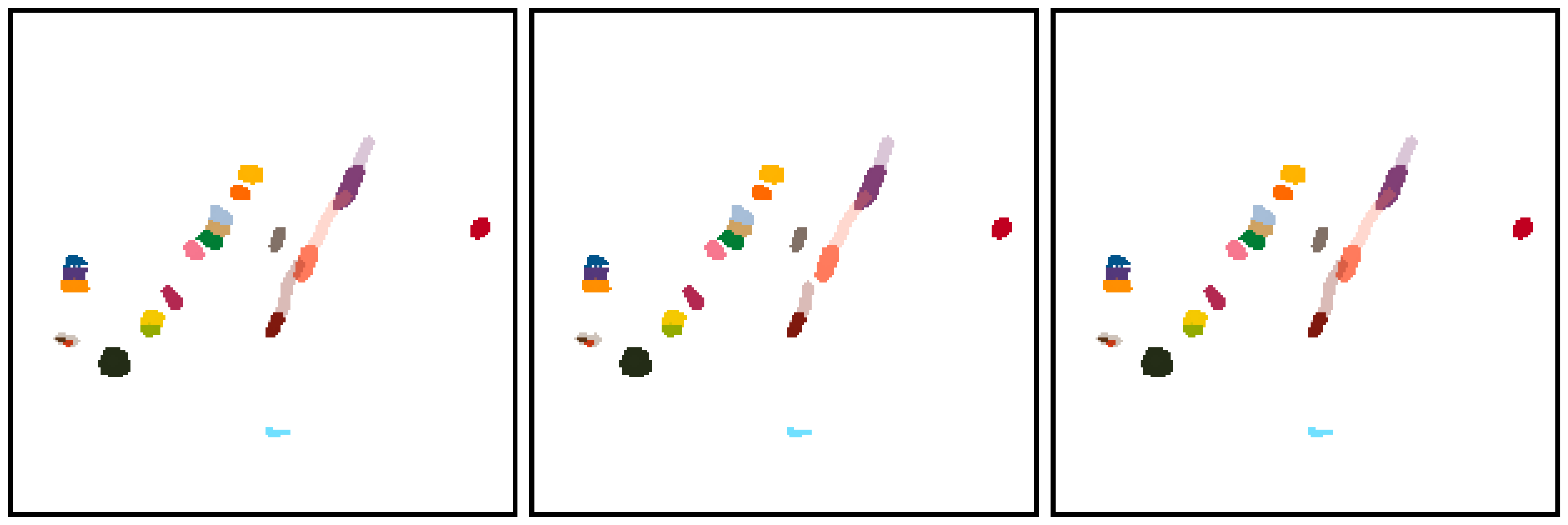} \includegraphics[width=.49\textwidth]{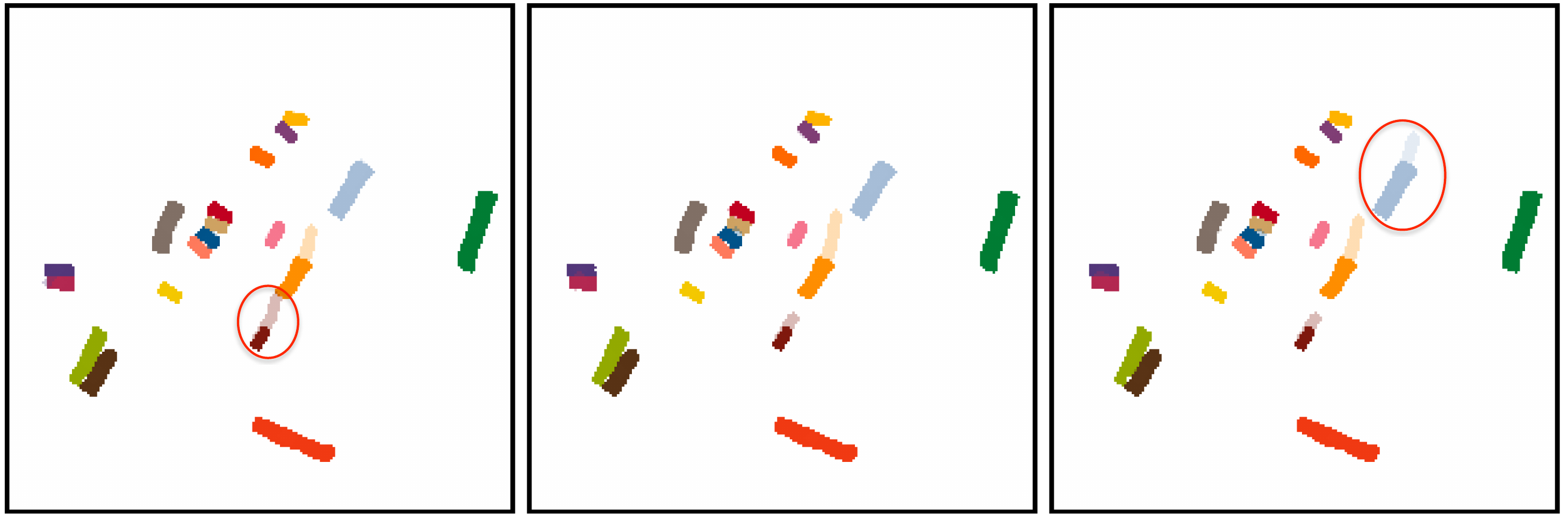}
    \caption[Qualitative Comparison of Diversity.]{\textbf{Qualitative Comparison of Diversity.} In this figure, we visualize random samples from FIERY \cite{Hu2021ICCV} \textbf{(left)} and our model StretchBEV-P \textbf{(right)} over short~\textbf{(top)}, mid~\textbf{(middle)}, and long~\textbf{(bottom)} temporal horizons.}
    \label{fig:comp_samples}
    \vspace{-0.5cm}
\end{figure}

We quantitatively evaluate diversity by computing $D_\mathrm{GED}$ over 10 samples and show the results in \tabref{tab:ged} for our models, StretchBEV and StretchBEV-P and FIERY (our reproduced version). Both our models outperforms FIERY with lower distance scores, demonstrating higher levels of diversity in the samples quantitatively. The difference is especially apparent in spatially far regions.
For qualitative comparison, in \figref{fig:comp_samples}, we visualize three samples from FIERY (left) and our model (right) over short, mid, and long temporal spans from top to bottom. While FIERY generates almost the same predictions in all three samples, our model can generate diverse predictions of future (marked with red). In the first row, our model can predict the turning behavior of the green vehicle at different speeds. In the second row, our model learns to adjust the speed of nearby vehicles proportionally, as in the case of purple, blue, and gray vehicles in the middle. Similarly, in the third row, our model can generate different predictions for the moving vehicles in the middle.

\boldparagraph{Run-time Comparison} 
We compare the inference speed of our model StretchBEV-P and FIERY \cite{Hu2021ICCV} by measuring the average time needed to process a validation example in inference over 250 forward passes. Both models have almost the same inference speed (FIERY: $0.6436$ seconds/example vs. StretchBEV-P: $0.6469$ seconds/example). Although our model processes each time step separately, it does not introduce any drawbacks in terms of speed and its inference speed is almost the same as FIERY.

\section{Conclusion}
We introduced StretchBEV, a stochastic future instance prediction method that improve over the state of the art, especially in challenging cases, with more diverse predictions. We proposed two versions of our method with and without the labels for output modalities  explicitly in the posterior while learning the dynamics. Both models improve the state of the art in spatially far regions and over temporally long horizons. Using labels in the posterior significantly improves the results in almost all metrics but introduces a dependency on the availability of labels in the conditioning frames during inference.

{\clearpage
\thispagestyle{plain}
~
\newpage}
\chapter{Conclusion}
\label{chap:conc}

This chapter provides a brief summary, explain the limitations of our work and .

\section{Summary}

In this thesis, we propose two novel methods for stochastic video prediction and another method for future instance segmentation in Bird's-eye view representation. 

In Chapter~\ref{chapter:slamp}, we introduced SLAMP, which can predict future frames of a video, given a few initial ones. SLAMP achieves state-of-the-art results on challenging real-world datasets with a moving background while performing on par with the other methods on the generic video prediction datasets. SLAMP can decompose video content into appearance and dynamic components, each of which is modeled separately with different latent variables. Moreover, we showed that motion history enriches the model's capacity to predict the future, leading to better predictions in challenging cases. 

In Chapter~\ref{chapter:chap3}, we have shown that using domain knowledge can improve the performance in stochastic video prediction. We proposed SLAMP-3D, a conditional stochastic video prediction model, by decomposing the scene into static and dynamic in driving videos. We showed that separate modeling of foreground and background motion leads to better future predictions. Also, by conditioning the dynamic latent variable on the static one, SLAMP-3D can predict the residual motion of foreground objects on top of ego-motion, which is predicted by the static part. We demonstrated that our model is among the top-performing methods overall while still being efficient.

In Chapter~\ref{chap:chapter4}, we have stated that by formulating the problem of future instance segmentation as a learning temporal dynamics through stochastic residual updates to a latent representation in each time step, both the performance and the diversity can be improved. We introduced StretchBEV, which is able to perform better than previous methods with a powerful posterior distribution, especially in challenging cases of spatially far regions and temporally long spans. We showed that our model achieves state-of-the-art results while still being efficient in terms of run-time performance.

\section{Limitations and Possible Future Directions}

In all of the methods we propose, there is no hierarchy. Introducing hierarchy for both latent variables and the predictors might increase the performance. In the literature, hierarchic methods have been used~\cite{Castrejon2019ICCV} and the results are improved using the hierarchy. Thus, we believe that introducing hierarchy in the SLAMP's architecture may improve the performance in both generic video prediction and real world datasets.

In SLAMP-3D, using the domain knowledge, we model video as the ego-motion and the residual motion in the scene. Since both parts warp the previous frames to synthesize the target frame, our model can predict the future motion of a car visible in the conditioning frames but it cannot predict a car appearing after that. This limitation can be solved by introducing a separate branch which predicts the target frame in pixel space. However, this improvement is against the motivation of SLAMP-3D. Moreover, we could not use any of the well-known tricks in view synthesis such as multi-scale, forward-backward consistency check, 3D convolutions, and better loss functions~\cite{Godard2019ICCV, Guizilini2020CVPR} due to computational reasons. Any improvement there can yield performance gains for stochastic video prediction with structure and motion.

Our StretchBEV has a powerful posterior distribution that leads to remarkable results. Although using labels in the posterior significantly improves the results in almost all metrics, it introduces a dependency on the availability of labels in the conditioning frames during inference. Future work on learning dynamics should focus on closing the gap between the two approaches, for example with scheduled sampling.

StretchBEV method can be interpreted as a Neural-ODE \cite{Chen2018NeurIPS} because of its residual update dynamics. In our model, we use only one update in between time steps but in future, we plan to explore increasing the number of updates in between time step as done in the previous work~\cite{Franceschi2020ICML}. Moreover, we think that focusing on objects more by following object-centric approaches might lead to better predictions.

One might explore driving policies that can utilize stochastic future predictions. Learned latent states at each time step can be directly fed into a policy learning algorithm, \eg as states in deep reinforcement learning. Furthermore, these states can be interpreted via supervised decoding into various future modalities that StretchBEV predicts.

\newpage
\bibliographystyle{apalike}
\bibliography{bibliography_long, references}

\newpage
\appendix

\end{document}